\documentclass[10pt,twocolumn,letterpaper]{article}

\usepackage{cvpr}              

\usepackage{graphicx}
\usepackage{amsmath}
\usepackage{amssymb}
\usepackage{booktabs}

\usepackage{makecell}
\usepackage{colortbl}
\usepackage{array} 
\usepackage[accsupp]{axessibility}  

\usepackage{marvosym}

\def\blu#1{\textbf{\color{blue} #1}} 
\def\red#1{\textbf{\color{red}  #1}} 
\usepackage{multirow}
\usepackage{makecell}
\usepackage{booktabs}  

\usepackage{colortbl}  
\usepackage{xcolor}

\newcommand{\Paragraph}[1]{\noindent\textbf{#1}}

\usepackage[pagebackref,breaklinks,colorlinks]{hyperref}

\usepackage[capitalize]{cleveref}
\crefname{section}{Sec.}{Secs.}
\Crefname{section}{Section}{Sections}
\Crefname{table}{Table}{Tables}
\crefname{table}{Tab.}{Tabs.}

\def\cvprPaperID{9516} 
\def\confName{CVPR}
\def\confYear{2023}

\begin{document}

\title{Discriminative Co-Saliency and Background Mining Transformer for Co-Salient Object Detection}

\author{
    Long Li$^{1}$\enspace 
    Junwei Han$^{1}$\enspace 
    Ni Zhang$^{1}$\enspace
    Nian Liu$^{2}$\footnotemark[2]
    \\
    Salman Khan$^{2,3}$\enspace
    Hisham Cholakkal$^{2}$\enspace
    Rao Muhammad Anwer$^{2}$\enspace 
    Fahad Shahbaz Khan$^{2,4}$ 
    \\
    $^1$ Northwestern Polytechnical University\enspace 
    $^2$ Mohamed bin Zayed University of Artificial Intelligence
    \\
    $^3$Australian National University\enspace 
    $^4$CVL, Linköping University
}

\maketitle
\footnotetext[2]{Corresponding author: liunian228@gmail.com.}
\maketitle

\vspace{-8mm}
\begin{abstract}
Most previous co-salient object detection works
mainly focus on extracting co-salient cues via mining the consistency relations across images while ignoring \textbf{explicit} exploration of background regions. In this paper, we propose a Discriminative co-saliency and background Mining Transformer framework (DMT)
based on several economical multi-grained correlation modules to \textbf{explicitly} mine both co-saliency and background information and
effectively model their discrimination.
Specifically, we first propose a region-to-region correlation module for introducing inter-image relations to pixel-wise segmentation features while maintaining computational efficiency. Then, we use two types of pre-defined tokens to mine co-saliency and background information via our proposed contrast-induced pixel-to-token correlation 
and co-saliency token-to-token correlation modules. We also design a token-guided feature refinement module to enhance the discriminability of the segmentation features under the guidance of the learned tokens. We perform iterative mutual promotion for the segmentation feature extraction and token construction. Experimental results on three benchmark datasets demonstrate the effectiveness of our proposed method. The source code is available at: \href{https://github.com/dragonlee258079/DMT}{https://github.com/dragonlee258079/DMT}.
\end{abstract}

\vspace{-4mm}
\section{Introduction}
\vspace{-1mm}
\label{sec:intro} 
Unlike standard Salient Object Detection (SOD) \cite{liu2016dhsnet,liu2018picanet,liu2021visual,fang2021densely,zhuge2022salient,zhang2019synthesizing}, which detects salient objects in a single image, Co-Salient Object Detection (CoSOD) aims to detect common salient objects across a group of relevant images. It often faces the following two challenges: 1) The foreground (FG) in CoSOD refers to co-salient objects, which are inherently hard to detect since they should satisfy both the intra-group commonality and the intra-image saliency. 2) The background (BG) in CoSOD might contain complex distractions, including extraneous salient objects that are salient but not "common", and similar concomitant objects appearing in multiple images (\eg performers often appear in guitar images). Such difficult distractors can easily seduce CoSOD models to make false positive predictions. Therefore, the effective exploration of both FG and BG and modeling their discrimination to precisely detect co-salient objects while suppressing interference from BG is crucial for CoSOD.

Although many works have achieved promising performance, most of them \cite{wei2017group, wang2019robust, li2019detecting, ren2020co, gao2020co, zhang2020adaptive, zhang2020gicd, zhang2021deepacg, zhang2021summarize} devoted to the ingenious mining of FG while ignored the \textbf{explicit} exploration of BG. They mainly constructed positive relations between co-salient objects but paid less attention to modeling negative ones between co-saliency regions and BG. \cite{jin2020icnet,yu2022democracy} followed some SOD methods \cite{chen2018reverse, zhang2021bilateral} to incorporate BG features for co-saliency representation learning or contrast learning. However, these methods can be regarded as an univariate FG\&BG modeling in which the essential optimization target is limited to FG and there is no explicit BG modeling, thus limiting the discriminative learning capability. To this end, this paper propose to conduct a bivariate FG\&BG modeling paradigm that \textbf{explicitly} models both FG and BG information and effectively facilitates their discriminative modeling.

As for co-saliency (FG) information, most previous works \cite{jin2020icnet,fan2021GCoNet,zhang2021summarize,yu2022democracy} detected them by exploring the inter-image similarity. Calculating the Pixel-to-Pixel (P2P) correlation between 3D CNN features in a group is widely used in many works \cite{jin2020icnet,fan2021GCoNet,yu2022democracy} and has demonstrated its effectiveness.
However, this method introduces heavy computation burdens and hinders sophisticated relation modeling. To alleviate this problem, we introduce economic multi-grained correlations among different images and the co-saliency and BG information, thus enabling modeling sophisticated relations to extract accurate co-saliency as well as BG knowledge.

Specifically, we construct a Discriminative co-saliency and BG Mining Transformer (DMT) following the paradigm of a semantic segmentation transformer architecture, \ie MaskFormer \cite{cheng2021per}, which enables explicit co-saliency and BG modeling and the construction of multi-grained correlations. Using this architecture, we decompose the CoSOD modeling into two sub-paths, \ie generating pixel-wise segmentation feature maps and extracting category information with pre-defined co-saliency and BG detection tokens.

In the first sub-path, to efficiently and thoroughly mine the common cues within the image group, we propose a Region-to-Region correlation (R2R) module to model the inter-image relation and plug it into each decoder layer. In the second sub-path, we transform the pixel-wise features into a co-saliency token and a BG token for each image, abstracting pixel-wise cues into high-level tokens. As such, we achieve sophisticated relation modeling among the tokens and features while largely reducing the computational costs. Concretely, we propose an intra-image Contrast-induced Pixel-to-Token correlation (CtP2T) module to extract the two tokens by considering the contrast relation between co-saliency and BG. Since the co-saliency tokens from CtP2T are separately learned on each image, we further design a Co-saliency Token-to-Token (CoT2T) correlation module to model their common relation.

After obtaining the tokens and pixel-wise features, the MaskFormer \cite{cheng2021per} architecture adopts dot production between them to obtain the final segmentation results.
However, such a scheme only achieves unidirectional information propagation, \ie conveying information from the feature maps to the tokens. We argue that the learned two tokens can also be used to improve the discriminability of the pixel-wise features, thus proposing our Token-Guided Feature Refinement (TGFR) module as a reverse information propagation path. Concretely, we first use the tokens as guidance to distill co-saliency and BG features from the pixel-wise feature maps, and then enhance the discriminability of the segmentation features between the two detection regions. In this way, the refined features become sensitive to both co-saliency and BG, reducing the affect of ambiguous distractors.

\begin{figure*}[t!]
\centering
\includegraphics[width=1\linewidth]{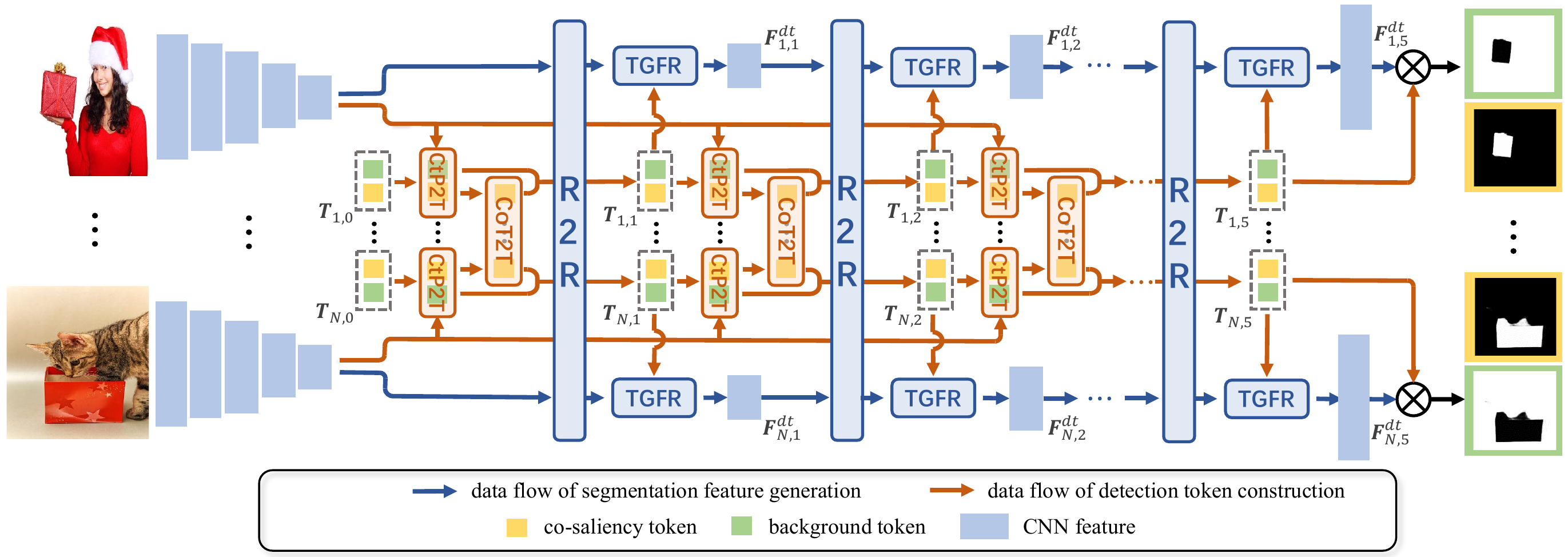}
\vspace{-7mm}
\caption{\textbf{Overall flowchart of our proposed DMT CoSOD model.} Specifically, the framework consists of four components, \ie R2R for segmentation feature generation, CtP2T and CoT2T for detection token construction, and TGFR for the segmentation feature refinement under the guidance of the tokens.
}
\label{fig:overview}
\vspace{-4mm}
\end{figure*}

Finally, as shown in Figure~\ref{fig:overview}, our DMT iteratively deploys CtP2T and CoT2T to leverage the segmentation features for updating the tokens, and then adopts TGFR to refine the corresponding decoder feature with the updated tokens. As a result, the learning processes can be effectively promoted, thus obtaining more accurate CoSOD results.

In summary, our major contributions are as follows:
\vspace{-2mm}
\begin{itemize}
\item We model CoSOD from the perspective of explicitly exploring both co-saliency and BG information and effectively modeling their discrimination.
\vspace{-2mm}
\item We introduce several computationally economical multi-grained correlation modules, \ie R2R, CtP2T, CoT2T, for  
the inter-image and intra-image relations modeling.
\vspace{-2mm}
\item We propose a novel TGFR module to use the learned tokens as guidance to refine the segmentation features for enhancing their discriminability between co-saliency and BG regions.
\vspace{-2mm}
\item Experimental results demonstrate that our DMT model outperforms previous state-of-the-art results on three benchmark datasets.
\end{itemize}

\vspace{-2mm}
\section{Related Work}
\vspace{-1mm}
\subsection{Co-Salient Object Detection}
\vspace{-1mm}

Recent CoSOD works \cite{zhang2020gicd,jin2020icnet,fan2021GCoNet,zhang2021summarize,su2022unified, yu2022democracy} have achieve promising performance and can be summarized as a unified paradigm, \ie first aggregating all image features in the group to form a consensus representation and then distributing it back to each image feature. We refer to these two processes as \emph{aggregation} and \emph{distribution} for expression convenience. 
For example, \cite{zhang2020gicd} summed up all features for \emph{aggregation} and leveraged a gradient feedback mechanism for \emph{distribution}. \cite{jin2020icnet} formed the consensus cues with a group of enhanced intra-saliency vectors and conducted the \emph{distribution} via a dense correlation module. \cite{fan2021GCoNet} generated a consensus attention map with an affinity module and multiplied it back to the individual image features. \cite{zhang2021summarize} encoded the consensus information with dynamic kernels and convolved the image features using these kernels as the \emph{distribution} process. \cite{yu2022democracy} first obtained consensus seeds by processing P2P affinity maps and then propagated the seeds using normalized convolution operations. However, most of them are limited in exploring BG regions, which hinders the discriminability learning. Unlike them, we propose to simultaneously detect the co-saliency and BG regions and sufficiently explore their discriminative modeling. Besides, we utilize tokens under a transformer architecture for \emph{aggregation}, and then use the learned tokens to conduct the \emph{distribution} process. 

\subsection{Transformer}
\vspace{-1mm}

After Vaswani \etal \cite{vaswani2017attention} first proposed the transformer architecture 
for machine translation, many successful transformer applications in the computer vision field emerged. Some works \cite{dosovitskiy2020image,touvron2021training,yuan2021tokens} directly apply the transformer architecture for feature learning. 
Some other works mainly focused on using transformers to extract specific semantic concepts, \eg the category or instance information for object detection \cite{carion2020end,zhu2020deformable,wang2021end}, semantic segmentation \cite{wang2021max,cheng2021per}, and the saliency and contour information for salient object detection \cite{liu2021visual}. 
Concretely, they first use a backbone to extract image feature maps and then adopt transformers to collect semantic concept information and store them in pre-created tokens. 

This paper follows the second type application 
to utilize the transformer for simultaneous FG and BG modeling. We further modify the transformer framework tailored for the CoSOD task by introducing economic multi-grained correlations for modeling sophisticated relations. We also propose to leverage the semantic information encoded in the learned tokens as a guide to refine the features, thus improving its discriminability. 

\section{Proposed Method}
\vspace{-1mm}
\subsection{Overview}
\vspace{-1mm}

Figure~\ref{fig:overview} illustrates our MaskFormer-style framework
for simultaneously detecting co-salient objects and BG regions.
It consists of two sub-paths, \ie pixel-wise segmentation feature generation and detection token construction. We use R2R in the first sub-path to enhance the segmentation features with inter-image consistency. 
In the second sub-path, CtP2T and CoT2T are designed to effectively construct the co-saliency and BG tokens from segmentation features, capturing the binary detection patterns. Finally, we propose TGFR to use the detection tokens as guidance for refining the segmentation features.

For ease of understanding, we first briefly describe
the vanilla MaskFormer-style framework for simultaneously detecting co-saliency and BG in CoSOD. 
Then, we progressively introduce the improvements in our proposed DMT, including R2R, CtP2T, CoT2T, and TGFR.

\subsection{Vanilla MaskFormer-style Framework}
\vspace{-1mm}
\label{Baseline}  
\subsubsection{Segmentation Feature Generation}
\vspace{-1mm}
Given a set of $N$ relevant images $\left\{\boldsymbol{I}_i \right\}^N_{i=1}$, 
we follow the original MaskFormer framework \cite{cheng2021per} to adopt an FPN \cite{lin2017feature} for generating pixel-wise segmentation features.
Specifically, we use VGG-16\cite{simonyan2014very} as the encoder and take $\left\{\boldsymbol{I}_i \right\}^N_{i=1}$ as the input to obtain the highest-level features $\boldsymbol{F}^e \in \mathbb{R}^{N \times H_0 \times W_0 \times C}$ from the last block.
Then, based on $\boldsymbol{F}^e$, the FPN decoder uses five decoder layers to progressively enlarge the feature resolution and obtain five decoder features
$\boldsymbol{F}^d_j \in \mathbb{R}^{N \times H_j \times W_j \times C}, j \in \{1 \cdots 5\}$. 

\vspace{-3mm}
\subsubsection{Detection Token Construction}
\vspace{-1mm}
Given the highest-level semantic feature $\boldsymbol{F}^e_i \in \mathbb{R}^{H_0 \times W_0 \times C}$ of image $\boldsymbol{I}_i$, 
we extract the detection tokens from it via a vanilla pixel-to-token correlation (P2T) module. First, we define two randomly initialized tokens for $\boldsymbol{I}_i$, \ie 
a co-saliency token $\boldsymbol{T}_{i, 0}^{c} \in \mathbb{R}^{1 \times C}$ and a BG token $\boldsymbol{T}_{i, 0}^{b} \in \mathbb{R}^{1 \times C}$, and denote their union as $\boldsymbol{T}_{i, 0} \in \mathbb{R}^{2 \times C}$.
We also flatten $\boldsymbol{F}^e_i$ along the spatial dimension as $\hat{\boldsymbol{F}^e_i} \in \mathbb{R}^{H_0W_0 \times C}$.
Then, we iteratively update the tokens five times. At each iteration $j \in \{1,...,5\}$, we obtain $\boldsymbol{T}_{i,j}$ by transforming the information from the feature $\hat{\boldsymbol{F}^e_i}$ to tokens in \eqref{F2T} and modeling the relationship between the co-saliency and BG tokens in \eqref{T2T}, formulated as
\vspace{-2mm}
\begin{equation}
\label{F2T}
\hat{\boldsymbol{T}}_{i, j} = \operatorname{Trans}(\boldsymbol{T}_{i, j-1}, \hat{\boldsymbol{F}^e_i}),
\end{equation}
\vspace{-6mm}
\begin{equation}
\label{T2T}
\boldsymbol{T}_{i, j} = \operatorname{Trans}(\hat{\boldsymbol{T}}_{i, j}, \hat{\boldsymbol{T}}_{i, j}),
\end{equation}
where $\operatorname{Trans}$ is a basic transformer operation following \cite{vaswani2017attention}:
\vspace{-2mm}
\begin{equation}
\label{Trans}
\operatorname{Trans}(\boldsymbol{X}, \boldsymbol{Y}) = \operatorname{rMLP}(\operatorname{rMHA}(\boldsymbol{X}, \boldsymbol{Y})).
\end{equation}
It can transfer the information from $\boldsymbol{Y} \in \mathbb{R}^{N_y \times C}$ to $\boldsymbol{X} \in \mathbb{R}^{N_x \times C}$
 under the guidance of their relation. 
$\operatorname{rMHA}$ and $\operatorname{rMLP}$ denote the residual multi-head attention \cite{vaswani2017attention} and residual multi-layer perception, respectively, formulated as
\begin{equation}
\operatorname{rMLP}(\boldsymbol{X}) = \boldsymbol{X} + \operatorname{MLP}(\operatorname{LN}(\boldsymbol{X})),
\end{equation}
\vspace{-9mm}
\begin{equation}
\label{rHMA}
\begin{gathered}
\operatorname{rMHA}(\boldsymbol{X}, \boldsymbol{Y}) = \boldsymbol{X} + \operatorname{MHA}(\operatorname{LN}(\boldsymbol{X}), \operatorname{LN}(\boldsymbol{Y}))),
\end{gathered}
\end{equation}
where $\operatorname{LN}$ denotes the layer normalization \cite{ba2016layer} and $\operatorname{MLP}$ is the multi-layer perception consisting of two fully connected layers with a GELU \cite{hendrycks2016gaussian} activation function.
$\operatorname{MHA}$ is the multi-head attention that can be formulated as
\vspace{-2mm}
\begin{equation}
\label{MHA}
\operatorname{MHA}(\boldsymbol{X},\boldsymbol{Y})=\operatorname{Cat}(\left[\operatorname{Att}_m(\boldsymbol{X}, \boldsymbol{Y}) V_m(\boldsymbol{Y})\right]_{m=1}^{M}),
\end{equation}
\vspace{-8mm}
\begin{equation}
\label{Att}
\operatorname{Att}_m,(\boldsymbol{X},\boldsymbol{Y}) = \operatorname{Softmax}\left(\frac{Q_m(\boldsymbol{X}) K_m(\boldsymbol{Y})^{\top}}{\sqrt{C/M}}\right),
\end{equation}
where $M$ is the number of used attention heads. The result of each head (with the shape of $N_x\!\times \! C/M$) is obtained via the matrix multiplication between $\operatorname{Att}_m(\boldsymbol{X},\boldsymbol{Y}) \in \mathbb{R}^{N_x \times N_y}$ and $V_m(\boldsymbol{Y}) \in \mathbb{R}^{N_y \times C/M}$. $\operatorname{Att}_m(\boldsymbol{X},\boldsymbol{Y})$ is the attention matrix calculated in \eqref{Att}. Here $Q_m(\cdot)$, $K_m(\cdot)$, and $V_m(\cdot)$ are the query, key, and value embedding functions in the $m$th head, respectively, and project corresponding tensors from $C$ channels to $C/M$ channels. 
Finally, $\operatorname{MHA}(\boldsymbol{X}, \boldsymbol{Y}) \in \mathbb{R}^{N_x \times C}$ can be obtained by concatenating ($\operatorname{Cat}$) the results of $M$ heads along the channel dimension.

\vspace{-4mm}
\subsubsection{Prediction}
\vspace{-1mm}
After performing the token construction five times on each image, we collect the final tokens of all images, \ie $\boldsymbol{T}_5^c,\boldsymbol{T}_5^b \in \mathbb{R}^{N \times 1 \times C}$. 
Then, we use the output of the first sub-path, \ie the segmentation feature $\boldsymbol{F}^d_5$, to generate the final predictions via the sigmoid matrix multiplication, formulated as
\vspace{-2mm}
\begin{equation}
\label{Prediction}
\boldsymbol{P}^c = \mathcal{P}(\boldsymbol{T}_5^c, \boldsymbol{F}^d_5) = \operatorname{Sigmoid}(\boldsymbol{T}^c_5(\boldsymbol{F}^d_5)^{\top}),
\end{equation}
\vspace{-6mm}
\begin{equation}
\boldsymbol{P}^b =  \mathcal{P}(\boldsymbol{T}_5^b, \boldsymbol{F}^d_5) =\operatorname{Sigmoid}(\boldsymbol{T}^b_5(\boldsymbol{F}^d_5)^{\top}),
\end{equation}
where $\boldsymbol{P}^c, \boldsymbol{P}^b \in \mathbb{R}^{N \times 1 \times H \times W}$ are the segmentation results of co-salient objects and BG regions, respectively.

\subsection{Our Improvements for DMT}
\subsubsection{Region-to-Region Correlation}
\vspace{-1mm}
In the first sub-path, the original FPN individually processes each image and lacks the inter-image correlation modeling, which is crucial for CoSOD. However, straightforward P2P correlation is computationally prohibitive for large feature maps and multiple images. To this end, we consider modeling correlations among images in an economical way, thus proposing our R2R module, which uses region-level features instead of pixel-level features to compute correlations.

Concretely, when given the features $\boldsymbol{F}^d_j \in \mathbb{R}^{N \times H_j \times W_j \times C}$ of $N$ relevant images from the $j$th decoder layer, 
we first adopt a transformation $\operatorname{R}_1$ to
divide the $H_j \! \times \! W_j$ feature maps into $K \! \times \! K$ local regions and use max-pooling to pick up the most representative feature for representing each local region. As a result, we can obtain the region-level query with shape $\mathbb{R}^{N \times K \times K \times C}$.

Then, we generate multi-scale region-level key and value via another transformation $\operatorname{R}_2$, which consists of three adaptive max-pooling operations with the output spatial sizes of $1 \times 1$, $3 \times 3$, and $6 \times 6$, respectively. The three pooled features are finally flattened and concatenated to generate the key and value with shape $\mathbb{R}^{N \times 46 \times C}$, encoding multi-scale robust region information.

Next, we perform the R2R inter-image correlation among the region-level query, key, and value via the transformer operation \eqref{Trans}, thus obtaining the enhanced features with the region-wise correlation. 

Finally, we upsample the enhanced features to the original resolution $H_j \! \times \! W_j$ via the nearest interpolation, denoted as $\operatorname{R}_1^{-1}$. A residual connection is also used to add the original features. Thus, the region correlation results are diffused to the corresponding internal pixels in each local region. 
The whole process of R2R on $\boldsymbol{F}^d_j$ is formulated as
\vspace{-2mm}
\begin{equation}
\boldsymbol{F}_j^{dr} = \boldsymbol{F}_j^{d} + \operatorname{R}_1^{-1}(\operatorname{Trans}(\operatorname{R}_1(\boldsymbol{F}_j^{d}), \operatorname{R}_2(\boldsymbol{F}_j^{d}))).
\end{equation}
\vspace{-8mm}

\vspace{-2mm}
\subsubsection{Contrast-induced Pixel-to-Token Correlation}
\vspace{-1mm}
In the second sub-path, the original P2T module uses a transformer operation in \eqref{T2T} to mine relations between the two types of tokens
in a data-driven way, while ignoring explicit CoSOD cues, especially the crucial contrast modeling between co-saliency and BG regions.
To enhance the discriminability between the tokens, we explicitly model the contrast relation with our proposed CtP2T module, which modifies the transformer layer in \eqref{F2T} and the remaining part keeps the same as P2T.

\begin{figure}
\centering
\includegraphics[width=1.0\linewidth]{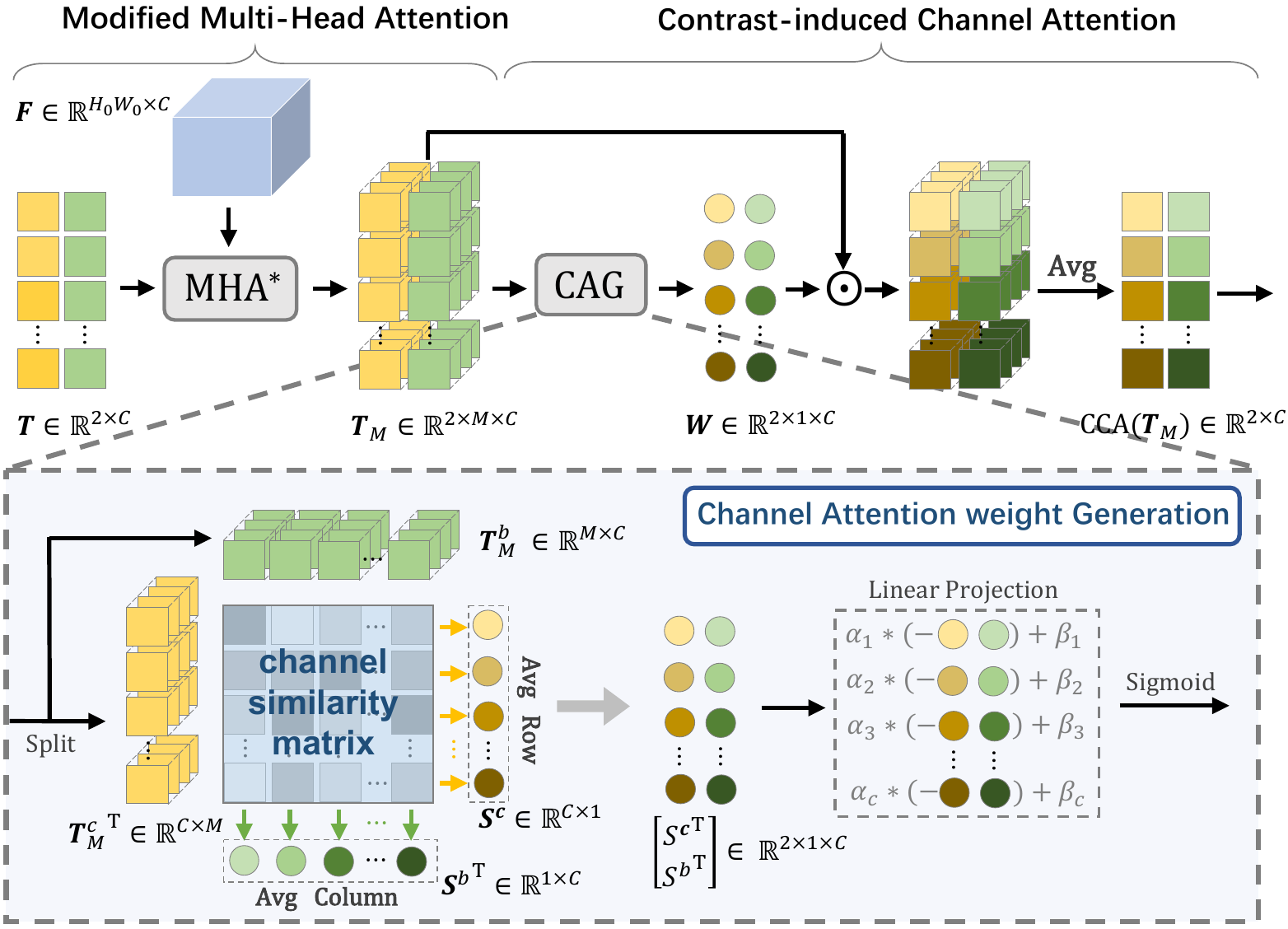}
\vspace{-7mm}
\caption{\textbf{Diagram of $\operatorname{MHA}^*$ and $\operatorname{CCA}$.} We first generate multi-head tokens $\boldsymbol{T}_M^c$ and $\boldsymbol{T}_M^b$ via $\operatorname{MHA}^*$. Then, we utilize matrix multiplication of the two tokens to generate the attention weights $\boldsymbol{W}$ for modulating the token channels in $\operatorname{CCA}$.}
\vspace{-4mm}
\label{fig:modifiedTrans}
\end{figure}

Overall, we modify the multi-head attention (denoted as $\operatorname{MHA}^*$) and 
propose a contrast-induced channel attention (CCA) mechanism. The basic idea is to suppress the channels that are not contrastive enough in the generated co-saliency and BG tokens. The contrast is modeled as the opposite of the channel similarity between the two types of tokens, which can be calculated via channel correlation. For brevity's sake, we slightly abuse the notation and use $\hat{\boldsymbol{T}}$, $\boldsymbol{T} \in \mathbb{R}^{2 \times C}$, and $\boldsymbol{F} \in \mathbb{R}^{H_0W_0 \times C}$ as shorthands for $\hat{\boldsymbol{T}}_{i, j}$, $\boldsymbol{T}_{i, j-1}$, and $\hat{\boldsymbol{F}^e_i}$ in \eqref{F2T}, respectively. Then, \eqref{F2T} can be modified for our CtP2T as below:
\begin{equation}
\begin{aligned}
\label{modifiedTrans}
\hat{\boldsymbol{T}} &= \operatorname{Trans}^{*}(\boldsymbol{T}, \boldsymbol{F})\\
&= \operatorname{rMLP}(\boldsymbol{T} + \operatorname{CCA}(\operatorname{MHA}^*(\boldsymbol{T}, \boldsymbol{F}))).
\end{aligned}
\end{equation}
Next, we introduce $\operatorname{MHA}^*$ and CCA as shown in Figure~\ref{fig:modifiedTrans}. The $\operatorname{LN}$ operations are omitted for expression convenience.

\vspace{-3mm}
\paragraph{Modified Multi-Head Attention.}
To generate co-saliency and BG tokens that can be used for calculating their channel similarity, we make our $\operatorname{MHA}^*$ able to generate tokens with multiple heads.
Concretely, we first replace the original $V_m$ in \eqref{MHA} with $V^*_m$ that embeds $\boldsymbol{F}$ to the identical channel number $C$. Thus, the shape of each head's result becomes $2 \! \times \! C$ instead of $2 \! \times \! C/M$. Next, we stack the results of $M$ heads to produce the output of $\operatorname{MHA}^*$. The whole process can be formulated as
\begin{equation}
\begin{aligned}
\boldsymbol{T}_M&=\operatorname{MHA}^*(\boldsymbol{T}, \boldsymbol{F})\\
&= \operatorname{Stack}(\left[\operatorname{Att}_m(\boldsymbol{T}, \boldsymbol{F})V^*_m(\boldsymbol{F})\right]_{m=1}^{M}).
\end{aligned}
\end{equation}
$\boldsymbol{T}_M \in \mathbb{R}^{2 \times M \times C}$ is composed of the co-saliency token  and the BG token $\boldsymbol{T}_M^c,\boldsymbol{T}_M^b \in \mathbb{R}^{M \times C}$ with $M$ heads.
Next, we can compute the channel similarity based on them.

\vspace{-3mm}
\paragraph{Contrast-induced Channel Attention.}
Given the multi-head tokens $\boldsymbol{T}_M^c$ and $\boldsymbol{T}_M^b$, we generate channel attention $\boldsymbol{W} \in \mathbb{R}^{2 \times 1 \times C}$ to suppress the token channels with
strong \emph{mutual} similarities since they cannot clearly distinguish between co-saliency and BG.

First, 
we compute a $C\!\times \!C$ \emph{channel similarity matrix} between $\boldsymbol{T}_M^c$ and $\boldsymbol{T}_M^b$ via matrix multiplication. Then, the channel similarity of each token to the other token can be computed as the average along the channel dimension of the other token.
The whole process can be denoted as
\vspace{-2mm}
\begin{equation}
\label{simi_c}
    \boldsymbol{S}^c = \operatorname{Avg}({\boldsymbol{T}_M^c}^{\top}\boldsymbol{T}_M^b),
\end{equation}
\vspace{-5mm}
\begin{equation}
\label{simi_b}
    \boldsymbol{S}^b = \operatorname{Avg}({\boldsymbol{T}_M^b}^{\top}\boldsymbol{T}_M^c),
\end{equation}
where $\boldsymbol{S}^c,\boldsymbol{S}^b \in \mathbb{R}^{C\times 1}$, representing how similar each channel is to the channels of the other token. $\operatorname{Avg}$ means calculating the average along the second dimension. 

Next, we multiply $\boldsymbol{S}^c$ and $\boldsymbol{S}^b$ with $-1$ to turn the similarity measurements into the \emph{contrast} scores and 
then compute the channel attention $\boldsymbol{W} \in \mathbb{R}^{2 \times 1 \times C}$ via 
\begin{equation}
\boldsymbol{W} = \operatorname{Sigmoid}\left(\alpha\left[\!\!\begin{array}{l}-{\boldsymbol{S}^c}^{\top} \\ -{\boldsymbol{S}^b}^{\top}\end{array}\!\!\right] + \beta\right).
\end{equation}
Here we use a learnable linear projection with parameters $\alpha,\beta$ on each channel of the stacked contrast scores to fit them for the sigmoid activation.

Once obtained $\boldsymbol{W}$, we adopt the element-wise multiplication between $\boldsymbol{T}_M$ and $\boldsymbol{W}$ to modulate the token channels based on their contrast 
and then eliminate the multi-head dimension of the tokens 
by averaging the head dimension and obtaining the modulated tokens:
\begin{equation}
\operatorname{CCA}(\boldsymbol{T}_M) = \operatorname{Avg}(\boldsymbol{W} \odot \boldsymbol{T}_M) \in \mathbb{R}^{2 \times C},
\end{equation}
where $\odot$ means element-wise multiplication with broadcasting.

\vspace{-3mm}
\subsubsection{Co-saliency Token-to-Token Correlation}
\vspace{-1mm}
CtP2T effectively explores the correlation between the two types of tokens within each image,
but lacks explicitly modeling the inter-image relation to capture the token-wise group consistency, thus being limited for consensus mining.
Therefore, we use co-saliency tokens from all images to model the consensus patterns via our CoT2T module.

Specifically, we first define a group token $\boldsymbol{G} \in \mathbb{R}^{1 \times C}$ to represent the group-wise consensus information, which is randomly initialized at the first iteration step. At the $j$th iteration, given the last group token $\boldsymbol{G}_{j-1}$ and the co-saliency tokens $\tilde{\boldsymbol{T}}_{j}^{c} \in \mathbb{R}^{N \times C}$ from the CtP2T module,
we aggregate the consensus information from all co-saliency tokens by using $\tilde{\boldsymbol{T}}_{j}^{c}$ to update $\boldsymbol{G}_{j-1}$, denoted as
\vspace{-2mm}
\begin{equation}
\label{aggregation}
\boldsymbol{G}_j = \operatorname{Trans}(\boldsymbol{G}_{j-1}, \tilde{\boldsymbol{T}}_{j}^{c}).
\end{equation}
\vspace{-5mm}

Finally, we distribute the aggregated consensus cues back to $\tilde{\boldsymbol{T}}_{j}^{c}$ and obtain the final co-saliency tokens $\boldsymbol{T}_{j}^{c}$:
\vspace{-2mm}
\begin{equation}
\label{distribution}
\boldsymbol{T}_{j}^{c} = \operatorname{Trans}(\tilde{\boldsymbol{T}}_{j}^{c}, \boldsymbol{G}_j).
\end{equation}
\vspace{-5mm}

\vspace{-4mm}
\subsubsection{Token-guided Feature Refinement}
\begin{figure}[t]
\centering
\includegraphics[width=1\linewidth]{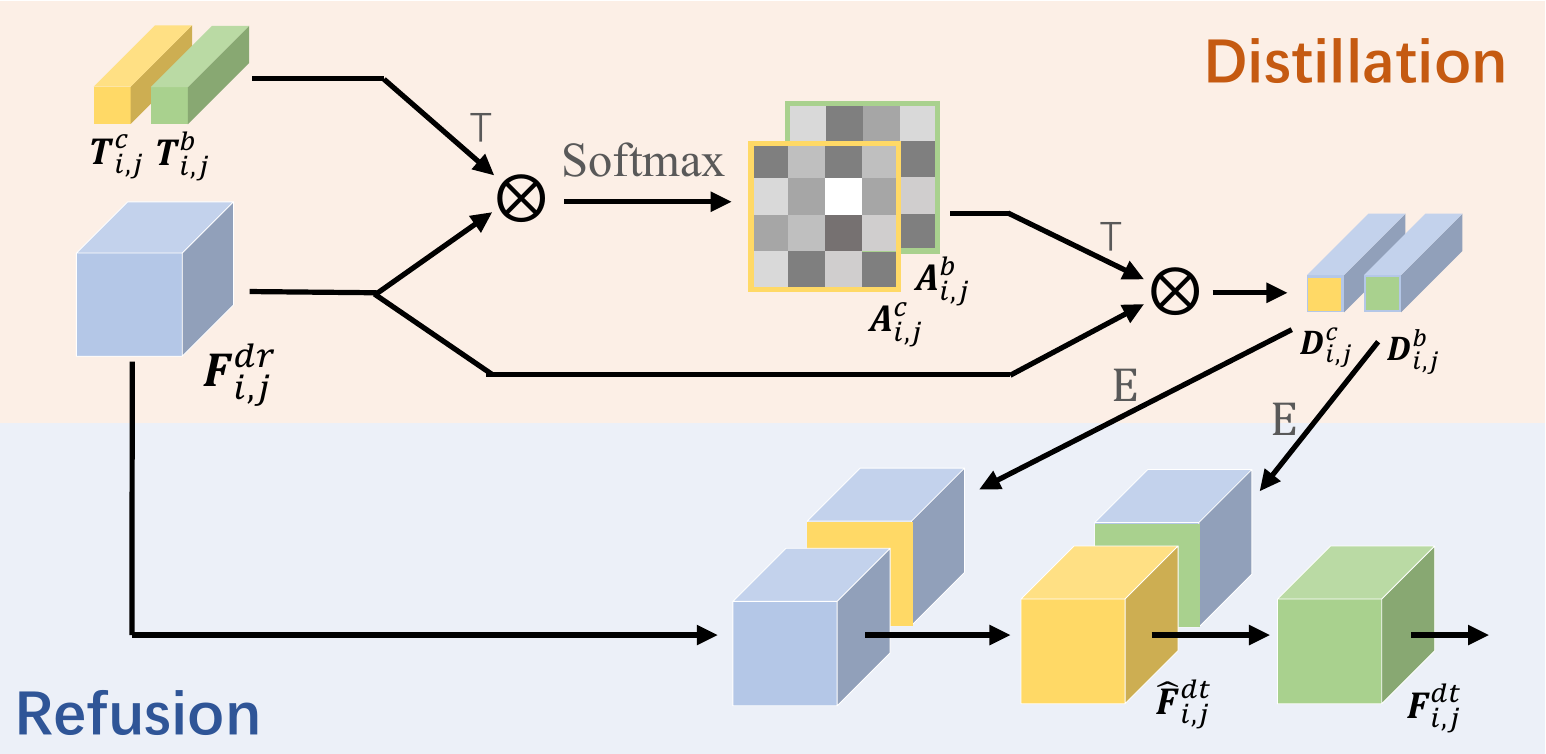}
\vspace{-7mm}
\caption{\textbf{Diagram of our proposed TGFR module.} Specifically, we first distill the co-saliency and BG features under the guidance of the two tokens. Then, we fuse them back to the original segmentation feature for discriminability enhancement.}
\label{fig:TGFR}
\vspace{-4mm}
\end{figure}

\vspace{-1mm}
The vanilla MaskFormer only transforms the information from the segmentation features to the tokens, hindering their complementary learning. To this end, we propose our TGFR module to improve the discriminability of the segmentation features via the detection cues of the tokens. As shown in Figure~\ref{fig:TGFR}, TGFR consists of two processes, \ie distillation and refusion. The distillation process is designed to distill the co-saliency and BG features from the segmentation feature under the guidance of the corresponding tokens. 
The refusion process is to fuse the distilled features back to the segmentation feature to enhance its discriminability.

\vspace{-3mm}
\paragraph{Distillation.}
For image $\boldsymbol{I}_i$ at the $j$th iteration, we have the final co-saliency token $\boldsymbol{T}_{i,j}^c \in \mathbb{R}^{1 \times C}$ generated from CoT2T and the final BG token $\boldsymbol{T}_{i,j}^{b} \in \mathbb{R}^{1 \times C}$ outputted by CtP2T, 
and the segmentation feature $\boldsymbol{F}_{i, j}^{dr} \in \mathbb{R}^{H_j \times W_j \times C}$ enhanced by R2R.
We first compute two attention maps $\boldsymbol{A}_{i, j}^{c} \in \mathbb{R}^{H_j \times W_j \times 1}$ and $\boldsymbol{A}_{i,j}^{b} \in \mathbb{R}^{H_j \times W_j \times 1}$ via performing the matrix multiplication between the segmentation feature and the tokens and then adopting a softmax normalization on the spatial dimension, formulated as
\vspace{-2mm}
\begin{equation}
\boldsymbol{A}_{i,j}^{c} = \operatorname{Softmax}(\boldsymbol{F}_{i,j}^{dr}(\boldsymbol{T}_{i,j}^c)^{\top}/\sqrt{C}),
\end{equation}
\vspace{-5mm}
\begin{equation}
\boldsymbol{A}_{i,j}^{b} = \operatorname{Softmax}(\boldsymbol{F}_{i,j}^{dr}(\boldsymbol{T}_{i,j}^{b})^{\top}/\sqrt{C}).
\end{equation}
Next, we adopt the computed attention maps to distill the detection features from the segmentation feature via matrix multiplication,
denoted as
\vspace{-2mm}
\begin{equation}
\boldsymbol{D}_{i,j}^{c} = (\boldsymbol{A}_{i,j}^{c})^{\top}\boldsymbol{F}_{i,j}^{dr},
\end{equation}
\vspace{-5mm}
\begin{equation}
\boldsymbol{D}_{i,j}^{b} = (\boldsymbol{A}_{i,j}^{b})^{\top}\boldsymbol{F}_{i,j}^{dr},
\end{equation}
where $\boldsymbol{D}_{i,j}^{c}, \boldsymbol{D}_{i,j}^{b} \in \mathbb{R}^{1 \times C}$ is the distilled features for co-saliency and BG, respectively. 

\vspace{-3mm}
\paragraph{Refusion.}
After producing $\boldsymbol{D}_{i,j}^{c}$ and $\boldsymbol{D}_{i,j}^{b}$, we conduct the refusion process to fuse them back to $\boldsymbol{F}_{i,j}^{dr}$ sequentially in a cascade way for activating the co-saliency and BG regions in $\boldsymbol{F}_{i,j}^{dr}$. In this way, we can effectively reduce ambiguous information and enhance feature discriminability. The details can be formulated as
\vspace{-2mm}
\begin{equation}
\label{TGFR_fc}
\hat{\boldsymbol{F}}_{i,j}^{dt} = \operatorname{Conv}_c\big(\operatorname{Cat}([\boldsymbol{F}_{i,j}^{dr},\, \operatorname{E}(\boldsymbol{D}_{i,j}^{c})])\big),
\end{equation}
\vspace{-5mm}
\begin{equation}
\label{TGFR_fb}
\boldsymbol{F}_{i,j}^{dt} = \operatorname{Conv}_b\big(\operatorname{Cat}([\hat{\boldsymbol{F}}_{i,j}^{dt}, \, \operatorname{E}(\boldsymbol{D}_{i,j}^{b})])\big),
\end{equation}
where $\operatorname{E}(*)$ replicates $\boldsymbol{D}_{i,j}^{c}$ and $\boldsymbol{D}_{i,j}^{b}$ along the spatial dimension to the same size as $\boldsymbol{F}_{i,j}^{dr}$. 
Then, we progressively concatenate them with $\boldsymbol{F}^{dr}_{i,j}$ and use a convolution layer to reduce the channel number to $C$.

\vspace{-3mm}
\subsubsection{Prediction and Loss Function}
\vspace{-1mm}
In the $j$th iteration,
after obtaining the learned
co-saliency and BG tokens, \ie $\boldsymbol{T}^c_j, \boldsymbol{T}^b_j \in \mathbb{R}^{N \times 1 \times C}$, from CoT2T and CtP2T, respectively, and the improved segmentation features $\boldsymbol{F}_{j}^{dt} \in \mathbb{R}^{N \times H_j \times W_j \times C}$ from TGFR,  
we use the prediction function $\mathcal{P}$ in \eqref{Prediction} to generate the co-saliency and the BG predictions, \ie $\boldsymbol{P}_j^c$ $\boldsymbol{P}_j^b$, as follows:
\vspace{-2mm}
\begin{equation}
\boldsymbol{P}_j^c = \mathcal{P}(\boldsymbol{T}^c_j, \boldsymbol{F}_{j}^{dt}),
\end{equation}
\vspace{-5mm}
\begin{equation}
\boldsymbol{P}_j^b = \mathcal{P}(\boldsymbol{T}^b_j, \boldsymbol{F}_{j}^{dt}).
\end{equation}
We also supervise the learning of 
the group token $\boldsymbol{G}_j \in \mathbb{R}^{1 \times 1 \times C}$ in CoT2T
and the middle feature $\hat{\boldsymbol{F}}^{dt}_j \in \mathbb{R}^{N \times H_j \times W_j \times C}$ in TGFR.
Two predictions can be obtained from them, respectively:
\begin{equation}
\begin{split}
\boldsymbol{P}^{g}_j &= \mathcal{P}(\operatorname{Repeat}(\boldsymbol{G}_j), \boldsymbol{F}_{j}^{dt}), \\   
\boldsymbol{P}^{dt}_j &= \mathcal{P}(\boldsymbol{T}^c_j, \hat{\boldsymbol{F}}_{j}^{dt}),
\end{split}
\end{equation}
where $\operatorname{Repeat}$ is to repeat $\boldsymbol{G}_j$ $N$ times. 

Our total loss $\mathcal{L}_{total}$ can be formulated as
\vspace{-2mm}
\begin{equation}
\begin{gathered}
\mathcal{L}_{total} =  \sum_{j=1}^{5} \Big(
\mathcal{L}_{1}(\boldsymbol{P}_j^c, \boldsymbol{M}_j^c) + 
\mathcal{L}_{2}(\boldsymbol{P}_j^c, \boldsymbol{M}_j^c) + \\
\mathcal{L}_{2}(\boldsymbol{P}_j^b, \boldsymbol{M}_j^b) + 
\mathcal{L}_{2}(\boldsymbol{P}_j^g, \boldsymbol{M}_j^c) + 
\mathcal{L}_{2}(\boldsymbol{P}_j^{dt}, \boldsymbol{M}_j^c) \Big),
\end{gathered}
\end{equation}
where $\mathcal{L}_{1}$ and $\mathcal{L}_{2}$ are the IoU \cite{jin2020icnet} and Binary Cross-Entropy (BCE) \cite{de2005tutorial} loss, respectively. $\boldsymbol{M}_j^c$ and $\boldsymbol{M}_j^b$ denote the co-saliency and BG ground truths, respectively, with the spatial shapes aligned to the $j$th decoder layer. 

\begin{table}[t]
\centering
\footnotesize
\renewcommand{\arraystretch}{0.6}
\renewcommand{\tabcolsep}{0.9mm}
\caption{\textbf{Quantitative results of different settings of our proposed model.} We show the results of progressively adding R2R, CtP2T, CoT2T, and TGFR on the baseline. ``Co" and ``Bg" mean explicitly modeling co-saliency and BG, respectively.}
\vspace{-2mm}
\begin{tabular}{c|c|c|c|c|c|cccc}
\toprule

\multicolumn{6}{c|}{Settings} & 
\multicolumn{4}{c}{CoCA \cite{zhang2020gicd}}  \\

\midrule

\multicolumn{1}{c|}{Co} &
\multicolumn{1}{c|}{Bg} &
\multicolumn{1}{c|}{R2R} &
\multicolumn{1}{c|}{CtP2T} &
\multicolumn{1}{c|}{CoT2T} &
\multicolumn{1}{c|}{TGFR} &
\multicolumn{1}{c}{$S_m \uparrow$} &  
\multicolumn{1}{c}{$E_\xi\uparrow$} &
\multicolumn{1}{c}{maxF$\uparrow$} & 
\multicolumn{1}{c}{MAE $\downarrow$}\\ \midrule

\multicolumn{1}{c|}{\checkmark} &
\multicolumn{1}{c|}{\checkmark} &
\multicolumn{1}{c|}{} &
\multicolumn{1}{c|}{} &
\multicolumn{1}{c|}{} &
\multicolumn{1}{c|}{} 
&0.6751 &0.7683 &0.5474 &0.1383 \\ \midrule

\multicolumn{1}{c|}{\checkmark} &
\multicolumn{1}{c|}{\checkmark} &
\multicolumn{1}{c|}{\checkmark} &
\multicolumn{1}{c|}{} &
\multicolumn{1}{c|}{} &
\multicolumn{1}{c|}{}   
&0.6945 &0.7824 &0.5815 &0.1234 \\ \midrule

\multicolumn{1}{c|}{\checkmark} &
\multicolumn{1}{c|}{\checkmark} &
\multicolumn{1}{c|}{\checkmark} &
\multicolumn{1}{c|}{\checkmark} &
\multicolumn{1}{c|}{} &
\multicolumn{1}{c|}{}     
&0.7038 &0.7868 &0.5984 &0.1230 \\ 

\multicolumn{1}{c|}{\checkmark} &
\multicolumn{1}{c|}{\checkmark} &
\multicolumn{1}{c|}{\checkmark} &
\multicolumn{1}{c|}{\checkmark} &
\multicolumn{1}{c|}{\checkmark} &
\multicolumn{1}{c|}{}    
&0.7140 &0.7880 &0.6003 &0.1139   \\ \midrule

\rowcolor[RGB]{214,220,229}
\multicolumn{1}{c|}{\checkmark} &
\multicolumn{1}{c|}{\checkmark} &
\multicolumn{1}{c|}{\checkmark} &
\multicolumn{1}{c|}{\checkmark} &
\multicolumn{1}{c|}{\checkmark} &
\multicolumn{1}{c|}{\checkmark}  
&\textbf{0.7246} &\textbf{0.8001} &\textbf{0.6190} &\textbf{0.1084}   \\  \midrule

\multicolumn{1}{c|}{\checkmark} &
\multicolumn{1}{c|}{} &
\multicolumn{1}{c|}{\checkmark} &
\multicolumn{1}{c|}{} &
\multicolumn{1}{c|}{\checkmark} &
\multicolumn{1}{c|}{\checkmark}    
&0.7059 &0.7920 &0.5996 &0.1259   \\ 

\bottomrule

\end{tabular}
\label{ablationTab}
\vspace{-2mm}
\end{table}

\begin{figure}[t]
		\scriptsize
		\renewcommand{\tabcolsep}{0.4pt} 
		\renewcommand{\arraystretch}{0.1} 
		\centering
		\begin{tabular}{ccccccccc}
		    \makecell[c]{\includegraphics[width=0.14\linewidth,height=0.1\linewidth]{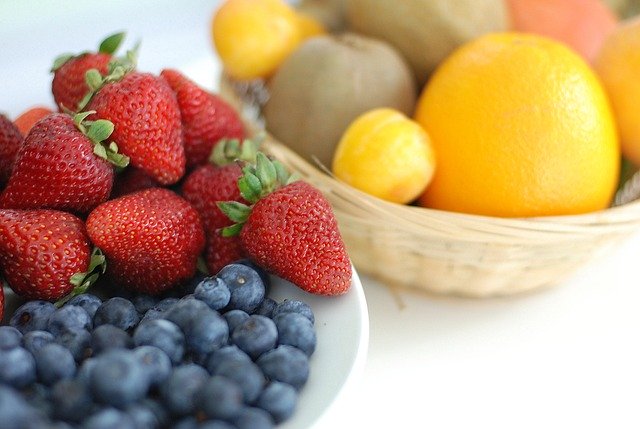}} &
			\makecell[c]{\includegraphics[width=0.14\linewidth,height=0.1\linewidth]{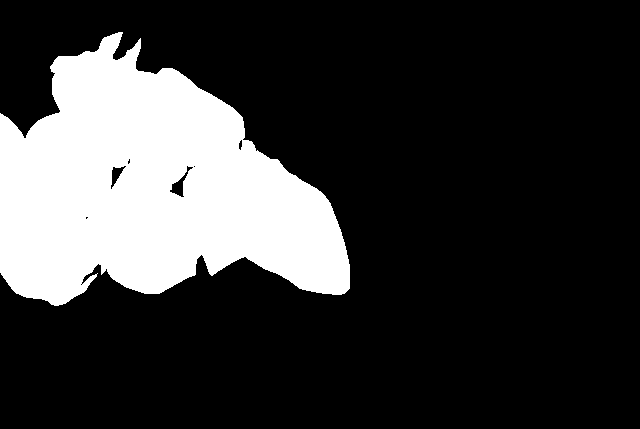}} &
			\makecell[c]{\includegraphics[width=0.14\linewidth,height=0.1\linewidth]{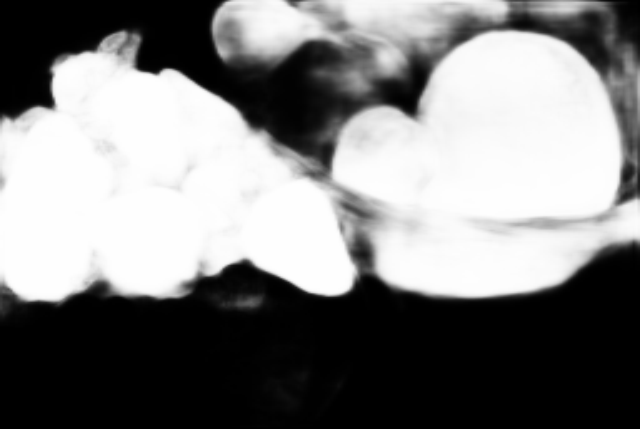}} &
			\makecell[c]{\includegraphics[width=0.14\linewidth,height=0.1\linewidth]{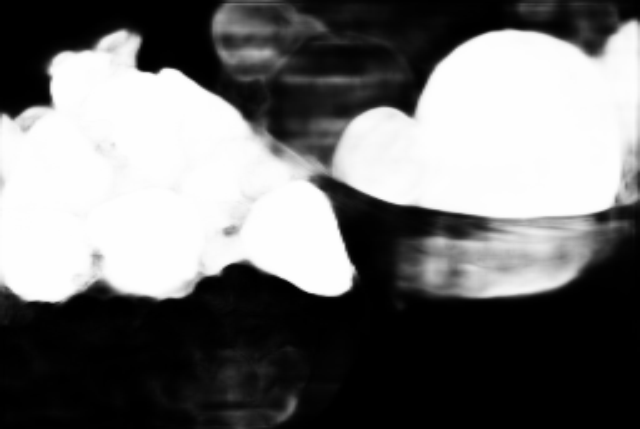}} & 
		    \makecell[c]{\includegraphics[width=0.14\linewidth,height=0.1\linewidth]{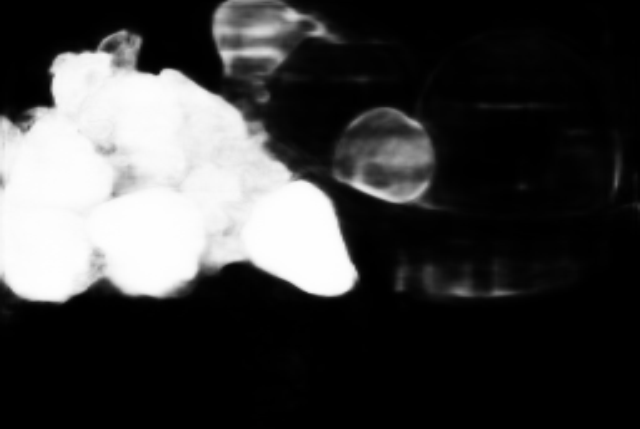}} & 
			\makecell[c]{\includegraphics[width=0.14\linewidth,height=0.1\linewidth]{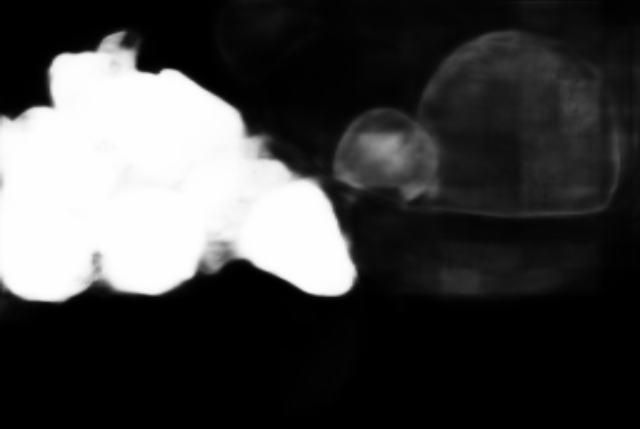}} & 
			\makecell[c]{\includegraphics[width=0.14\linewidth,height=0.1\linewidth]{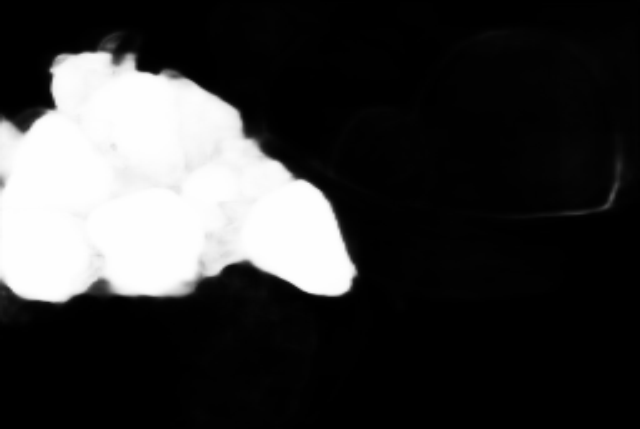}} 
                \vspace{-0.5mm}
		    \\
		    \makecell[c]{\includegraphics[width=0.14\linewidth,height=0.1\linewidth]{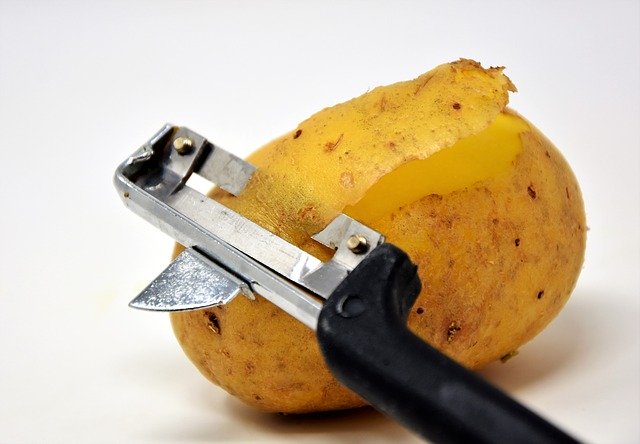}} &
			\makecell[c]{\includegraphics[width=0.14\linewidth,height=0.1\linewidth]{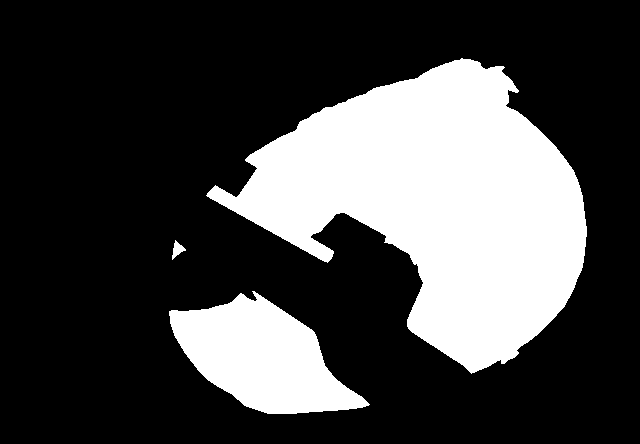}} &
			\makecell[c]{\includegraphics[width=0.14\linewidth,height=0.1\linewidth]{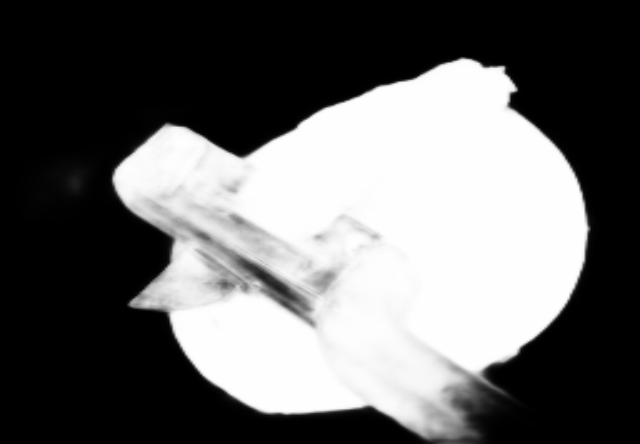}} &
			\makecell[c]{\includegraphics[width=0.14\linewidth,height=0.1\linewidth]{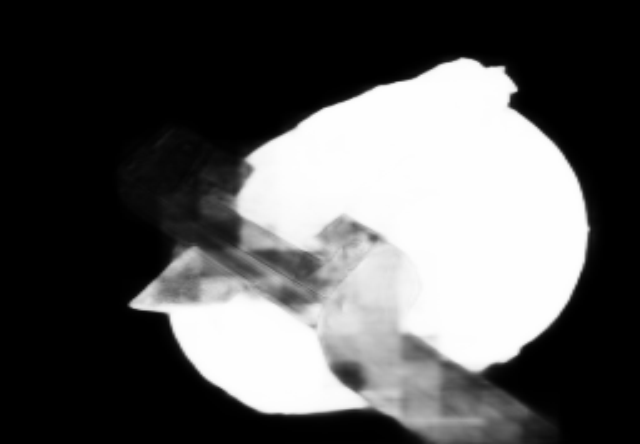}} & 
		    \makecell[c]{\includegraphics[width=0.14\linewidth,height=0.1\linewidth]{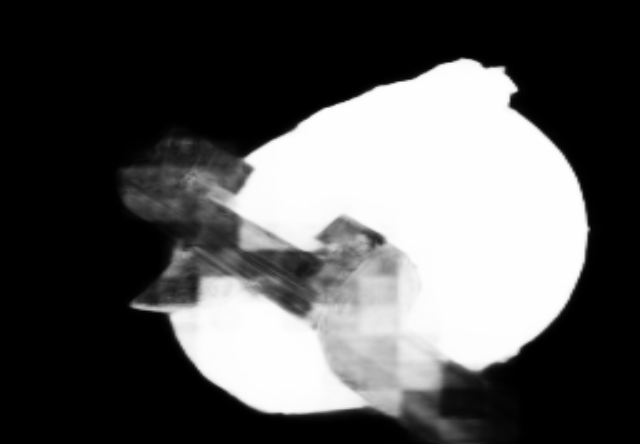}} & 
			\makecell[c]{\includegraphics[width=0.14\linewidth,height=0.1\linewidth]{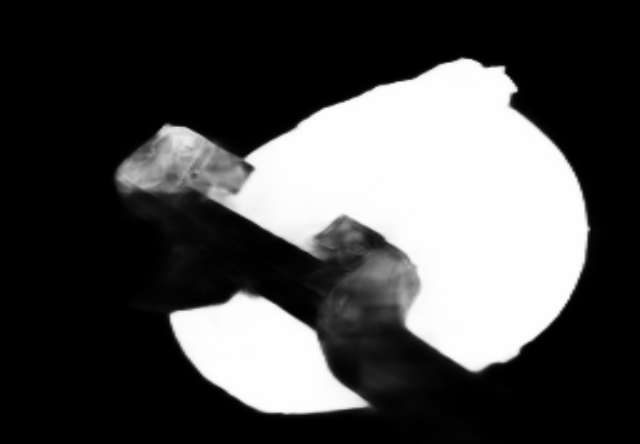}} & 
			\makecell[c]{\includegraphics[width=0.14\linewidth,height=0.1\linewidth]{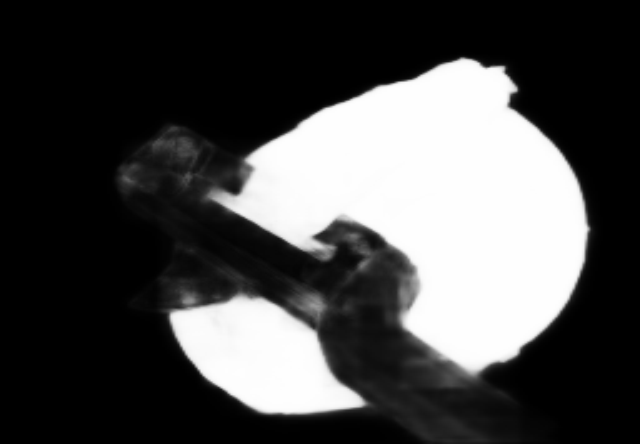}} 
                \vspace{-0.5mm}
                \\
		    \makecell[c]{\includegraphics[width=0.14\linewidth,height=0.1\linewidth]{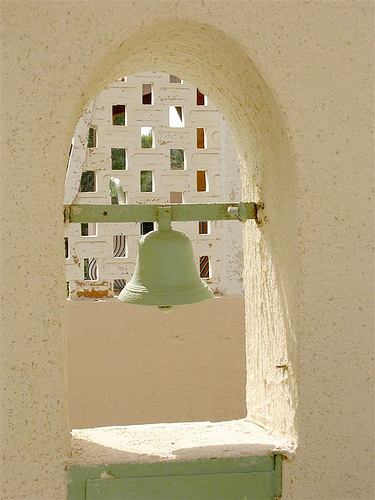}} &
			\makecell[c]{\includegraphics[width=0.14\linewidth,height=0.1\linewidth]{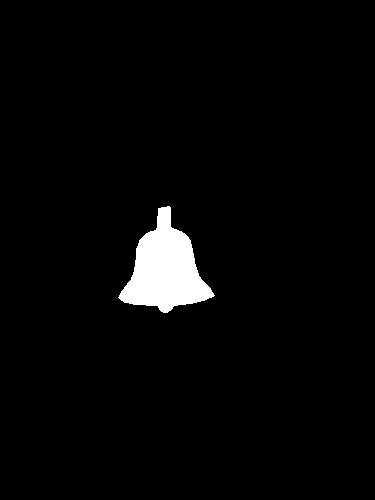}} &
			\makecell[c]{\includegraphics[width=0.14\linewidth,height=0.1\linewidth]{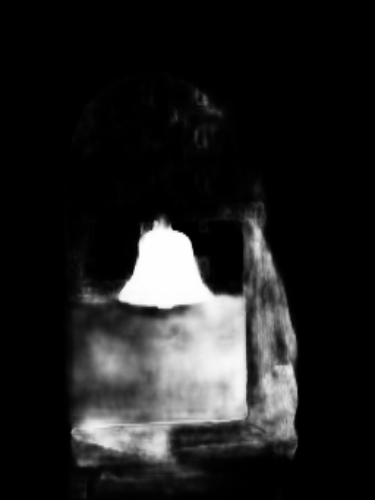}} &
			\makecell[c]{\includegraphics[width=0.14\linewidth,height=0.1\linewidth]{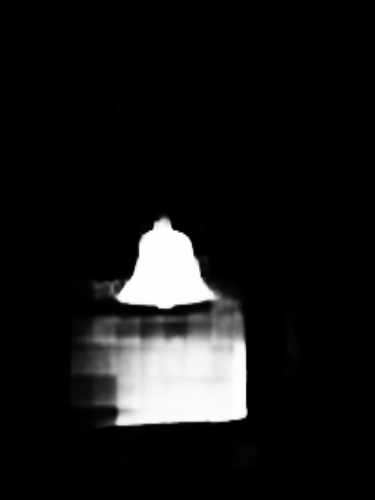}} & 
		    \makecell[c]{\includegraphics[width=0.14\linewidth,height=0.1\linewidth]{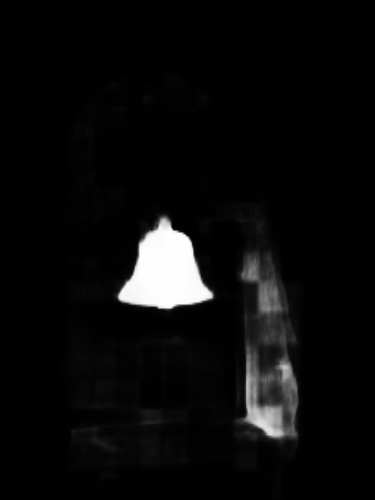}} & 
			\makecell[c]{\includegraphics[width=0.14\linewidth,height=0.1\linewidth]{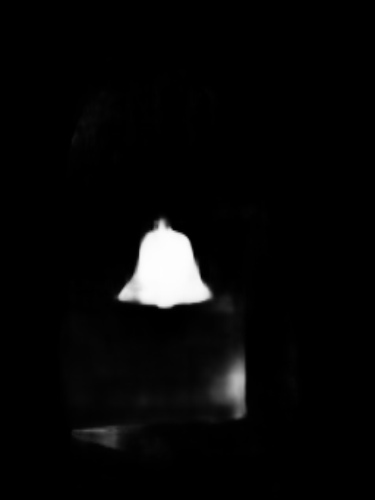}} & 
			\makecell[c]{\includegraphics[width=0.14\linewidth,height=0.1\linewidth]{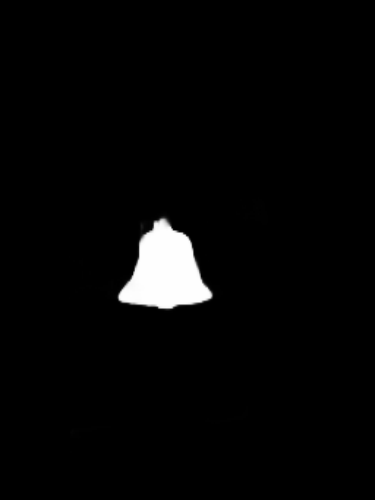}} 
            \\
		  Image &
		  GT &
		  Baseline &
		  +R2R & 
		  +CtP2T &
            +CoT2T &
            +TGFR
		    \\
		\end{tabular}
		\vspace{-0.3cm}
		\caption{
			\textbf{Qualitative results of different settings of our proposed model.} We show the results of progressively adding the R2R, CtP2T, CoT2T, and TGFR on the baseline.
		}
		\label{ablationFig}
		\vspace{-5mm}
\end{figure}

\section{Experiments}
\subsection{Evaluation Datasets and Metrics}
\vspace{-1mm}
We follow \cite{fan2021GCoNet, su2022unified, yu2022democracy} to evaluate our proposed model on three CoSOD benchmark datasets.
CoSal2015 \cite{zhang2015co} and CoSOD3k \cite{fan2020taking} collect 50 groups with 2015 images and 160 groups with 3316 images, respectively. 
CoCA \cite{zhang2020gicd} is the most challenging dataset 
and contain 1295 images of 80 groups. We employ four widely-used metrics for quantitative evaluation, \ie Structure-measure $S_m$ \cite{fan2017structure}, Enhanced-alignment measure $E_\xi$ \cite{Fan2018Enhanced}, Maximum F-measure (maxF) \cite{5206596}, and Mean Absolute Error (MAE) \cite{6751300}.

\subsection{Implementation Details}
\vspace{-1mm}
We follow \cite{zhang2021summarize} to use the COCO-9k\cite{lin2014microsoft} (9213 images of 65 groups) and the DUTS class \cite{zhang2020gicd} (8250 images of 291 groups) with the synthesis strategy \cite{zhang2021summarize} to construct our training set. We follow \cite{liu2018picanet} to perform data augmentation and adopt the Adam optimizer \cite{kingma2014adam} with an initial learning rate of 0.0001, ${\beta}_1=0.9$, and ${\beta}_2=0.99$ to train our model for 80,000 iterations. The learning rate is divided by 10 at the $60000^{\text{th}}$ iteration. 
We select at most eight images from each group as a mini-batch to train our network. The training and testing image size is set as $256 \times 256$.
Our method is implemented using Pytorch \cite{paszke2019pytorch}.

\begin{figure}[t]
		\scriptsize
		\renewcommand{\tabcolsep}{0.4pt} 
		\renewcommand{\arraystretch}{0.1} 
		\centering
		\begin{tabular}{ccccccccc}
		    \makecell[c]{\includegraphics[width=0.14\linewidth,height=0.1\linewidth]{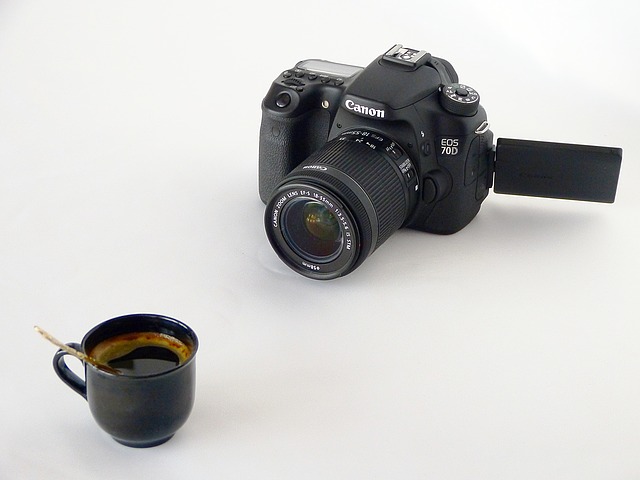}} &
			\makecell[c]{\includegraphics[width=0.14\linewidth,height=0.1\linewidth]{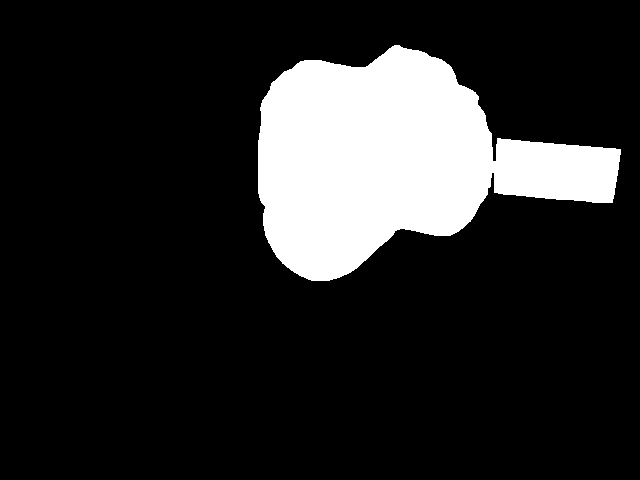}} & 
            \hspace{0.001mm}
			\makecell[c]{\includegraphics[width=0.14\linewidth,height=0.1\linewidth]{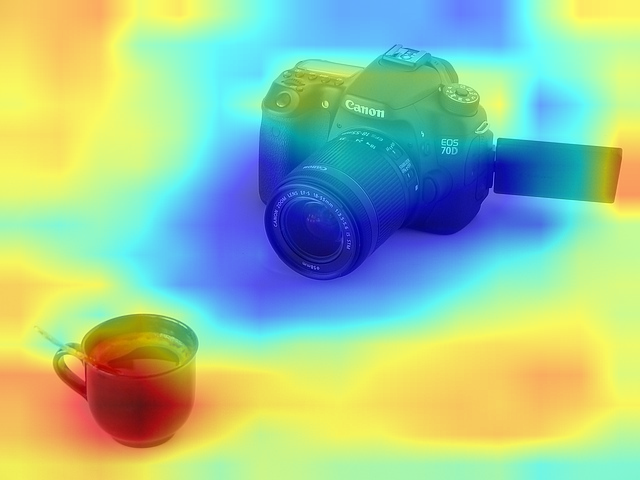}} & 
		    \makecell[c]{\includegraphics[width=0.14\linewidth,height=0.1\linewidth]{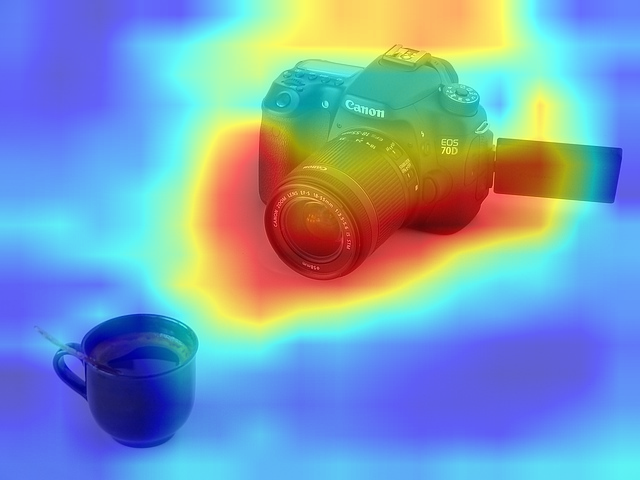}} & \hspace{0.001mm}
			\makecell[c]{\includegraphics[width=0.14\linewidth,height=0.1\linewidth]{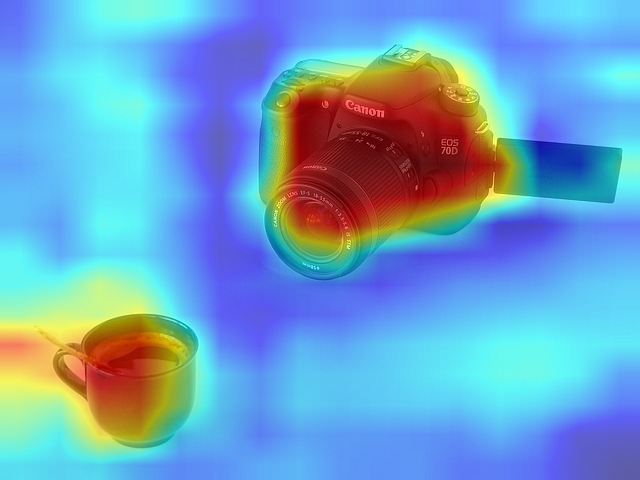}} & 
			\makecell[c]{\includegraphics[width=0.14\linewidth,height=0.1\linewidth]{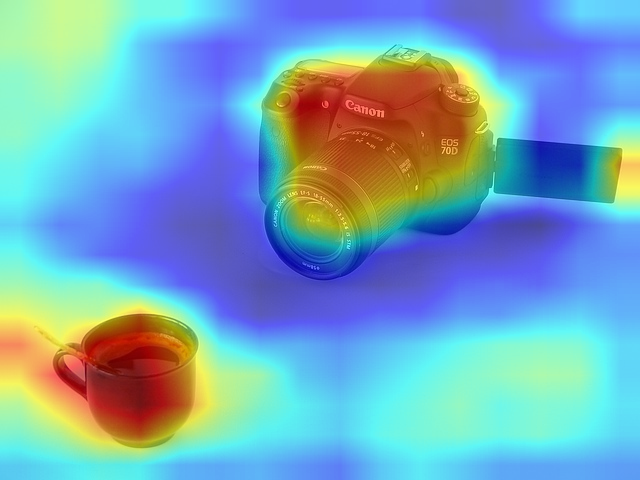}} & 
            \hspace{0.001mm}
			\makecell[c]{\includegraphics[width=0.14\linewidth,height=0.1\linewidth]{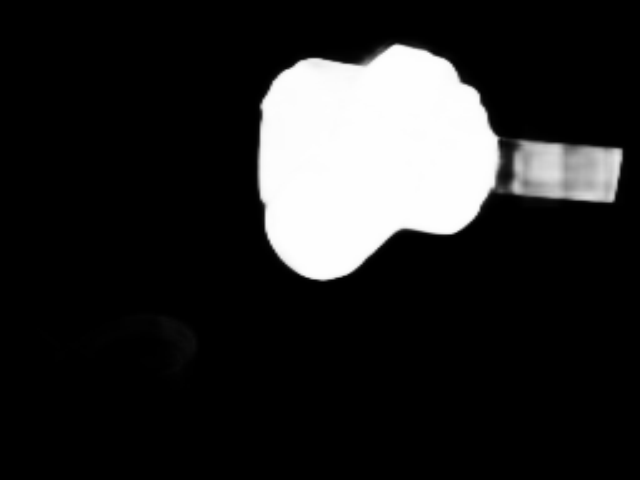}} 
                \vspace{-0.5mm}
		    \\
		    \makecell[c]{\includegraphics[width=0.14\linewidth,height=0.1\linewidth]{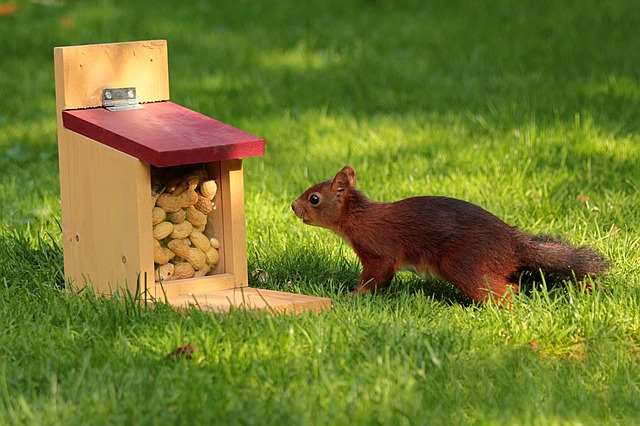}} &
			\makecell[c]{\includegraphics[width=0.14\linewidth,height=0.1\linewidth]{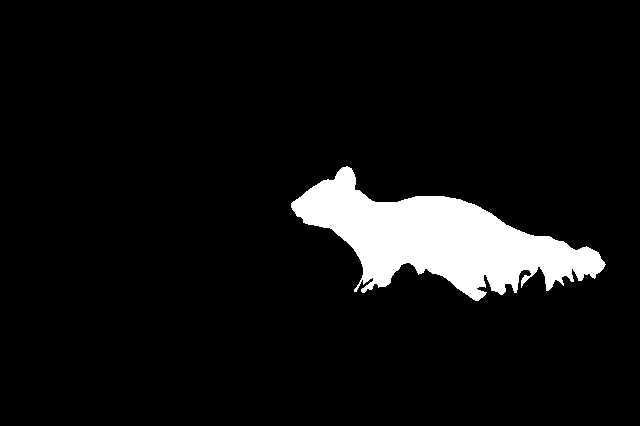}} &
            \hspace{0.001mm}
			\makecell[c]{\includegraphics[width=0.14\linewidth,height=0.1\linewidth]{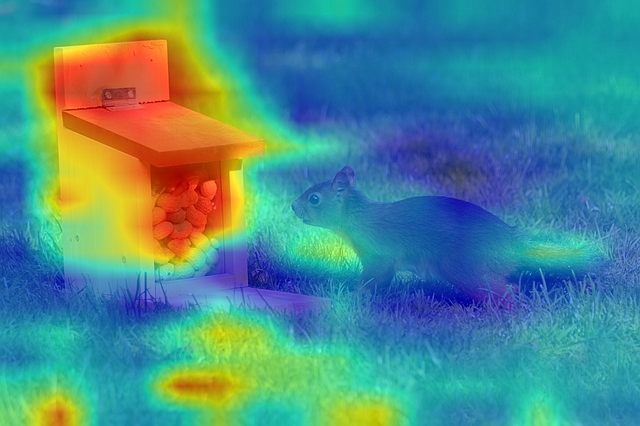}} &
			\makecell[c]{\includegraphics[width=0.14\linewidth,height=0.1\linewidth]{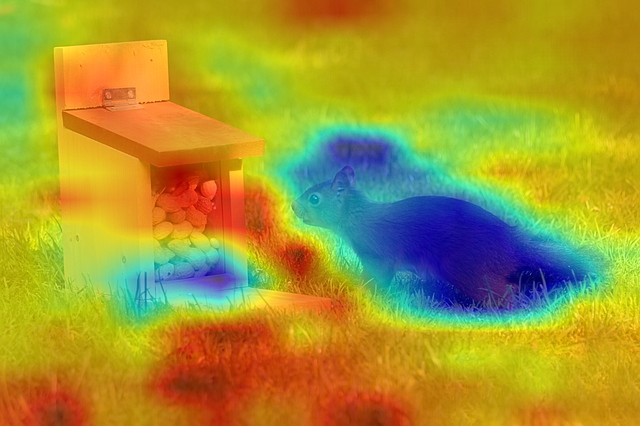}} & 
            \hspace{0.001mm}
		    \makecell[c]{\includegraphics[width=0.14\linewidth,height=0.1\linewidth]{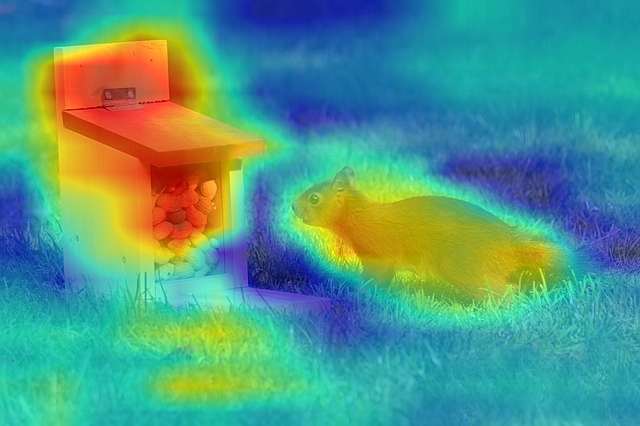}} & 
			\makecell[c]{\includegraphics[width=0.14\linewidth,height=0.1\linewidth]{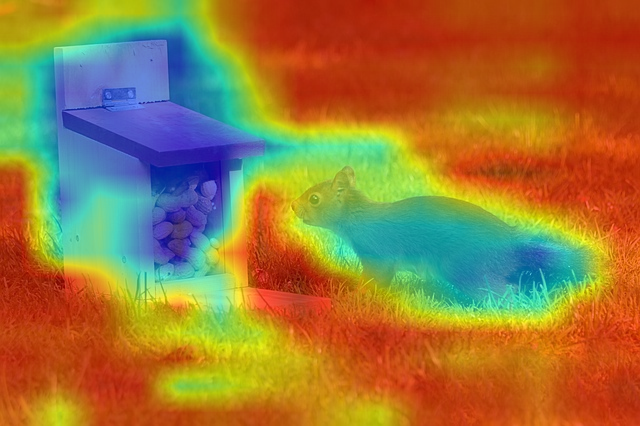}} & 
            \hspace{0.001mm}
			\makecell[c]{\includegraphics[width=0.14\linewidth,height=0.1\linewidth]{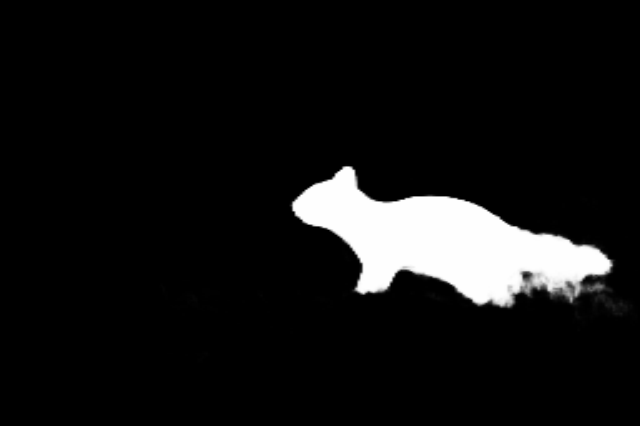}} 
                \vspace{-0.5mm}
                \\
            \makecell[c]{\includegraphics[width=0.14\linewidth,height=0.1\linewidth]{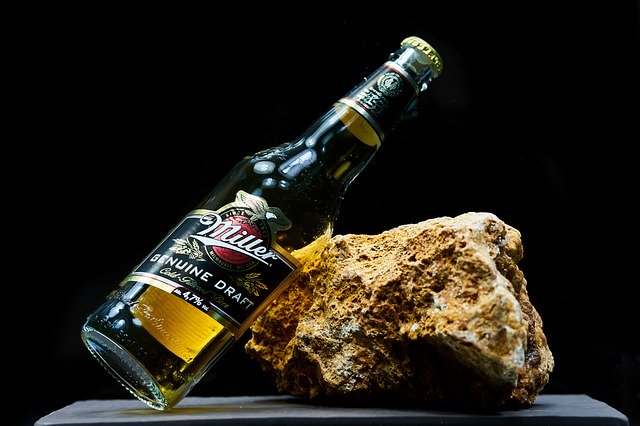}} &
			\makecell[c]{\includegraphics[width=0.14\linewidth,height=0.1\linewidth]{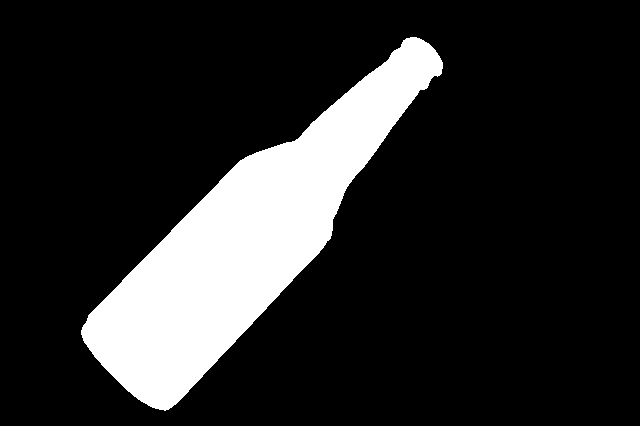}} &
            \hspace{0.001mm}
			\makecell[c]{\includegraphics[width=0.14\linewidth,height=0.1\linewidth]{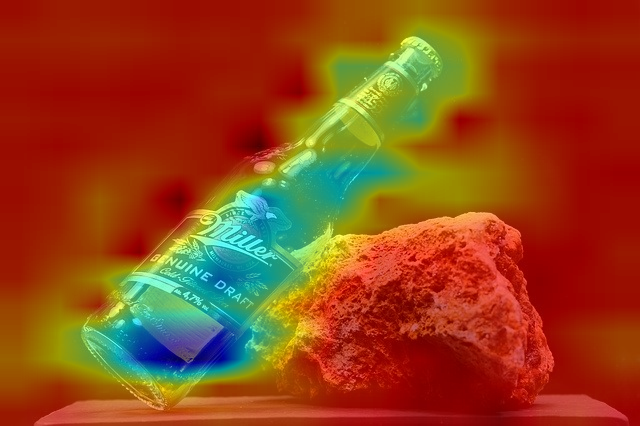}} &
			\makecell[c]{\includegraphics[width=0.14\linewidth,height=0.1\linewidth]{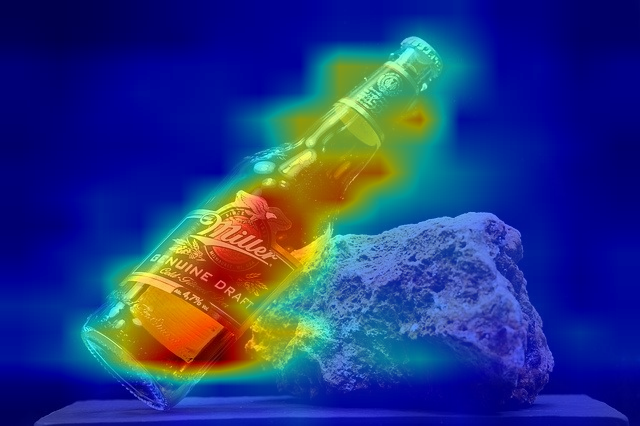}} &
            \hspace{0.001mm}
		    \makecell[c]{\includegraphics[width=0.14\linewidth,height=0.1\linewidth]{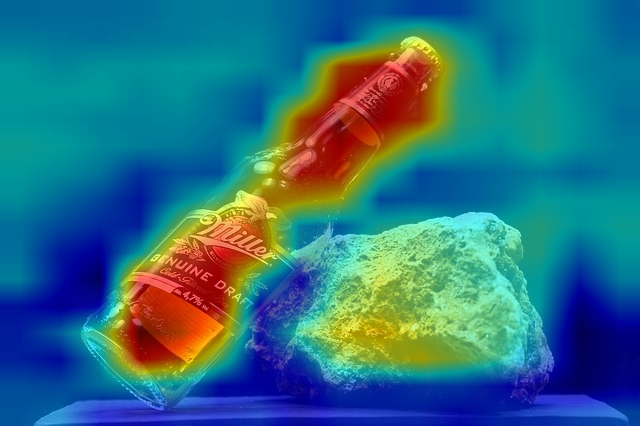}} & 
			\makecell[c]{\includegraphics[width=0.14\linewidth,height=0.1\linewidth]{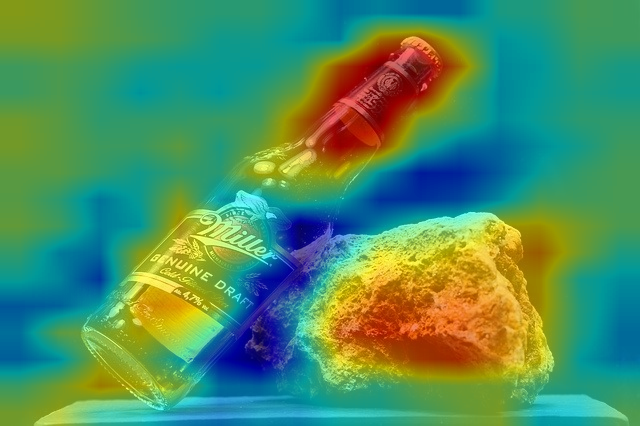}} & 
            \hspace{0.001mm}
			\makecell[c]{\includegraphics[width=0.14\linewidth,height=0.1\linewidth]{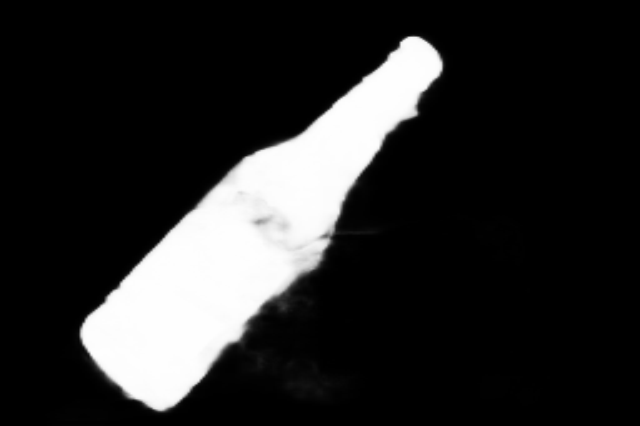}} 
            \\
		  Image &
		  GT &
		  \multicolumn{2}{c}{Large CA} &
            \multicolumn{2}{c}{Small CA} &
            Prediction
		    \\
		\end{tabular}
		\vspace{-0.3cm}
		\caption{
			\textbf{Visual comparison among the channels with different channel attention weights in CtP2T.} We visualize some feature maps in $V_m^*(\boldsymbol{F})$ for the channels with large and small channel attention (CA) in CtP2T. We visualize two channels for large and small CA, respectively.
		}
		\label{VisualCtP2T}
		\vspace{-5mm}
\end{figure}

\subsection{Ablation Study}
\vspace{-1mm}
We conduct ablation studies on the challenging CoCA \cite{zhang2020gicd} dataset to verify the effectiveness of our proposed components. 
As shown in Table~\ref{ablationTab},
we treat the vanilla MaskFormer-style framework
as our baseline, shown in the first row, and progressively add our proposed R2R, CtP2T, CoT2T, and TGFR on it for effectiveness analysis.

\Paragraph{Effectiveness of R2R.} First, we plug R2R into each decoder layer to enhance the segmentation features. It shows that using R2R largely improves the model performance compared to the baseline, while using vanilla P2P causes the out-of-memory error. The results verify the necessity of using our R2R for inter-image correlation modeling.

\Paragraph{Effectiveness of CtP2T.} Next, we consider the contrast relation modeling between the co-saliency and BG tokens, thus replacing the original P2T module to our proposed CtP2T module. By using CtP2T, the model performance can be further improved, indicating that CtP2T is beneficial for enhancing the discriminability between the two types of tokens. We also provide some visual samples in Figure~\ref{VisualCtP2T}. Since the channels of the tokens correspond to those of the values in $\operatorname{MHA}^*$, we visualize some feature maps of $V^*_m(\boldsymbol{F})$ of the channels with large or small channel attention weights in $\boldsymbol{W}$. We can see that the channels with large channel attention (CA) can easily distinguish co-salient objects and distracting objects, while those with small CA usually confuse them. The results demonstrate our generated channel attention is meaningful
for accurate co-salient object detection.

\Paragraph{Effectiveness of CoT2T.} Furthermore, we supplement CoT2T to explore the inter-image correlations for all co-saliency tokens. CoT2T explicitly promotes consensus information propagation among all co-saliency tokens, thus obtaining obvious improvements.

\begin{table}[t]
\centering
\footnotesize
\renewcommand{\arraystretch}{1}
\renewcommand{\tabcolsep}{3.5mm}
\caption{\textbf{Quantitative results of different settings in TGFR.}
}
\vspace{-3mm}
\begin{tabular}{l|l|cccc}
\toprule

\multicolumn{2}{c|}{\multirow{2}{*}{Settings}} & \multicolumn{4}{c}{CoCA \cite{zhang2020gicd}}  \\

\multicolumn{2}{l|}{} & 
\multicolumn{1}{l}{$S_m \uparrow$} &  
\multicolumn{1}{l}{$E_\xi\uparrow$} &
\multicolumn{1}{l}{maxF$\uparrow$} & 
\multicolumn{1}{l}{MAE $\downarrow$}\\ \hline

\multicolumn{2}{l|}{w/o TGFR}
&0.7140 &0.7880 &0.6003 &0.1139 \\ \hline
\multicolumn{2}{l|}{w/o Distillation}
&0.7141 &0.7921 &0.6046 &0.1114 \\ \hline
\multicolumn{2}{l|}{w/ co}
&0.7171 &0.7965 &0.6076 &0.1144 \\ 
\multicolumn{2}{l|}{w/ bg}     
&0.7155 &0.7935 &0.6064 &0.1112 \\ 
\multicolumn{2}{l|}{w/ bg\&co}    
&0.7197 &0.7907 &0.6092 &0.1069 \\ 
\rowcolor[RGB]{214,220,229}
\multicolumn{2}{l|}{w/ co\&bg}   
&\textbf{0.7246} &\textbf{0.8001} &\textbf{0.6190} &\textbf{0.1084} \\ 
\bottomrule

\end{tabular}
\label{tab: detail_TGFR}
\vspace{-3mm}
\end{table}

\begin{figure}[t]
		\scriptsize
		\renewcommand{\tabcolsep}{0.7pt} 
		\renewcommand{\arraystretch}{0.1} 
		\centering
		\begin{tabular}{cccccccc}
		    \makecell[c]{\includegraphics[width=0.16\linewidth,height=0.12\linewidth]{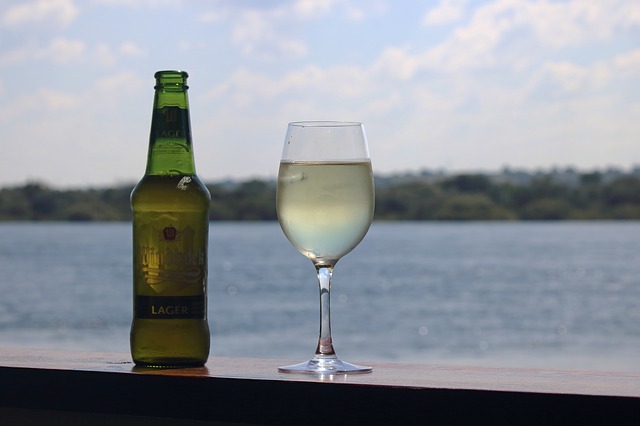}} &
			\makecell[c]{\includegraphics[width=0.16\linewidth,height=0.12\linewidth]{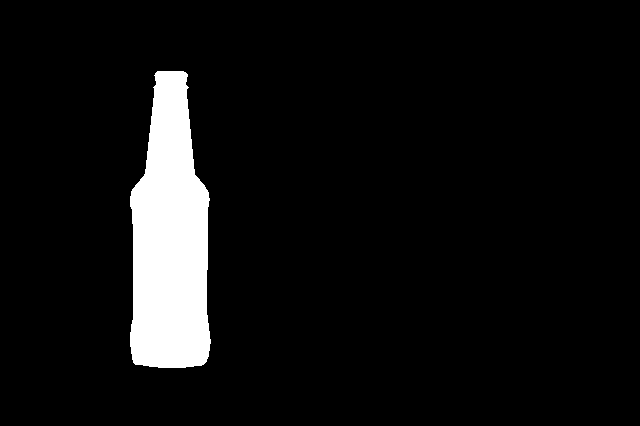}} &
			\makecell[c]{\includegraphics[width=0.16\linewidth,height=0.12\linewidth]{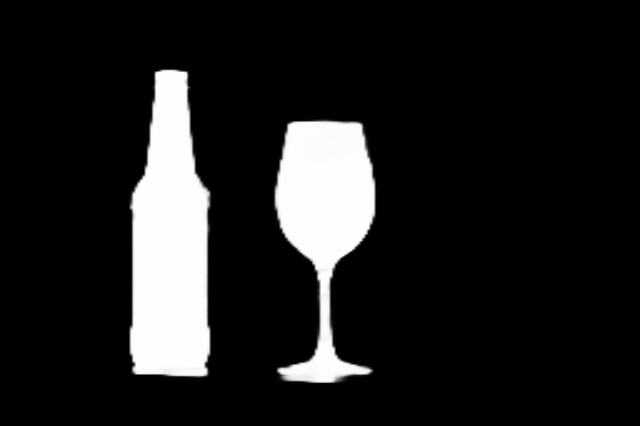}} &
			\makecell[c]{\includegraphics[width=0.16\linewidth,height=0.12\linewidth]{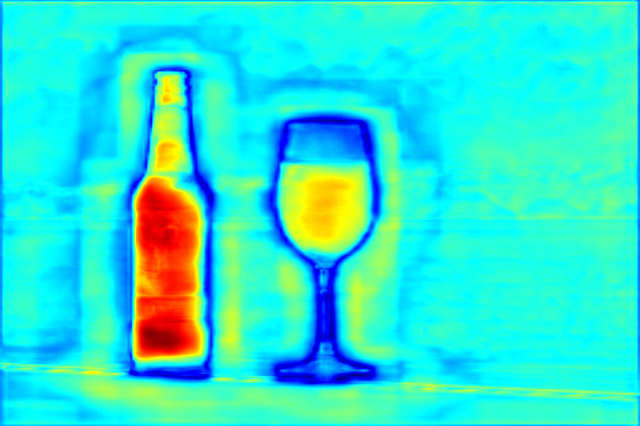}} & 
		    \makecell[c]{\includegraphics[width=0.16\linewidth,height=0.12\linewidth]{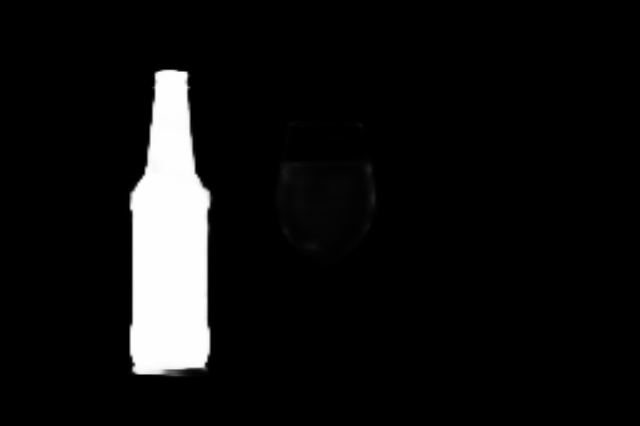}} & 
			\makecell[c]{\includegraphics[width=0.16\linewidth,height=0.12\linewidth]{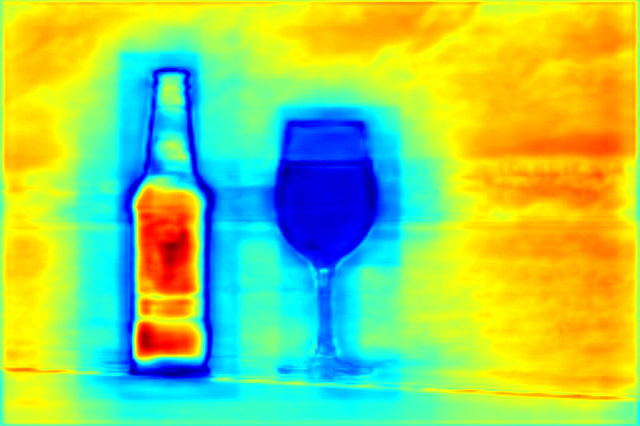}} & 
                \vspace{-0.3mm}
		    \\
		    \makecell[c]{\includegraphics[width=0.16\linewidth,height=0.12\linewidth]{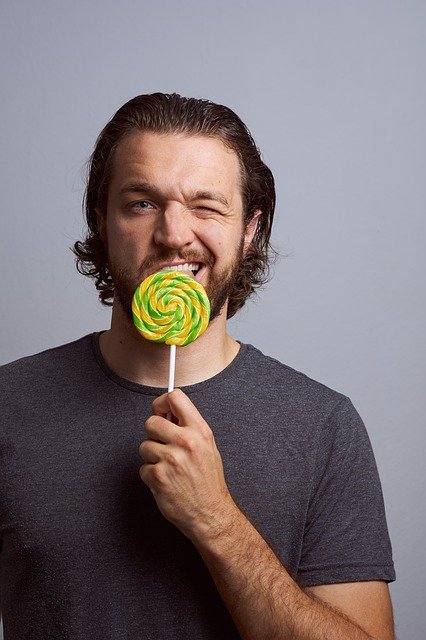}} &
			\makecell[c]{\includegraphics[width=0.16\linewidth,height=0.12\linewidth]{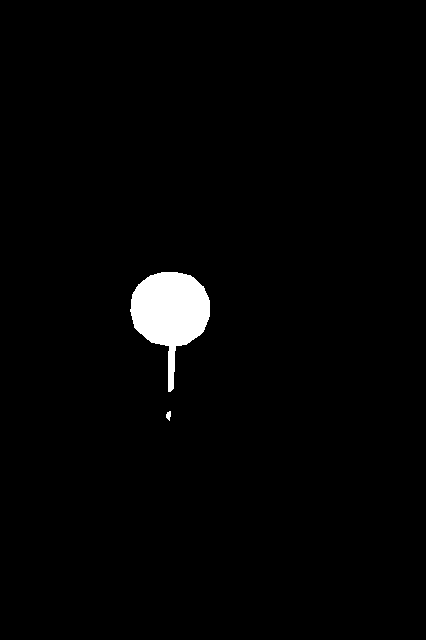}} &
			\makecell[c]{\includegraphics[width=0.16\linewidth,height=0.12\linewidth]{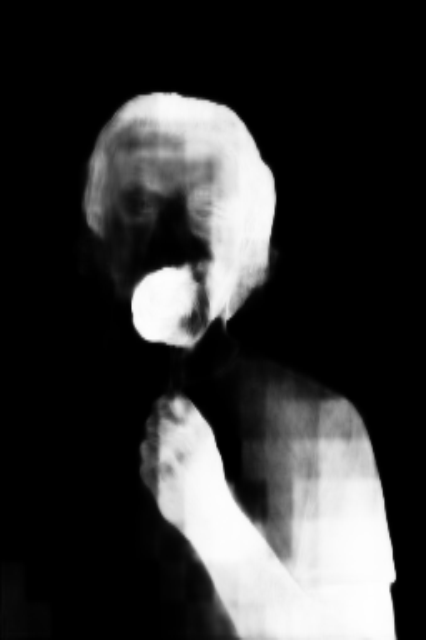}} &
			\makecell[c]{\includegraphics[width=0.16\linewidth,height=0.12\linewidth]{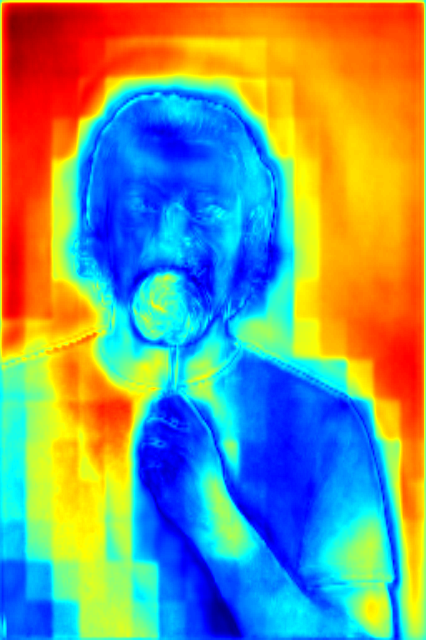}} & 
		    \makecell[c]{\includegraphics[width=0.16\linewidth,height=0.12\linewidth]{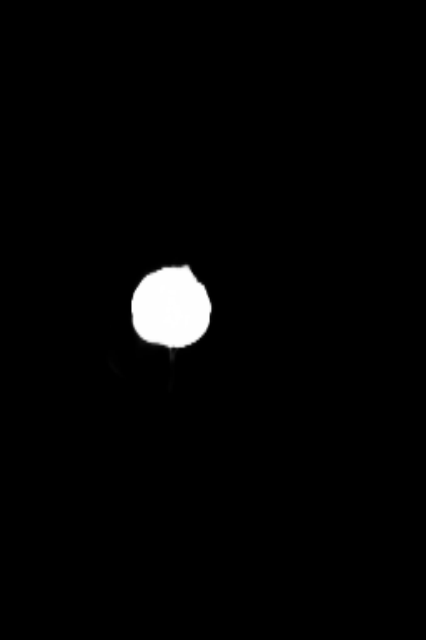}} & 
			\makecell[c]{\includegraphics[width=0.16\linewidth,height=0.12\linewidth]{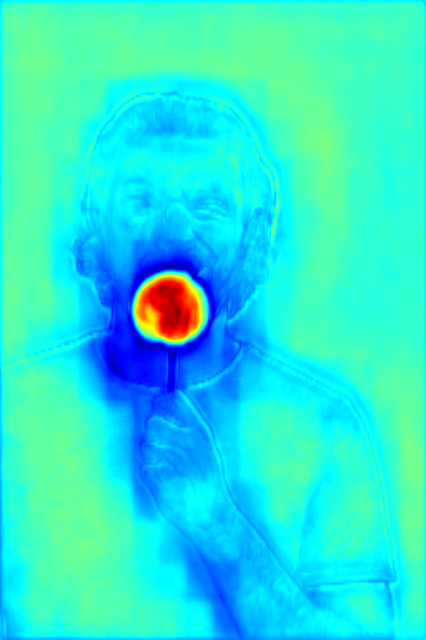}} 
                \vspace{-0.3mm}
                \\
		  Image &
		  GT &
		  Pred. w/o &
		  Fea. w/o & 
		  Pred. w/ &
            Fea. w/ 
		    \\
		\end{tabular}
		\vspace{-0.3cm}
		\caption{
		        \textbf{Visualization of some feature maps (Fea.) and predictions (Pred.) of the models with (w/) or without (w/o) using TGFR.}
		}
		\label{fig:visual_TGFR}
		\vspace{-5mm}
\end{figure}

\begin{table*}[t]
  \centering
  \footnotesize
  \renewcommand{\arraystretch}{0.8}
  \renewcommand{\tabcolsep}{1.7mm}
 \caption{\textbf{Quantitative comparison of our model with other state-of-the-art methods.} We conduct the comparison on three benchmark CoSOD datasets. \red{Red} and \blu{blue} denote the best and the second-best results, respectively.}
  \vspace{-0.3cm}
  \begin{tabular}{lr|cccc|cccc|cccc}
  \toprule
  
   \multicolumn{2}{c|}{\multirow{2}{*}{Methods}} &
   \multicolumn{4}{c|}{CoCA\cite{zhang2020gicd}} &
   \multicolumn{4}{c|}{CoSal2015\cite{zhang2015co}} &
   \multicolumn{4}{c}{CoSOD3k\cite{fan2020taking}} 
   \\
   \cmidrule{3-14}

   &
   &
   $S_m\uparrow$ &
   $E_\xi\uparrow$ &
   maxF$\uparrow$ &
   MAE$\downarrow$ &
   
   $S_m\uparrow$ &
   $E_\xi\uparrow$ &
   maxF$\uparrow$ &
   MAE$\downarrow$ &   
   
   $S_m\uparrow$ &
   $E_\xi\uparrow$ &
   maxF$\uparrow$ &
   MAE$\downarrow$ 

   \\
   
   \midrule

   \multicolumn{2}{c|}{$\text{CSMG}_{(\text{CVPR2019})}$ \cite{zhang2019co}} & 
   0.6276 & 0.7324 & 0.4988 & 0.1273 &
   0.7757 & 0.8436 & 0.7869 & 0.1309 &
   0.7272 & 0.8208 & 0.7297 & 0.1480 
   \\

   \multicolumn{2}{c|}{$\text{GICD}_{(\text{ECCV2020})}$ \cite{zhang2020gicd}} &
   0.6579 & 0.7149 & 0.5126 & 0.1260 &
   0.8437 & 0.8869 & 0.8441 & 0.0707 &
   0.7967 & 0.8478 & 0.7698 & 0.0794 
   \\

   \multicolumn{2}{c|}{$\text{ICNet}_{(\text{NIPS2020})}$ \cite{jin2020icnet}} &
   0.6541 & 0.7042 & 0.5133 & 0.1470 &
   0.8571 & 0.9011 & 0.8583 & 0.0579 &
   0.7942 & 0.8450 & 0.7623 & 0.0891 
   \\

   \multicolumn{2}{c|}{$\text{GCoNet}_{(\text{CVPR2021})}$ \cite{fan2021GCoNet}} &
   0.6730 & 0.7598 & 0.5438 & 0.1050 &
   0.8453 & 0.8879 & 0.8471 & 0.0681 &
   0.8018 & 0.8601 & 0.7771 & 0.0712 
   \\   

   \multicolumn{2}{c|}{$\text{CADC}_{(\text{ICCV2021})}$ \cite{zhang2021summarize}} &
   0.6800 & 0.7443 & 0.5487 & 0.1330 &
   \blu{0.8666} & \blu{0.9063} & \blu{0.8645} & \blu{0.0641} &
   0.8150 & 0.8543 & 0.7781 & 0.0875 
   \\   

   \multicolumn{2}{c|}{$\text{UFO}_{(\text{ArXiv2022})}$ \cite{su2022unified}} &
   0.6971 & 0.7802 & 0.5681 & \blu{0.0939} &
   0.8578 & 0.9057 & 0.8621 & 0.0648 &
   \blu{0.8191} & 0.8694 & 0.7954 & 0.0735 
   \\
   
   \multicolumn{2}{c|}{$\text{DCFM}_{(\text{CVPR2022})}$ \cite{yu2022democracy}} &
   \blu{0.7101} & \blu{0.7826} & \blu{0.5981} & \red{0.0845} &
   0.8380 & 0.8929 & 0.8559 & 0.0672 &
   0.8094 & \blu{0.8742} & \blu{0.8045} & \blu{0.0674}
   \\
   \midrule

   \rowcolor[RGB]{214,220,229}
   \multicolumn{2}{c|}{DMT (Ours)} &
   \red{0.7246} & \red{0.8001} & \red{0.6190} & 0.1084 &
   \red{0.8974} & \red{0.9362}	& \red{0.9052} & \red{0.0454} &
   \red{0.8514} & \red{0.8950} & \red{0.8353} & \red{0.0633} 
   \\
   \bottomrule
   
  \end{tabular}
  \label{SOTATab}
\end{table*}

\begin{figure*}[t]
	\scriptsize
	\renewcommand{\tabcolsep}{0.5pt} 
	\renewcommand{\arraystretch}{0.9} 
	\centering
	\begin{tabular}{ccccccccccccccccc}
	    &
            \multicolumn{4}{c}{\cellcolor[RGB]{255,230,153}$\bold{Helicopter}$}
            &
            \multicolumn{4}{c}{
            \cellcolor[RGB]{244,177,131}$\bold{Perfume}$}
	    &
            \multicolumn{4}{c}{
            \cellcolor[RGB]{169,209,142}$\bold{Coffeecup}$}
            &
            \multicolumn{4}{c}{
            \cellcolor[RGB]{143,170,220}$\bold{Scorpion}$}
            \vspace{0.5mm}
	    \\
            \arrayrulecolor{red}
            \rotatebox[origin=c]{90}{Images}
            &
		\makecell[c]{\includegraphics[width=0.06\linewidth,height=0.05\linewidth]{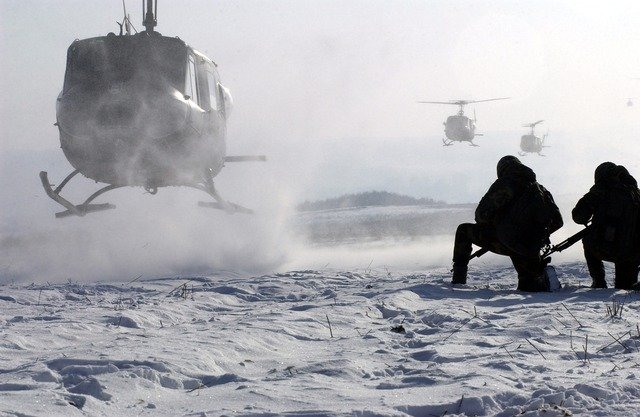}} 
		&
		\makecell[c]{\includegraphics[width=0.06\linewidth,height=0.05\linewidth]{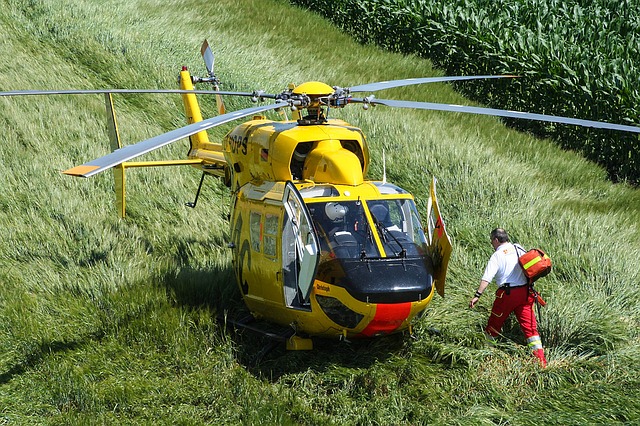}} 
		&
		\makecell[c]{\includegraphics[width=0.06\linewidth,height=0.05\linewidth]{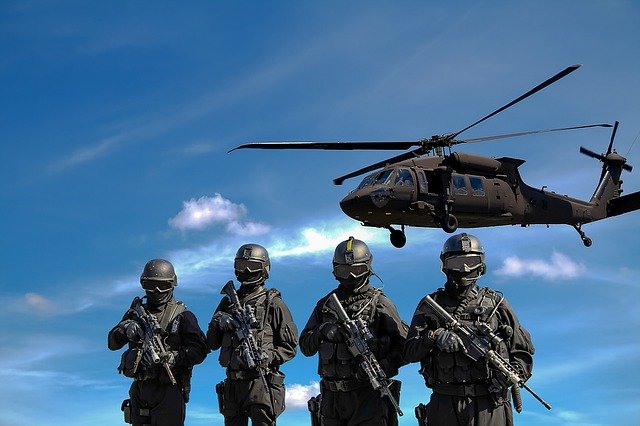}} 
		&
		\makecell[c]{\includegraphics[width=0.06\linewidth,height=0.05\linewidth]{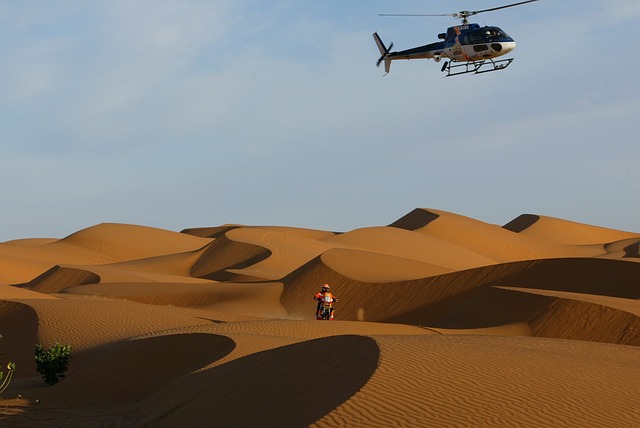}} 
		&
		\makecell[c]{\includegraphics[width=0.06\linewidth,height=0.05\linewidth]{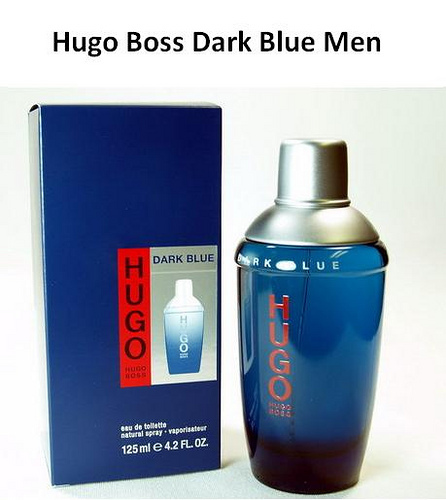}} 
		&
		\makecell[c]{\includegraphics[width=0.06\linewidth,height=0.05\linewidth]{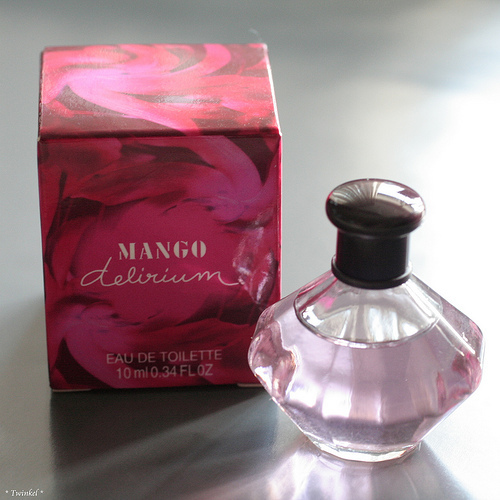}} 
		&
		\makecell[c]{\includegraphics[width=0.06\linewidth,height=0.05\linewidth]{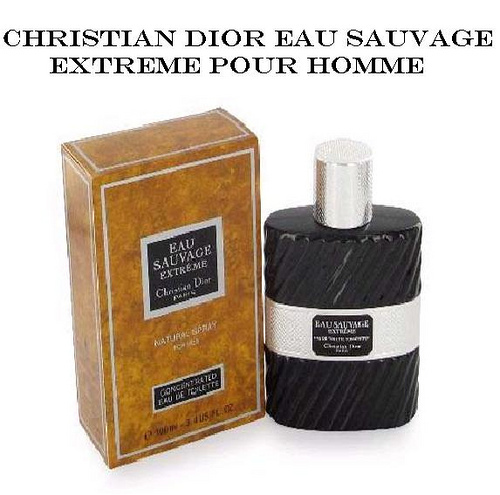}} 
		&
		\makecell[c]{\includegraphics[width=0.06\linewidth,height=0.05\linewidth]{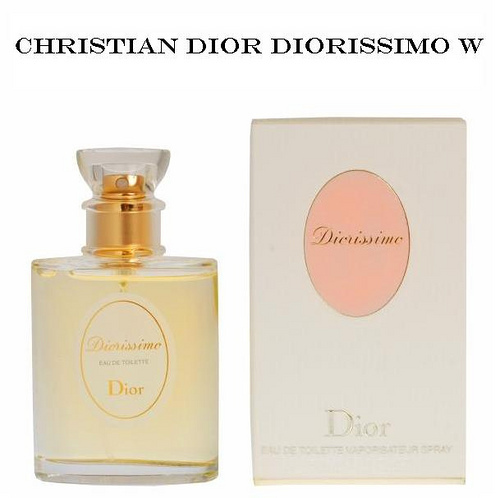}}
            &
		\makecell[c]{\includegraphics[width=0.06\linewidth,height=0.05\linewidth]{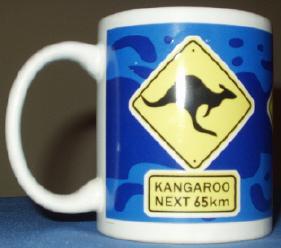}} 
		&
		\makecell[c]{\includegraphics[width=0.06\linewidth,height=0.05\linewidth]{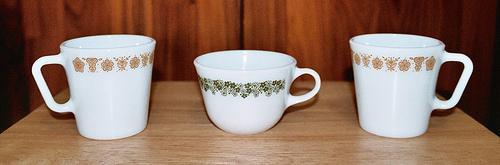}} 
		&
		\makecell[c]{\includegraphics[width=0.06\linewidth,height=0.05\linewidth]{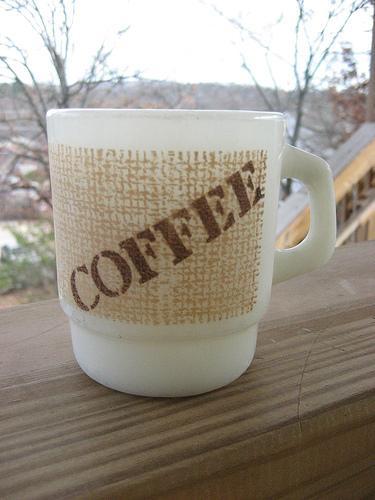}}
		&
		\makecell[c]{\includegraphics[width=0.06\linewidth,height=0.05\linewidth]{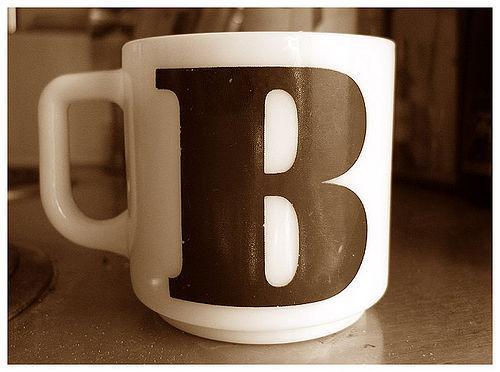}}
		&
		\makecell[c]{\includegraphics[width=0.06\linewidth,height=0.05\linewidth]{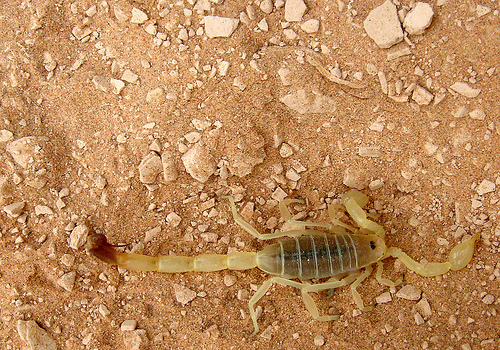}} 
		&
		\makecell[c]{\includegraphics[width=0.06\linewidth,height=0.05\linewidth]{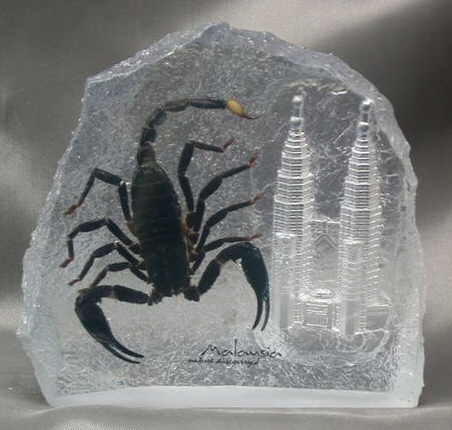}} 
		&
		\makecell[c]{\includegraphics[width=0.06\linewidth,height=0.05\linewidth]{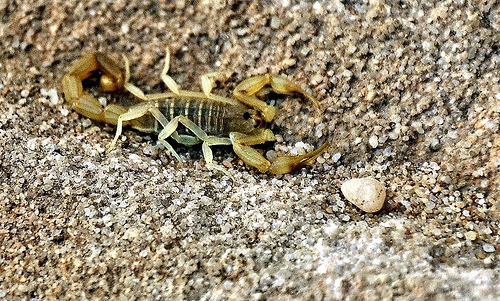}} 
		&
		\makecell[c]{\includegraphics[width=0.06\linewidth,height=0.05\linewidth]{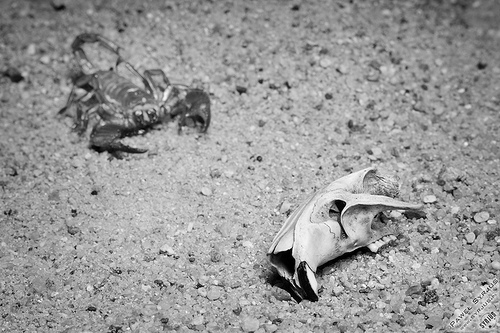}} 
		\vspace{-0.05cm}
		\\
            \rotatebox[origin=c]{90}{GT}
            &
		\makecell[c]{\includegraphics[width=0.06\linewidth,height=0.05\linewidth]{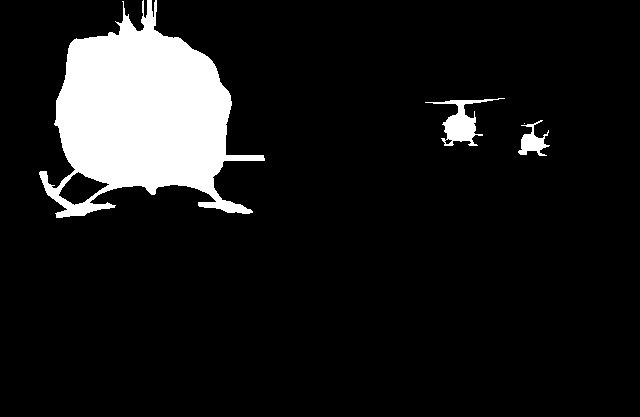}} 
		&
		\makecell[c]{\includegraphics[width=0.06\linewidth,height=0.05\linewidth]{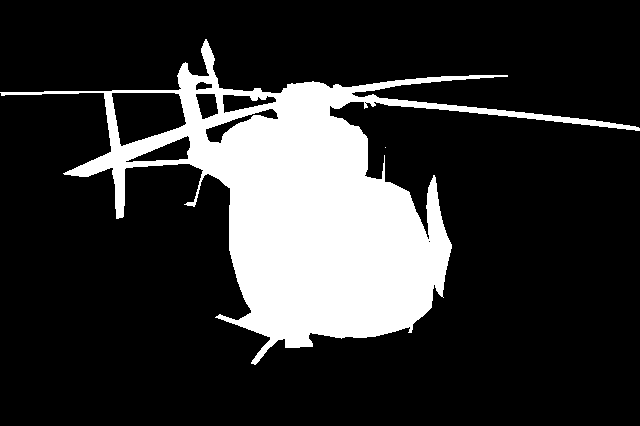}}
		&
		\makecell[c]{\includegraphics[width=0.06\linewidth,height=0.05\linewidth]{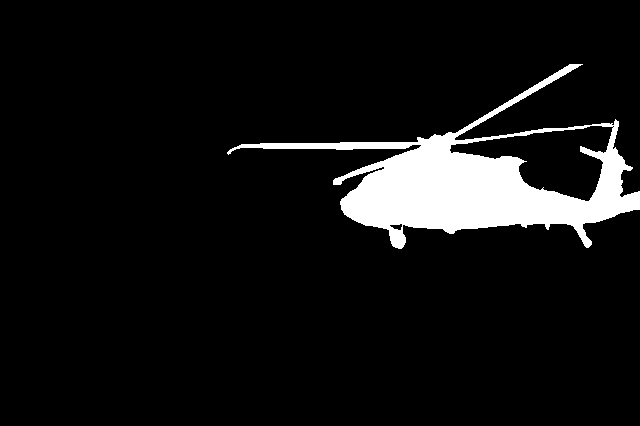}} 
		&
		\makecell[c]{\includegraphics[width=0.06\linewidth,height=0.05\linewidth]{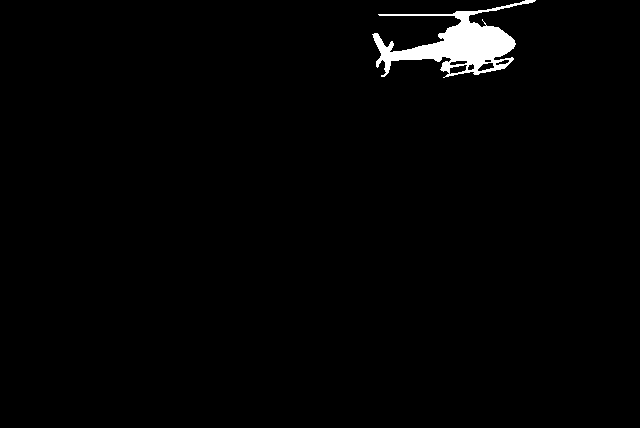}} 
		&
		\makecell[c]{\includegraphics[width=0.06\linewidth,height=0.05\linewidth]{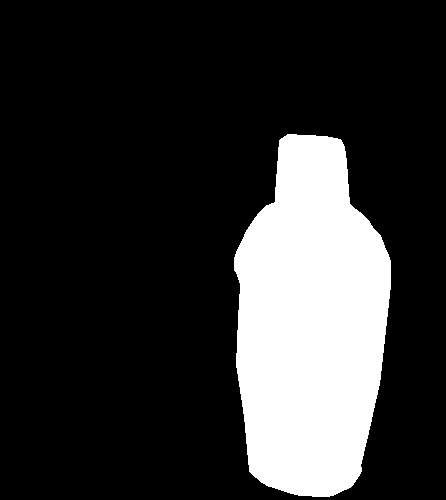}} 
		&
		\makecell[c]{\includegraphics[width=0.06\linewidth,height=0.05\linewidth]{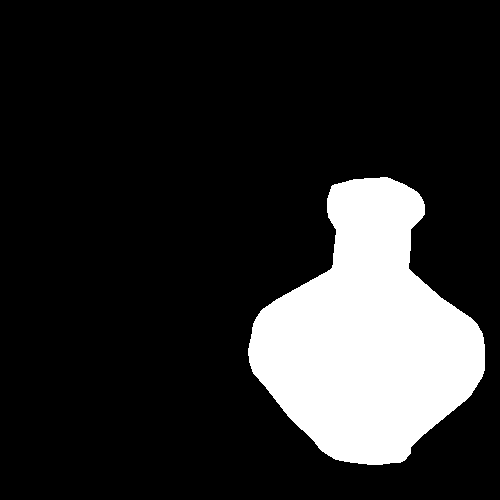}} 
		&
		\makecell[c]{\includegraphics[width=0.06\linewidth,height=0.05\linewidth]{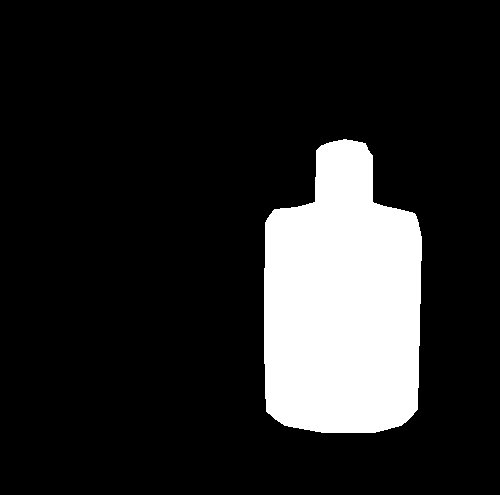}} 
		&
		\makecell[c]{\includegraphics[width=0.06\linewidth,height=0.05\linewidth]{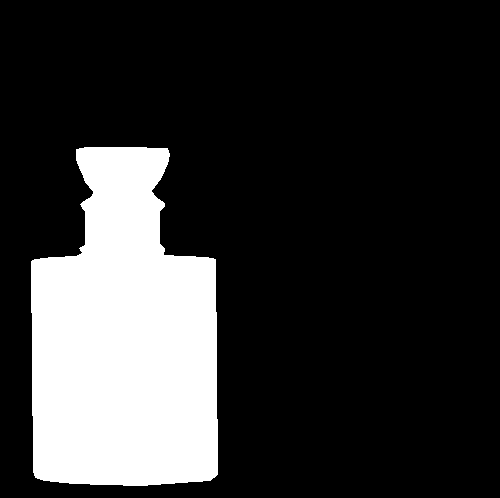}}
            &
		\makecell[c]{\includegraphics[width=0.06\linewidth,height=0.05\linewidth]{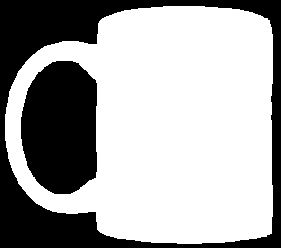}} 
		&
		\makecell[c]{\includegraphics[width=0.06\linewidth,height=0.05\linewidth]{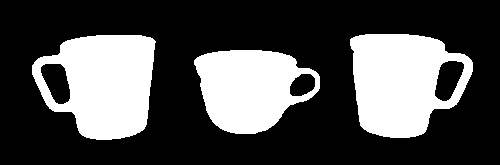}} 
		&
		\makecell[c]{\includegraphics[width=0.06\linewidth,height=0.05\linewidth]{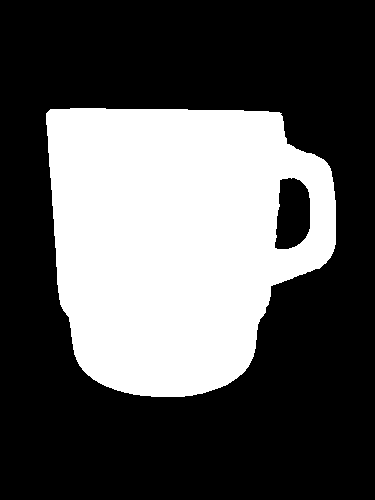}}
		&
		\makecell[c]{\includegraphics[width=0.06\linewidth,height=0.05\linewidth]{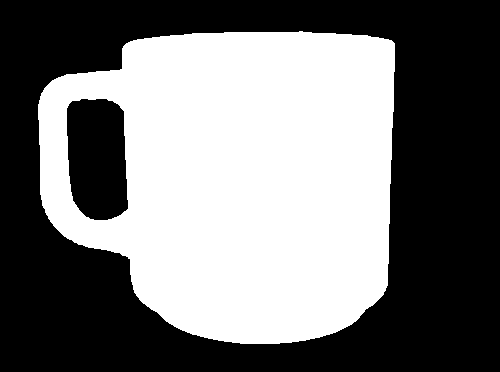}}
		&
		\makecell[c]{\includegraphics[width=0.06\linewidth,height=0.05\linewidth]{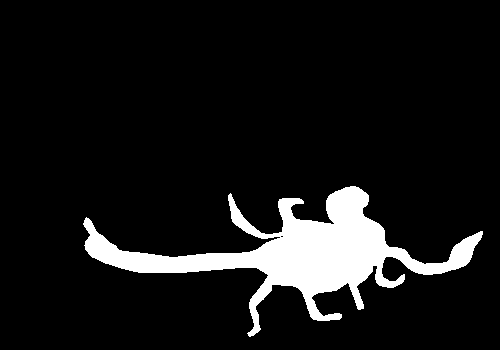}} 
		&
		\makecell[c]{\includegraphics[width=0.06\linewidth,height=0.05\linewidth]{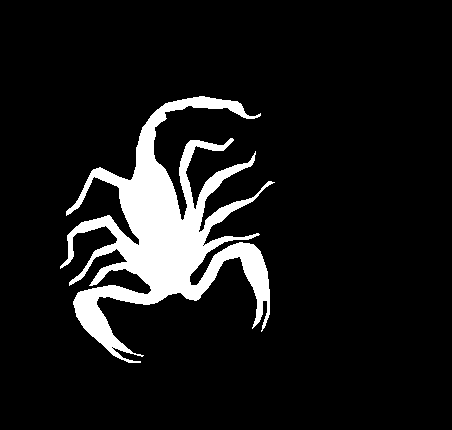}} 
		&
		\makecell[c]{\includegraphics[width=0.06\linewidth,height=0.05\linewidth]{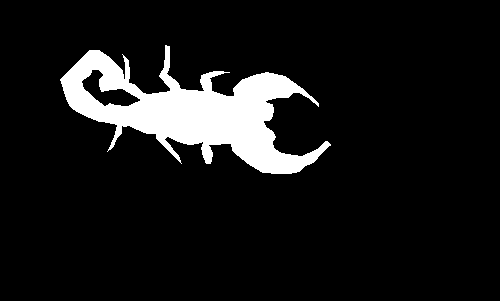}} 	
		&
		\makecell[c]{\includegraphics[width=0.06\linewidth,height=0.05\linewidth]{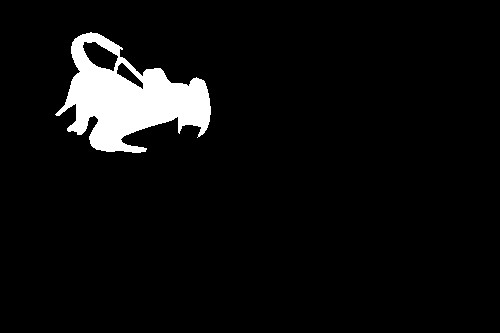}} 
		\vspace{-0.05cm}
            \\
            \arrayrulecolor{red}
            \rotatebox[origin=c]{90}{Ours}
            &
		\makecell[c]{\includegraphics[width=0.06\linewidth,height=0.05\linewidth]{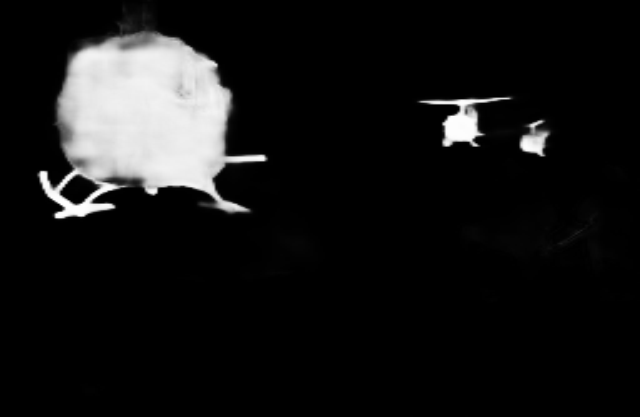}} 
		&
		\makecell[c]{\includegraphics[width=0.06\linewidth,height=0.05\linewidth]{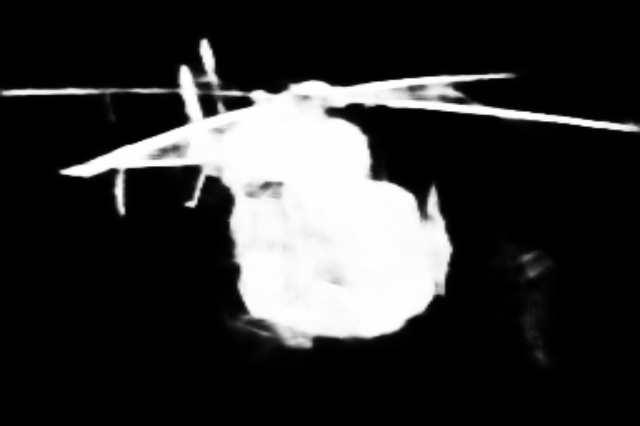}}
		&
		\makecell[c]{\includegraphics[width=0.06\linewidth,height=0.05\linewidth]{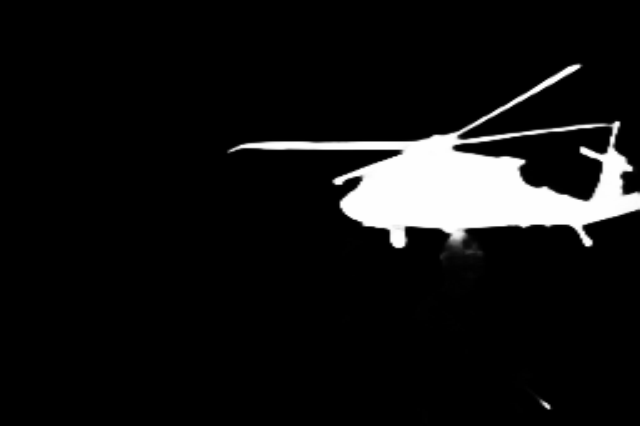}} 
		&
		\makecell[c]{\includegraphics[width=0.06\linewidth,height=0.05\linewidth]{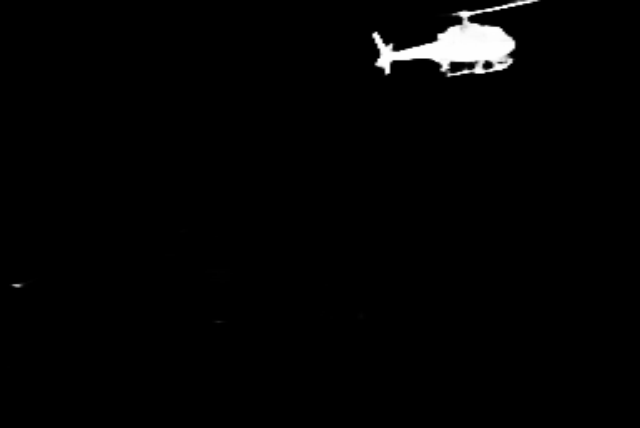}} 
		&
		\makecell[c]{\includegraphics[width=0.06\linewidth,height=0.05\linewidth]{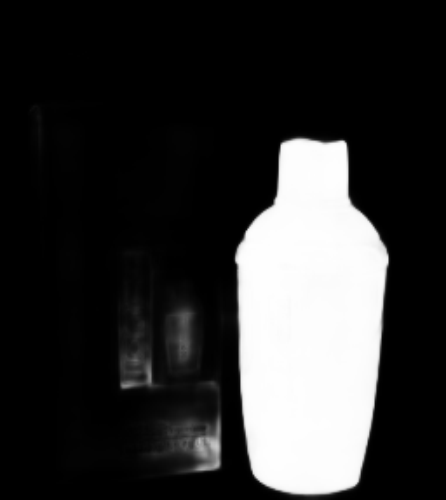}} 
		&
		\makecell[c]{\includegraphics[width=0.06\linewidth,height=0.05\linewidth]{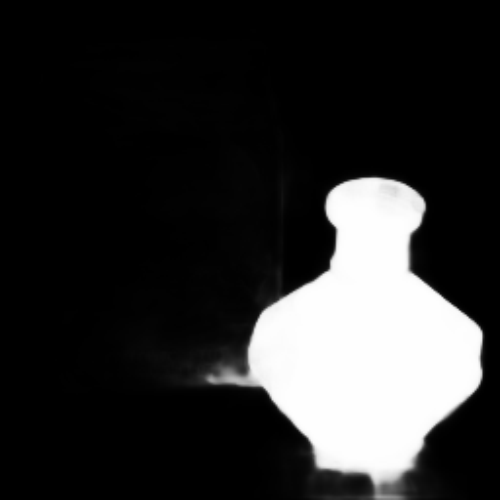}} 
		&
		\makecell[c]{\includegraphics[width=0.06\linewidth,height=0.05\linewidth]{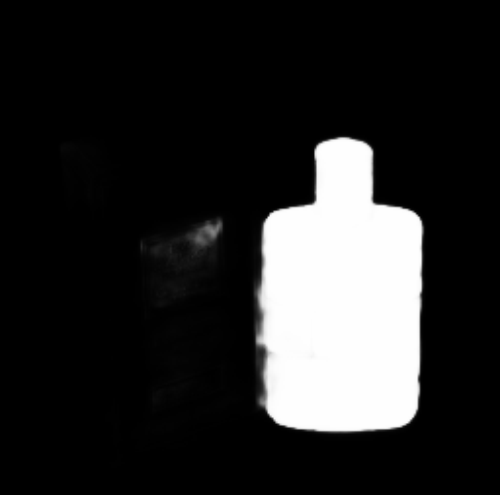}} 
		&
		\makecell[c]{\includegraphics[width=0.06\linewidth,height=0.05\linewidth]{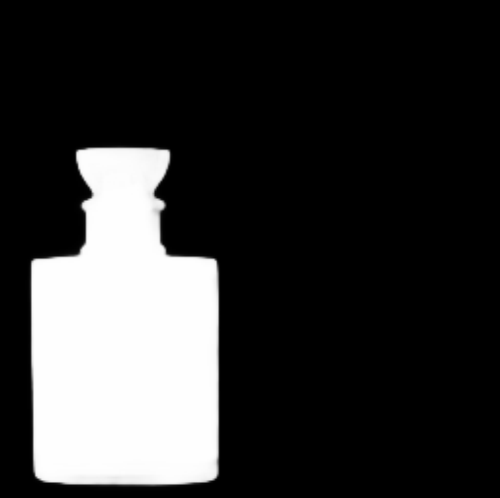}}
            &
		\makecell[c]{\includegraphics[width=0.06\linewidth,height=0.05\linewidth]{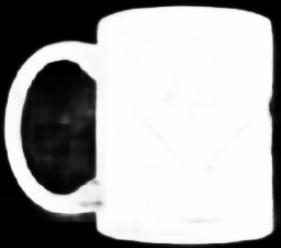}} 
		&
		\makecell[c]{\includegraphics[width=0.06\linewidth,height=0.05\linewidth]{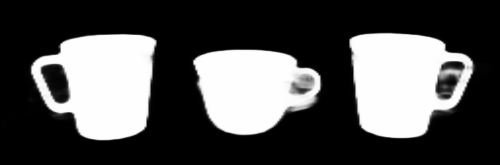}} 
		&
		\makecell[c]{\includegraphics[width=0.06\linewidth,height=0.05\linewidth]{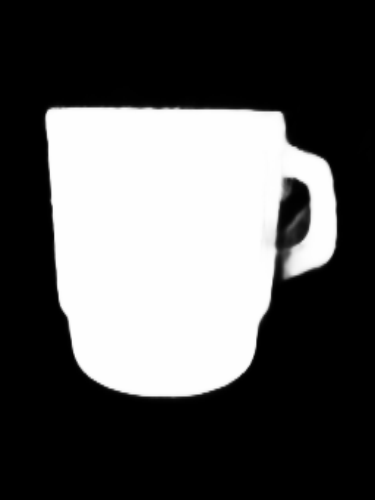}}
		&
		\makecell[c]{\includegraphics[width=0.06\linewidth,height=0.05\linewidth]{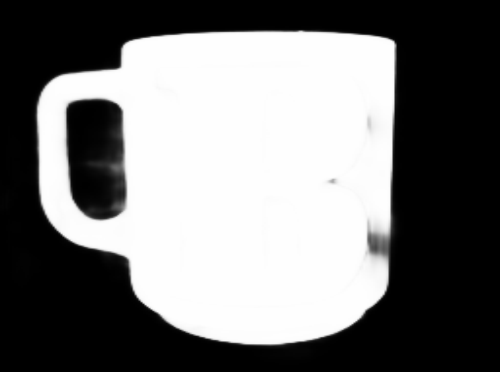}}
		&
		\makecell[c]{\includegraphics[width=0.06\linewidth,height=0.05\linewidth]{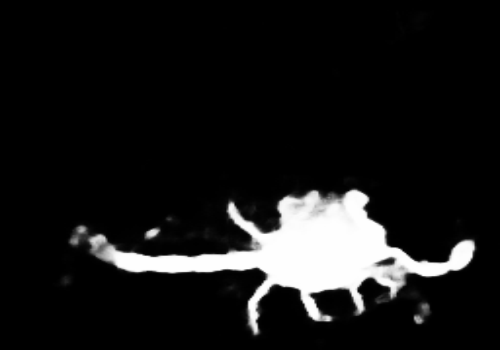}} 
		&
		\makecell[c]{\includegraphics[width=0.06\linewidth,height=0.05\linewidth]{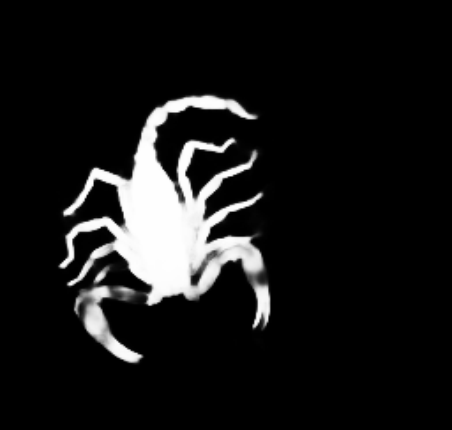}} 
		&
		\makecell[c]{\includegraphics[width=0.06\linewidth,height=0.05\linewidth]{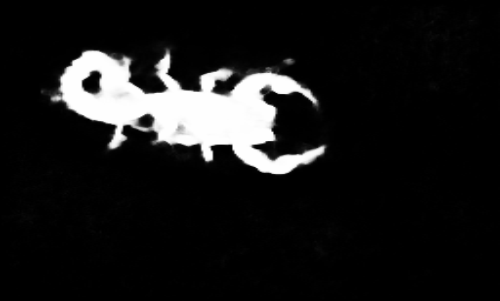}} 	
		&
		\makecell[c]{\includegraphics[width=0.06\linewidth,height=0.05\linewidth]{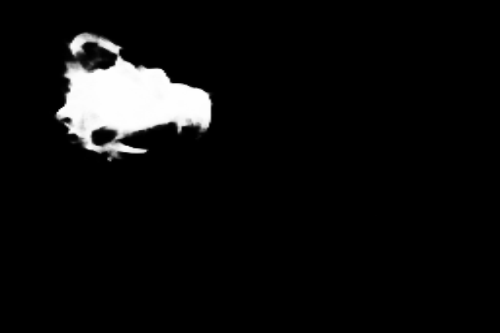}} 
		\vspace{-0.05cm}
            \\
            
            \rotatebox[origin=c]{90}{DCFM}
            &
		\makecell[c]{\includegraphics[width=0.06\linewidth,height=0.05\linewidth]{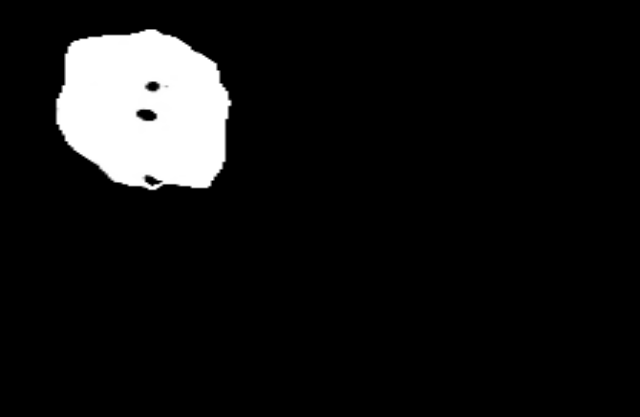}} 
		&
		\makecell[c]{\includegraphics[width=0.06\linewidth,height=0.05\linewidth]{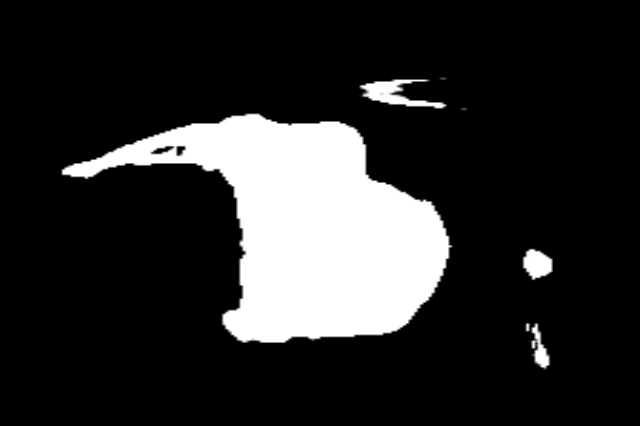}}
		&
		\makecell[c]{\includegraphics[width=0.06\linewidth,height=0.05\linewidth]{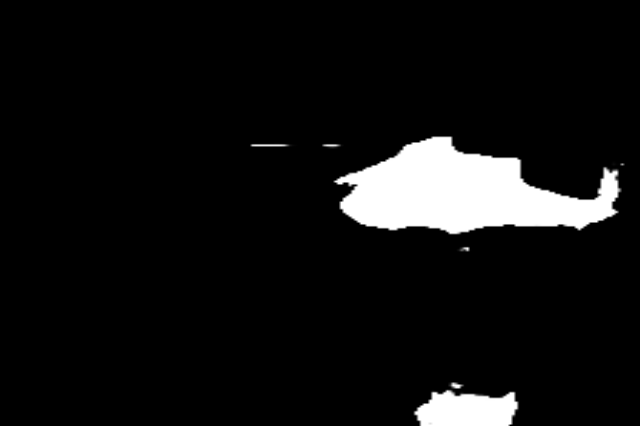}} 
		&
		\makecell[c]{\includegraphics[width=0.06\linewidth,height=0.05\linewidth]{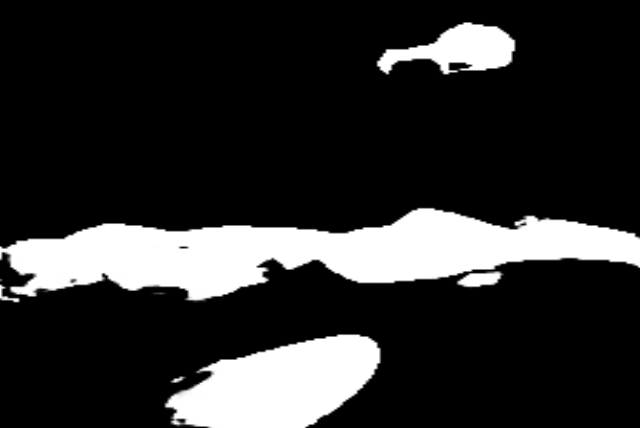}} 
		&
		\makecell[c]{\includegraphics[width=0.06\linewidth,height=0.05\linewidth]{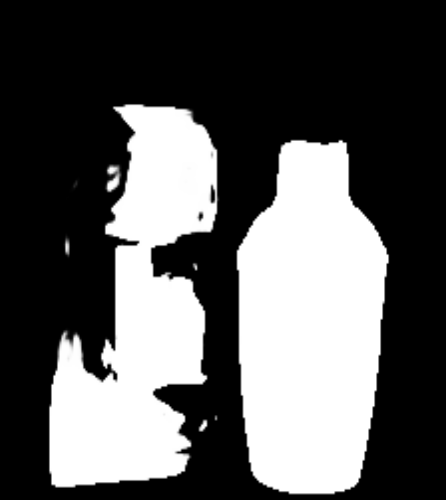}} 
		&
		\makecell[c]{\includegraphics[width=0.06\linewidth,height=0.05\linewidth]{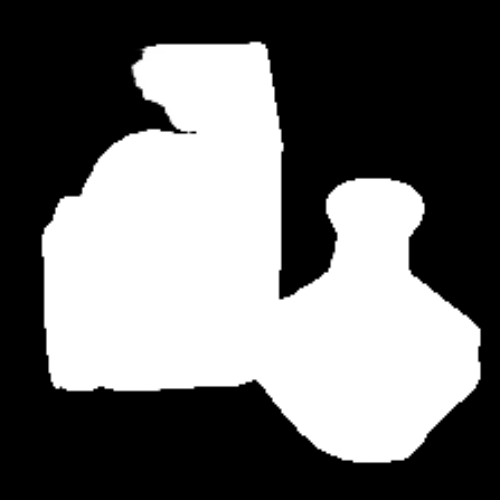}} 
		&
		\makecell[c]{\includegraphics[width=0.06\linewidth,height=0.05\linewidth]{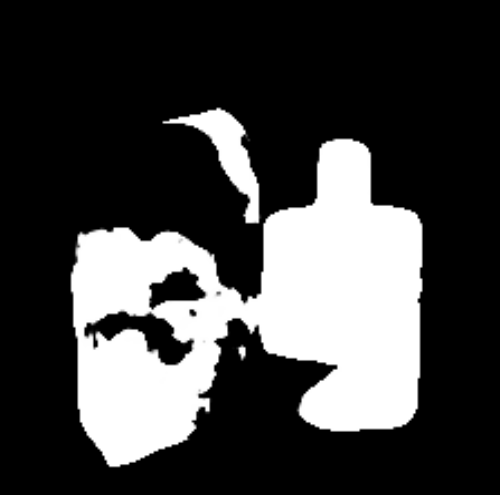}} 
		&
		\makecell[c]{\includegraphics[width=0.06\linewidth,height=0.05\linewidth]{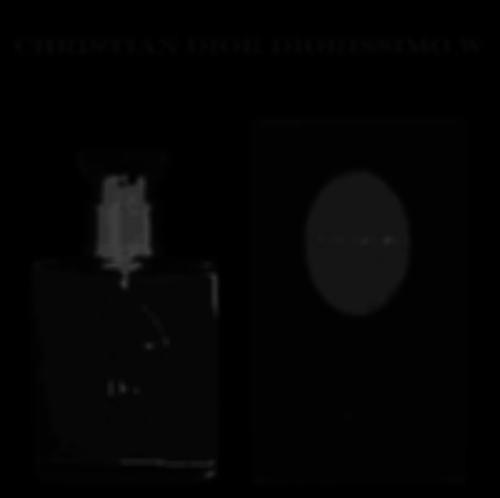}}
            &
		\makecell[c]{\includegraphics[width=0.06\linewidth,height=0.05\linewidth]{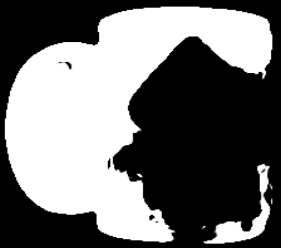}} 
		&
		\makecell[c]{\includegraphics[width=0.06\linewidth,height=0.05\linewidth]{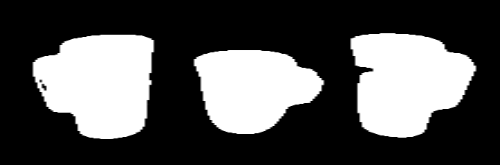}} 
		&
		\makecell[c]{\includegraphics[width=0.06\linewidth,height=0.05\linewidth]{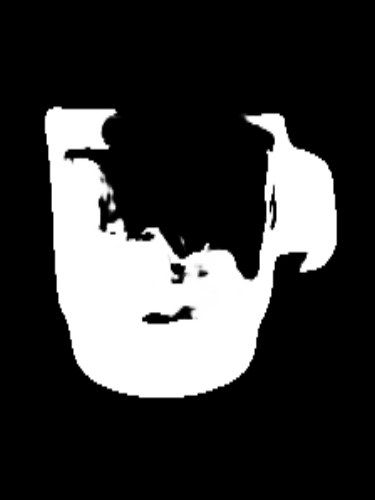}}
		&
		\makecell[c]{\includegraphics[width=0.06\linewidth,height=0.05\linewidth]{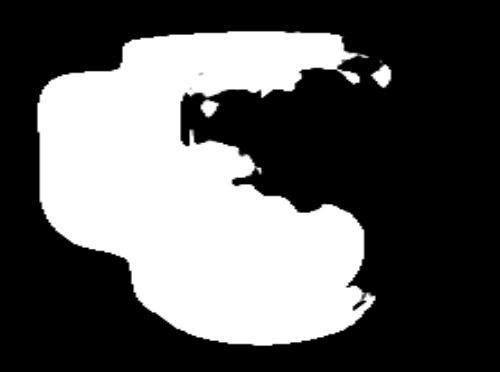}}
		&
		\makecell[c]{\includegraphics[width=0.06\linewidth,height=0.05\linewidth]{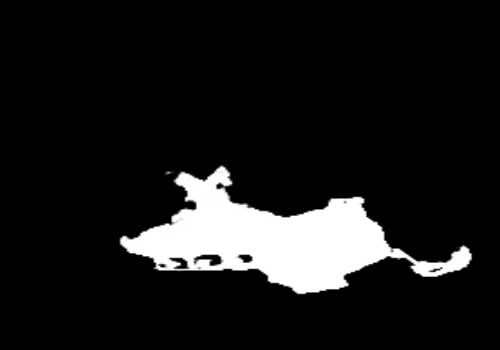}} 
		&
		\makecell[c]{\includegraphics[width=0.06\linewidth,height=0.05\linewidth]{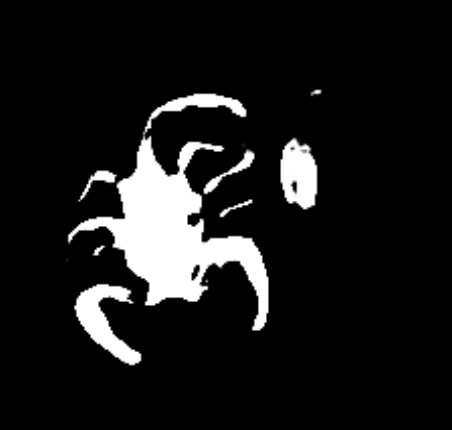}} 
		&
		\makecell[c]{\includegraphics[width=0.06\linewidth,height=0.05\linewidth]{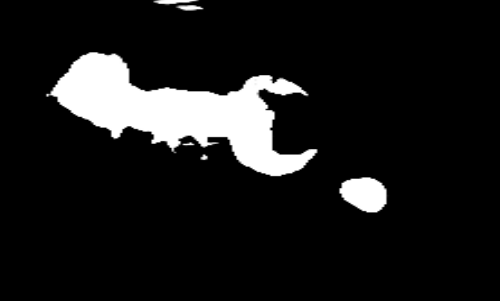}} 	
		&
		\makecell[c]{\includegraphics[width=0.06\linewidth,height=0.05\linewidth]{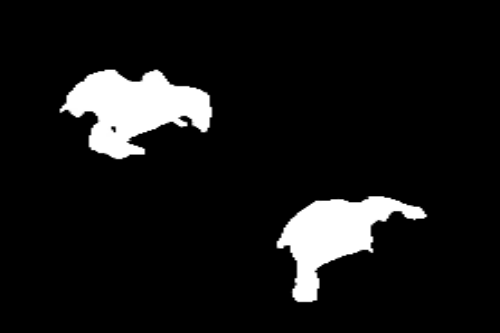}} 
		\vspace{-0.05cm}
            \\

            \rotatebox[origin=c]{90}{UFO}
            &
		\makecell[c]{\includegraphics[width=0.06\linewidth,height=0.05\linewidth]{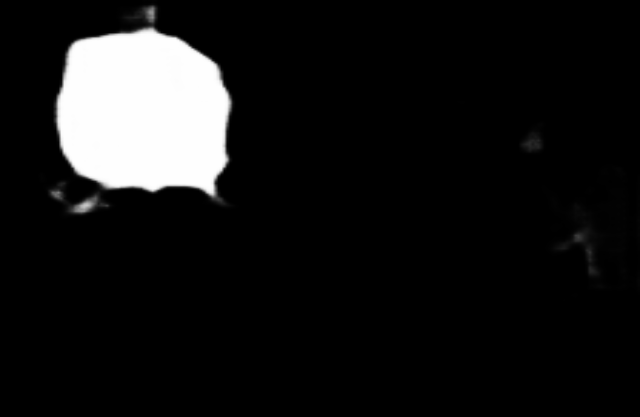}} 
		&
		\makecell[c]{\includegraphics[width=0.06\linewidth,height=0.05\linewidth]{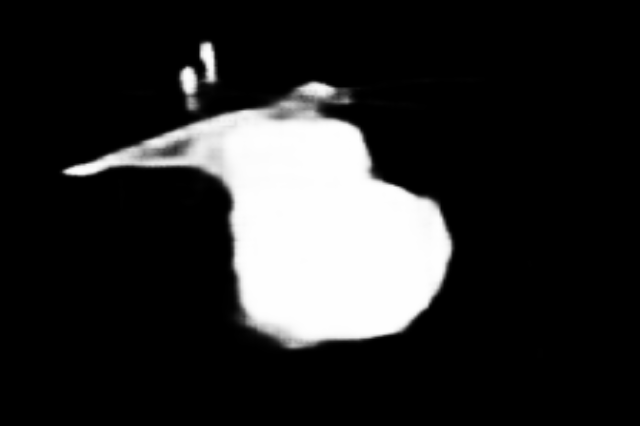}}
		&
		\makecell[c]{\includegraphics[width=0.06\linewidth,height=0.05\linewidth]{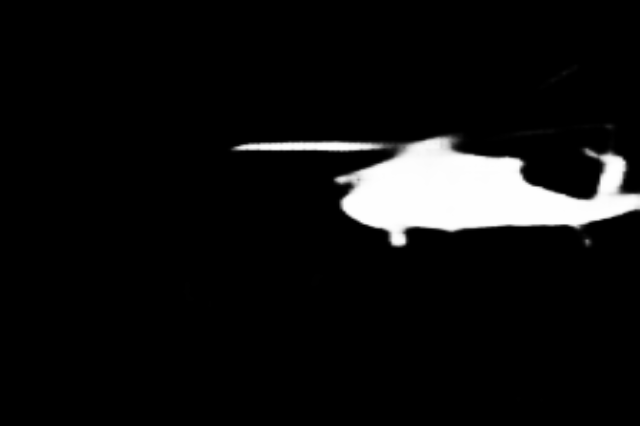}} 
		&
		\makecell[c]{\includegraphics[width=0.06\linewidth,height=0.05\linewidth]{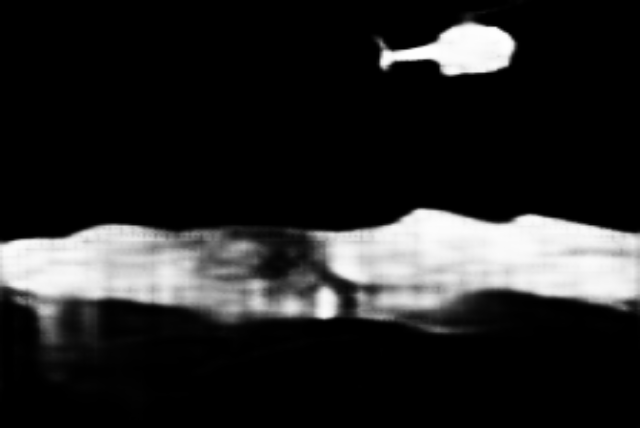}} 
		&
		\makecell[c]{\includegraphics[width=0.06\linewidth,height=0.05\linewidth]{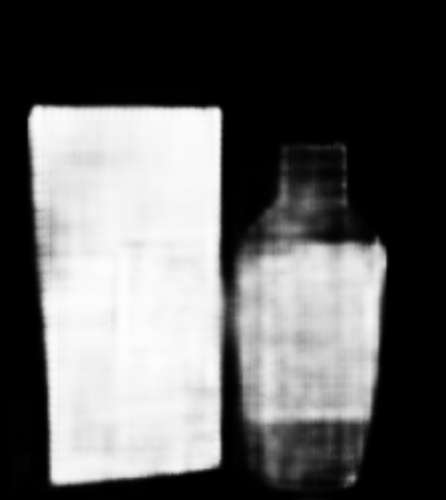}} 
		&
		\makecell[c]{\includegraphics[width=0.06\linewidth,height=0.05\linewidth]{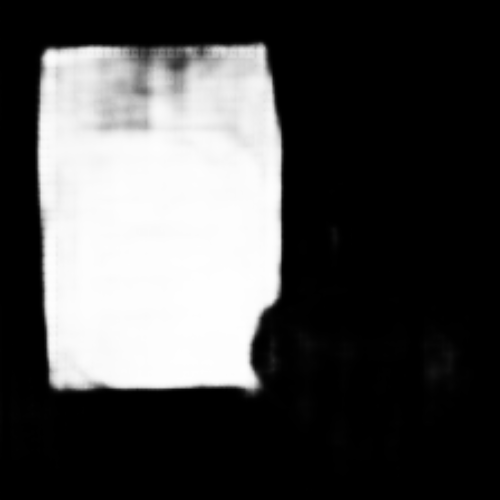}} 
		&
		\makecell[c]{\includegraphics[width=0.06\linewidth,height=0.05\linewidth]{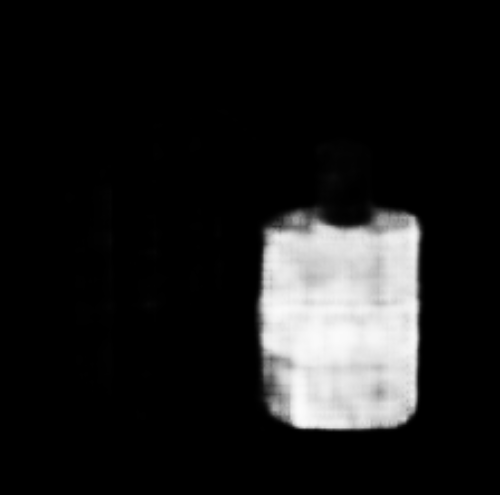}} 
		&
		\makecell[c]{\includegraphics[width=0.06\linewidth,height=0.05\linewidth]{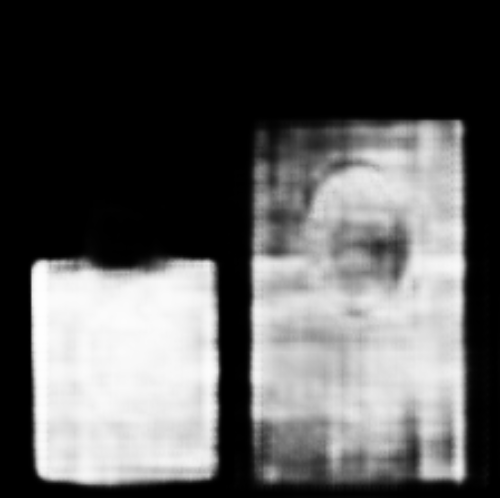}}
            &
		\makecell[c]{\includegraphics[width=0.06\linewidth,height=0.05\linewidth]{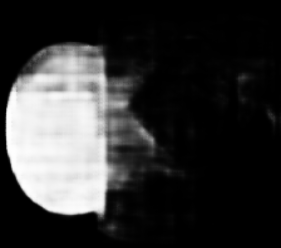}} 
		&
		\makecell[c]{\includegraphics[width=0.06\linewidth,height=0.05\linewidth]{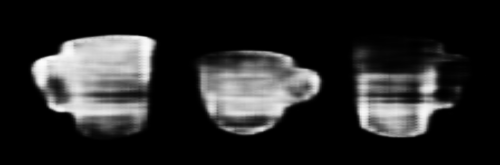}} 
		&
		\makecell[c]{\includegraphics[width=0.06\linewidth,height=0.05\linewidth]{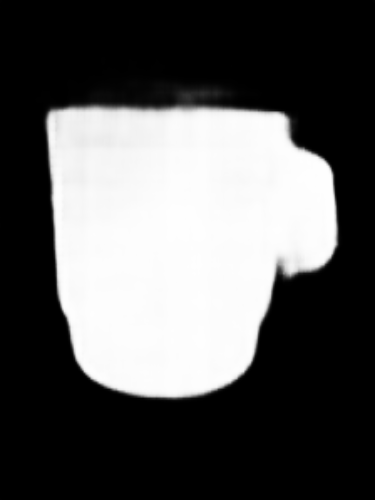}}
		&
		\makecell[c]{\includegraphics[width=0.06\linewidth,height=0.05\linewidth]{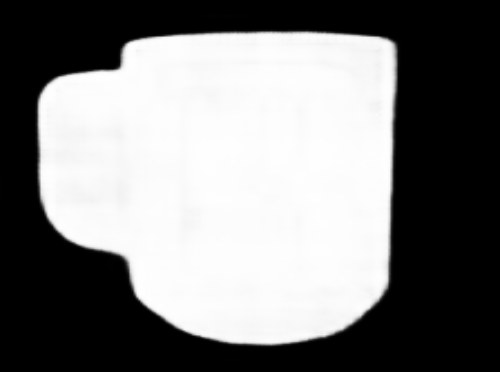}}
		&
		\makecell[c]{\includegraphics[width=0.06\linewidth,height=0.05\linewidth]{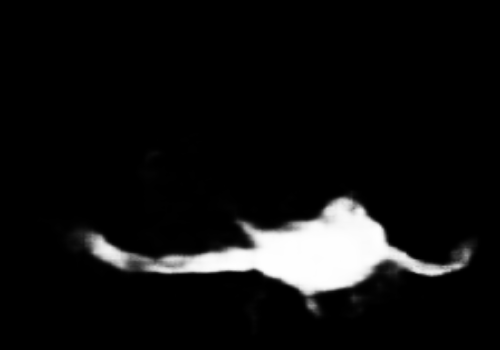}} 
		&
		\makecell[c]{\includegraphics[width=0.06\linewidth,height=0.05\linewidth]{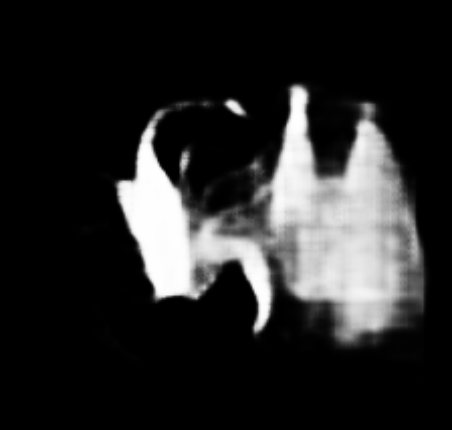}} 
		&
		\makecell[c]{\includegraphics[width=0.06\linewidth,height=0.05\linewidth]{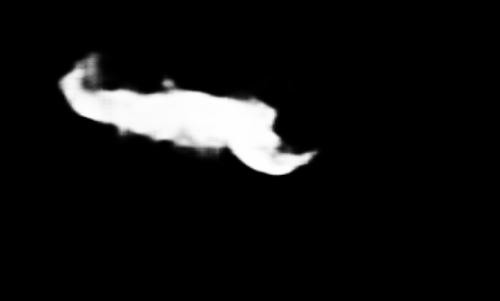}} 	
		&
		\makecell[c]{\includegraphics[width=0.06\linewidth,height=0.05\linewidth]{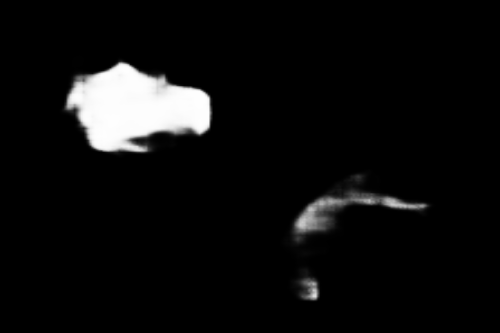}} 
		\vspace{-0.05cm}
            \\

            \rotatebox[origin=c]{90}{CADC}
            &
		\makecell[c]{\includegraphics[width=0.06\linewidth,height=0.05\linewidth]{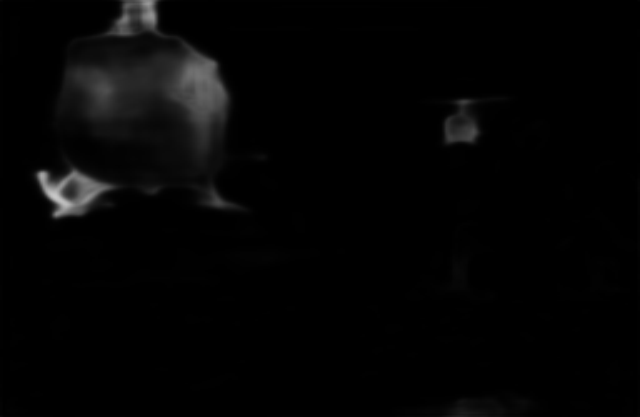}} 
		&
		\makecell[c]{\includegraphics[width=0.06\linewidth,height=0.05\linewidth]{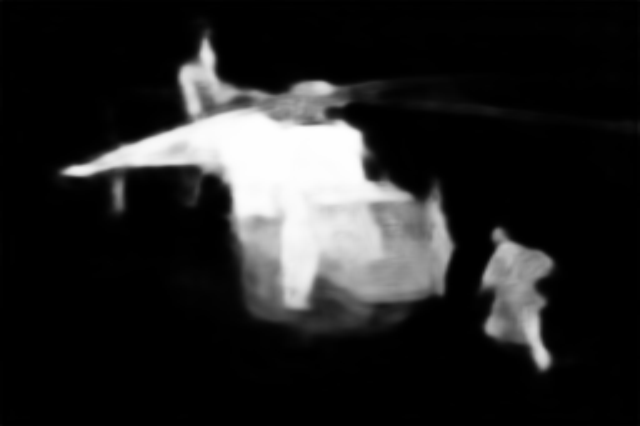}}
		&
		\makecell[c]{\includegraphics[width=0.06\linewidth,height=0.05\linewidth]{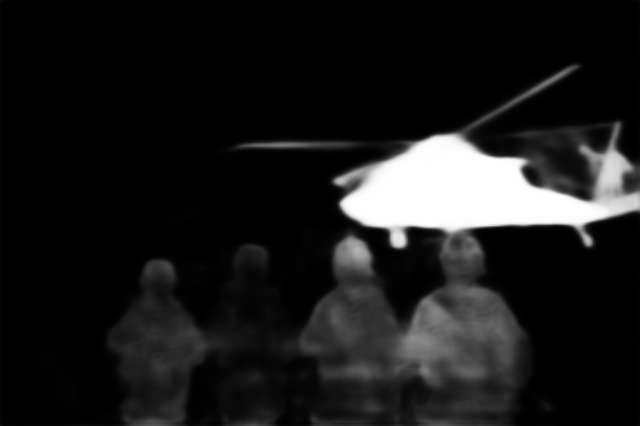}} 
		&
		\makecell[c]{\includegraphics[width=0.06\linewidth,height=0.05\linewidth]{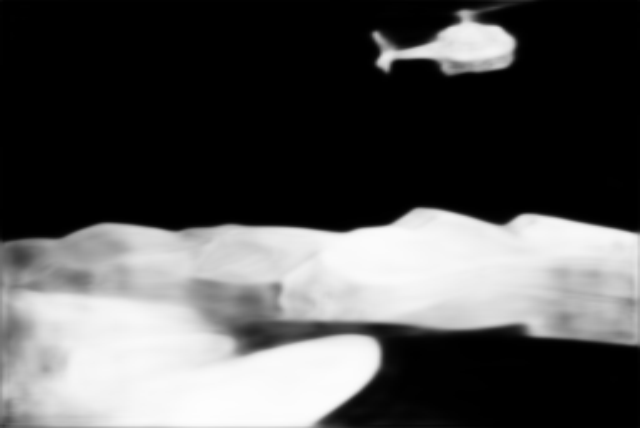}} 
		&
		\makecell[c]{\includegraphics[width=0.06\linewidth,height=0.05\linewidth]{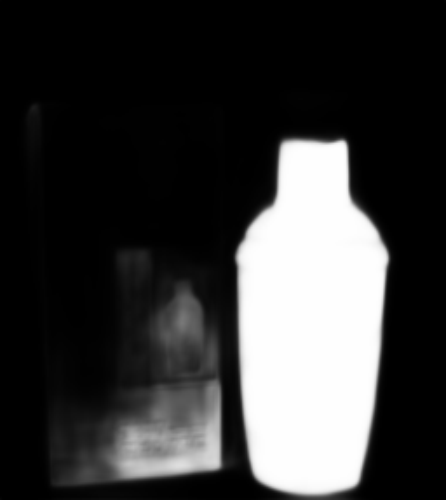}} 
		&
		\makecell[c]{\includegraphics[width=0.06\linewidth,height=0.05\linewidth]{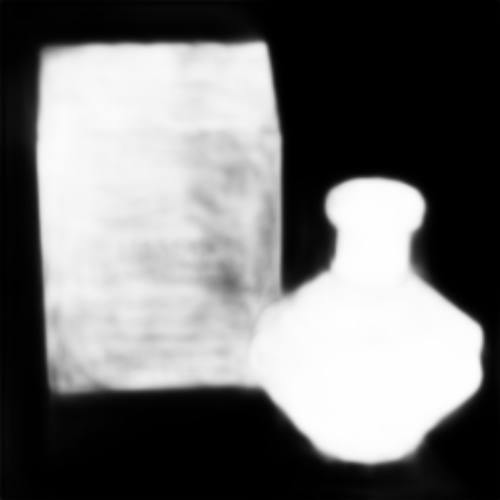}} 
		&
		\makecell[c]{\includegraphics[width=0.06\linewidth,height=0.05\linewidth]{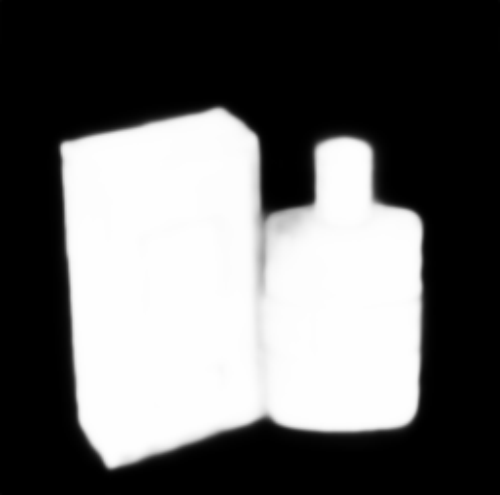}} 
		&
		\makecell[c]{\includegraphics[width=0.06\linewidth,height=0.05\linewidth]{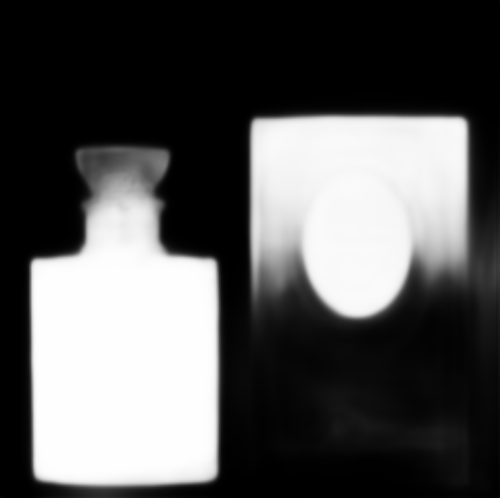}}
            &
		\makecell[c]{\includegraphics[width=0.06\linewidth,height=0.05\linewidth]{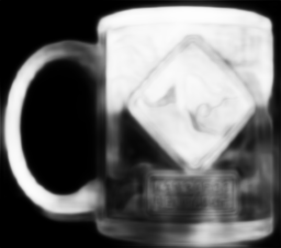}} 
		&
		\makecell[c]{\includegraphics[width=0.06\linewidth,height=0.05\linewidth]{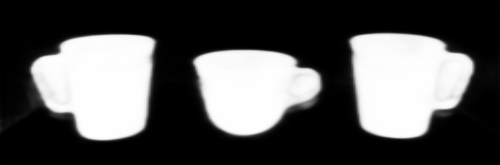}} 
		&
		\makecell[c]{\includegraphics[width=0.06\linewidth,height=0.05\linewidth]{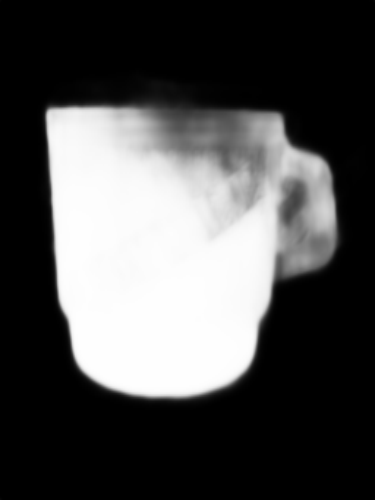}}
		&
		\makecell[c]{\includegraphics[width=0.06\linewidth,height=0.05\linewidth]{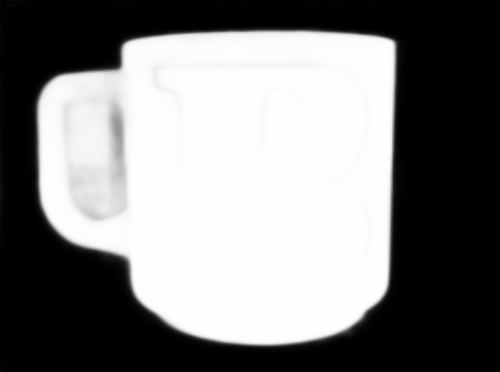}}
		&
		\makecell[c]{\includegraphics[width=0.06\linewidth,height=0.05\linewidth]{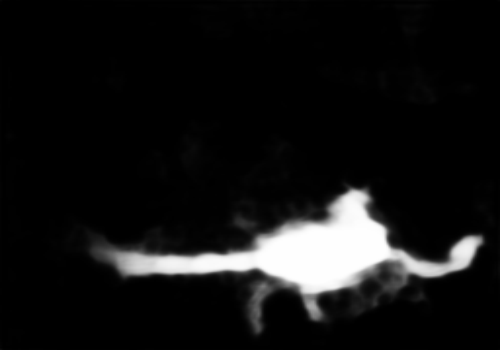}} 
		&
		\makecell[c]{\includegraphics[width=0.06\linewidth,height=0.05\linewidth]{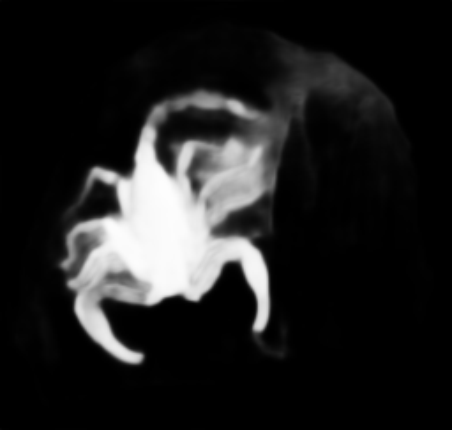}} 
		&
		\makecell[c]{\includegraphics[width=0.06\linewidth,height=0.05\linewidth]{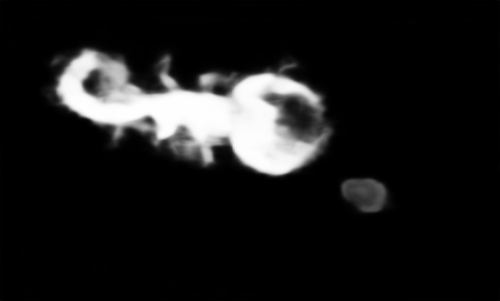}} 	
		&
		\makecell[c]{\includegraphics[width=0.06\linewidth,height=0.05\linewidth]{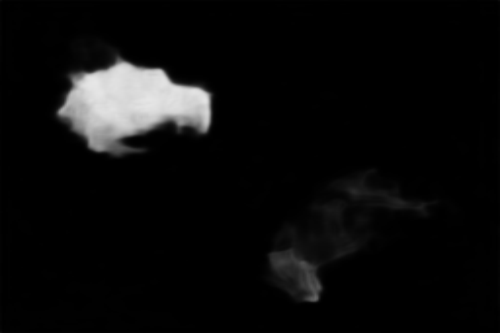}} 
		\vspace{-0.05cm}
            \\

            \rotatebox[origin=c]{90}{GCoNet}
            &
		\makecell[c]{\includegraphics[width=0.06\linewidth,height=0.05\linewidth]{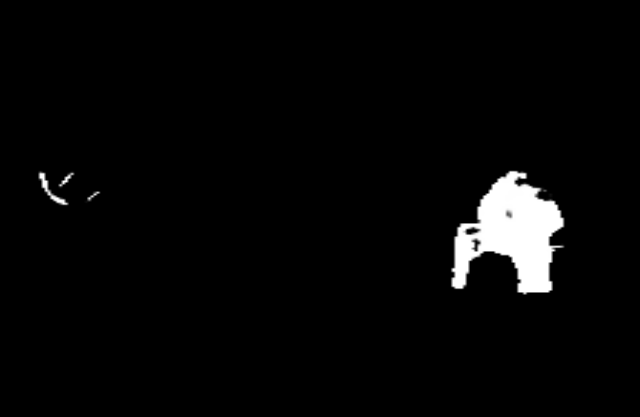}} 
		&
		\makecell[c]{\includegraphics[width=0.06\linewidth,height=0.05\linewidth]{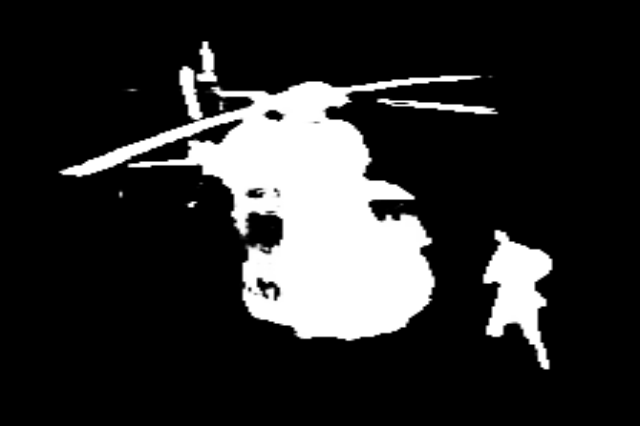}}
		&
		\makecell[c]{\includegraphics[width=0.06\linewidth,height=0.05\linewidth]{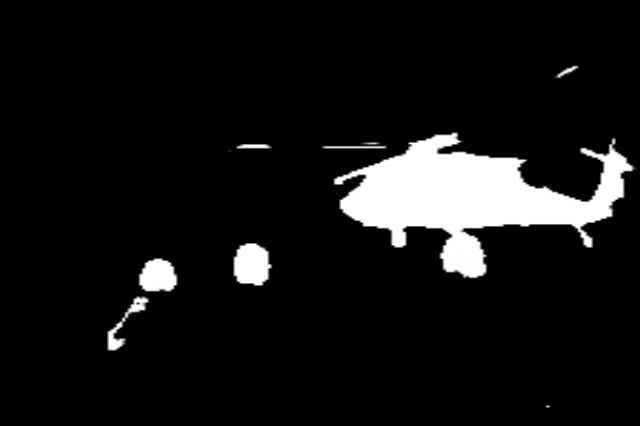}} 
		&
		\makecell[c]{\includegraphics[width=0.06\linewidth,height=0.05\linewidth]{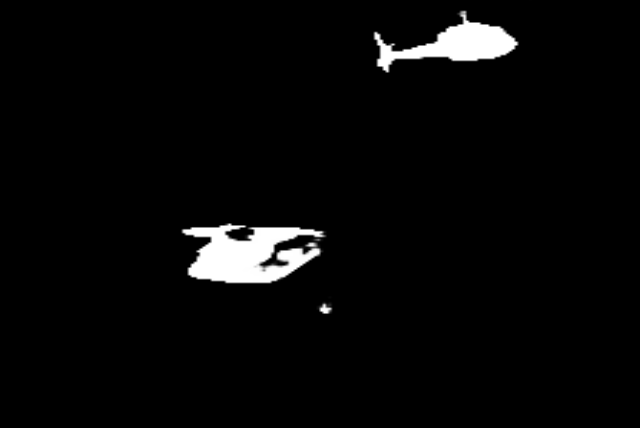}} 
		&
		\makecell[c]{\includegraphics[width=0.06\linewidth,height=0.05\linewidth]{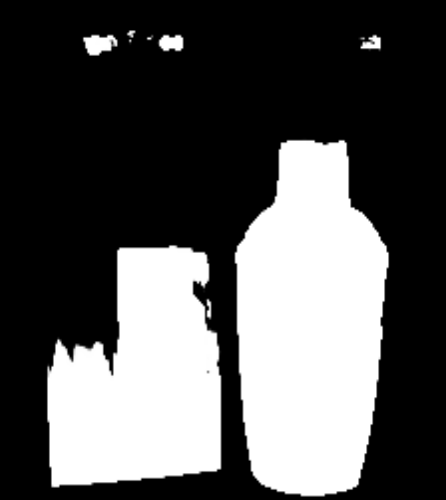}} 
		&
		\makecell[c]{\includegraphics[width=0.06\linewidth,height=0.05\linewidth]{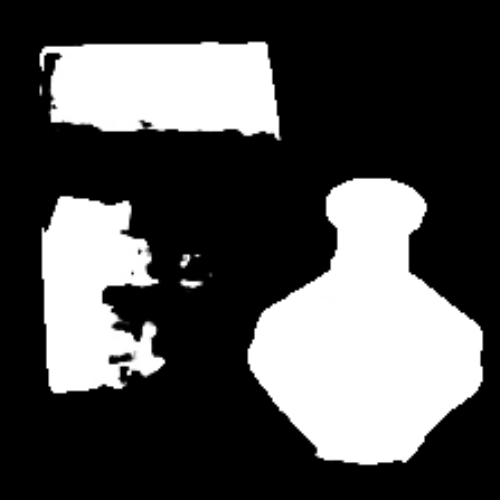}} 
		&
		\makecell[c]{\includegraphics[width=0.06\linewidth,height=0.05\linewidth]{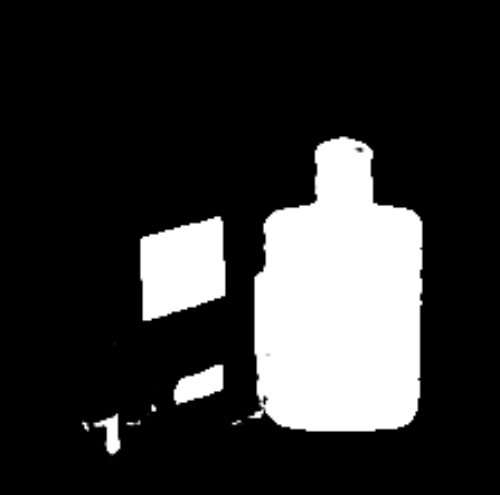}} 
		&
		\makecell[c]{\includegraphics[width=0.06\linewidth,height=0.05\linewidth]{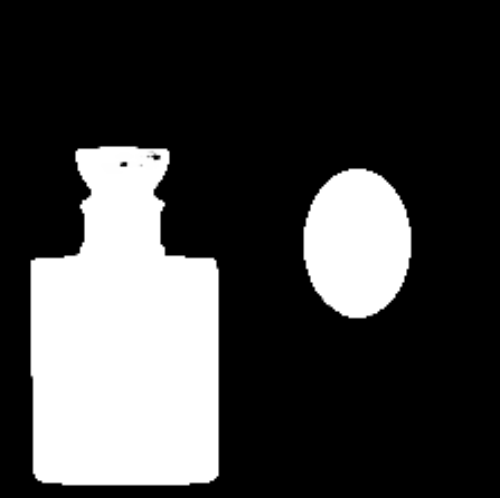}}
            &
		\makecell[c]{\includegraphics[width=0.06\linewidth,height=0.05\linewidth]{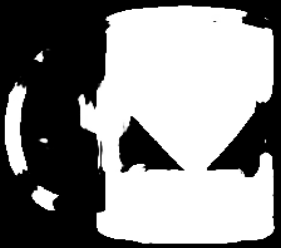}} 
		&
		\makecell[c]{\includegraphics[width=0.06\linewidth,height=0.05\linewidth]{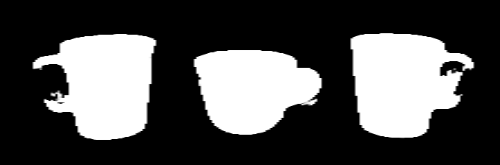}} 
		&
		\makecell[c]{\includegraphics[width=0.06\linewidth,height=0.05\linewidth]{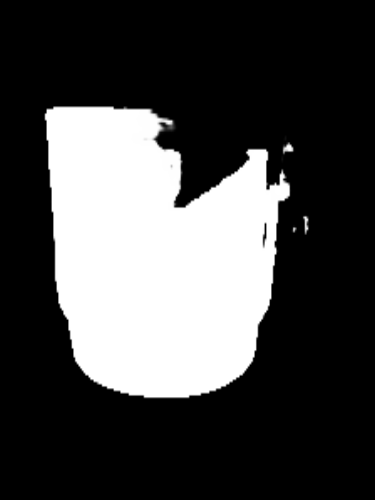}}
		&
		\makecell[c]{\includegraphics[width=0.06\linewidth,height=0.05\linewidth]{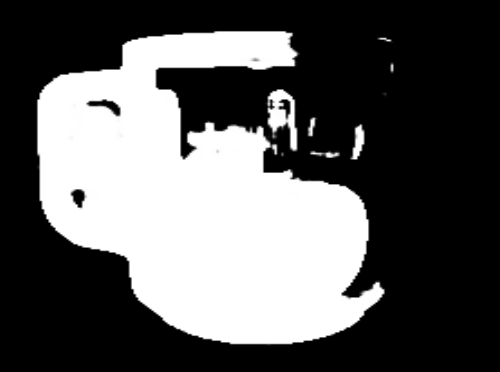}}
		&
		\makecell[c]{\includegraphics[width=0.06\linewidth,height=0.05\linewidth]{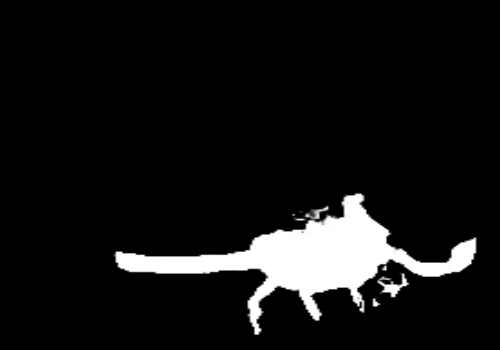}} 
		&
		\makecell[c]{\includegraphics[width=0.06\linewidth,height=0.05\linewidth]{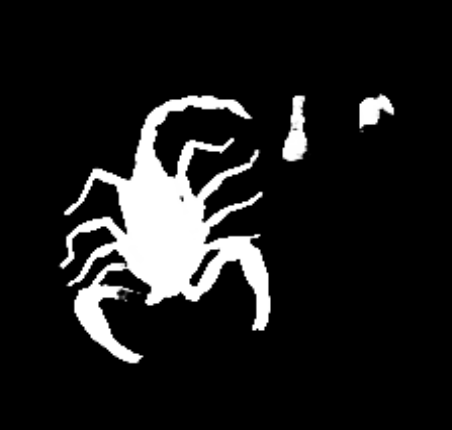}} 
		&
		\makecell[c]{\includegraphics[width=0.06\linewidth,height=0.05\linewidth]{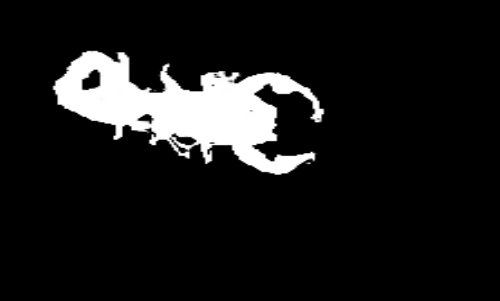}} 	
		&
		\makecell[c]{\includegraphics[width=0.06\linewidth,height=0.05\linewidth]{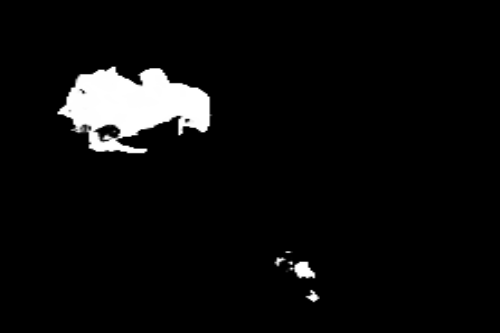}} 
		\vspace{-0.05cm}
            \\

            \rotatebox[origin=c]{90}{ICNet}
            &
		\makecell[c]{\includegraphics[width=0.06\linewidth,height=0.05\linewidth]{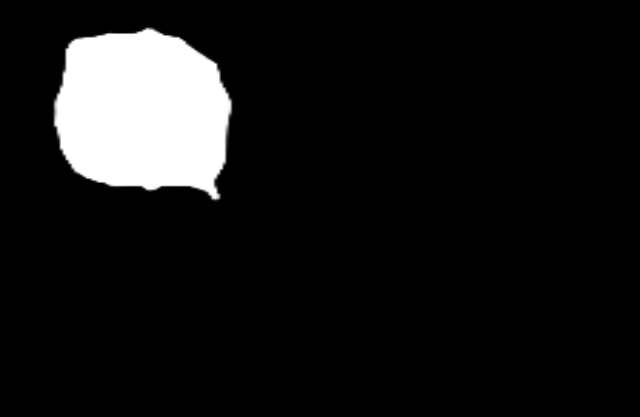}} 
		&
		\makecell[c]{\includegraphics[width=0.06\linewidth,height=0.05\linewidth]{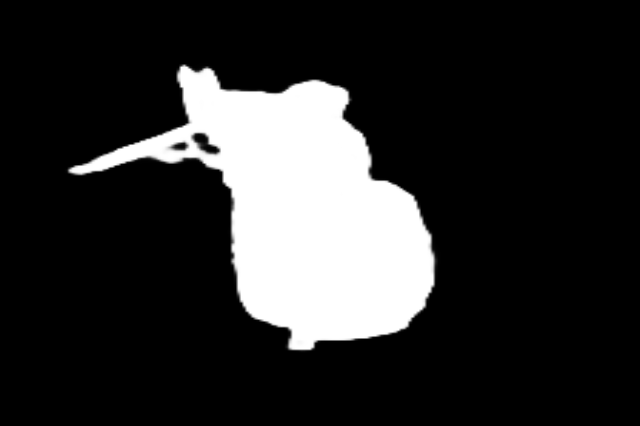}}
		&
		\makecell[c]{\includegraphics[width=0.06\linewidth,height=0.05\linewidth]{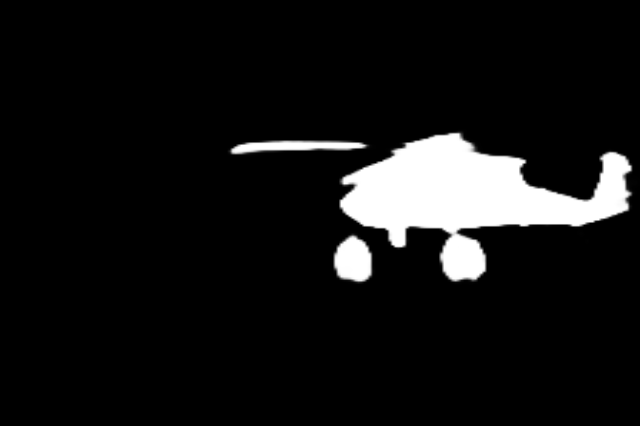}} 
		&
		\makecell[c]{\includegraphics[width=0.06\linewidth,height=0.05\linewidth]{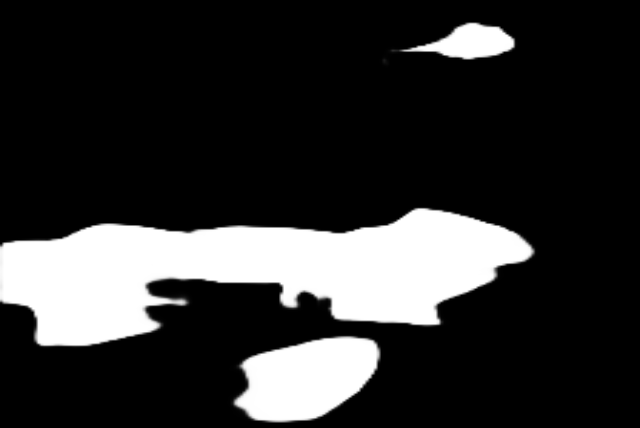}} 
		&
		\makecell[c]{\includegraphics[width=0.06\linewidth,height=0.05\linewidth]{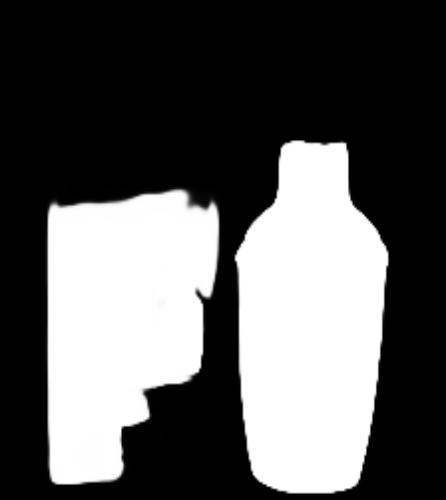}} 
		&
		\makecell[c]{\includegraphics[width=0.06\linewidth,height=0.05\linewidth]{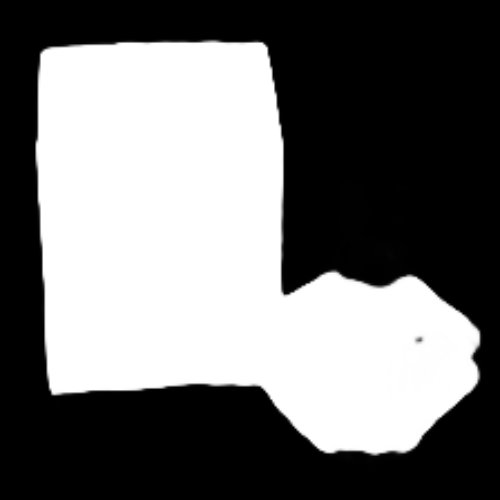}} 
		&
		\makecell[c]{\includegraphics[width=0.06\linewidth,height=0.05\linewidth]{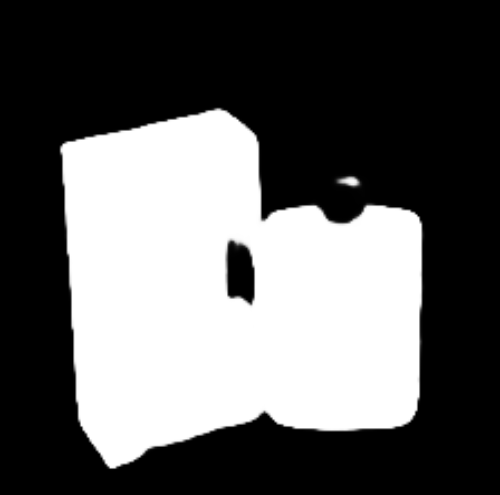}} 
		&
		\makecell[c]{\includegraphics[width=0.06\linewidth,height=0.05\linewidth]{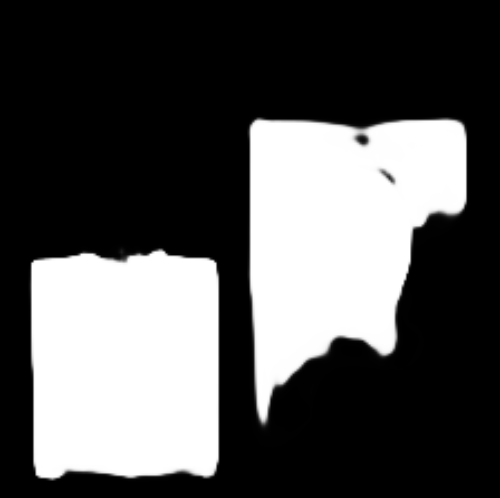}}
            &
		\makecell[c]{\includegraphics[width=0.06\linewidth,height=0.05\linewidth]{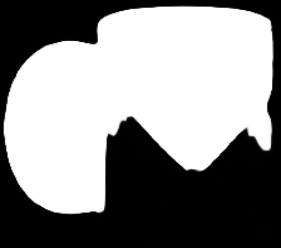}} 
		&
		\makecell[c]{\includegraphics[width=0.06\linewidth,height=0.05\linewidth]{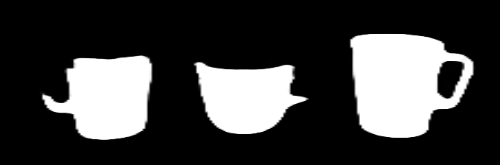}} 
		&
		\makecell[c]{\includegraphics[width=0.06\linewidth,height=0.05\linewidth]{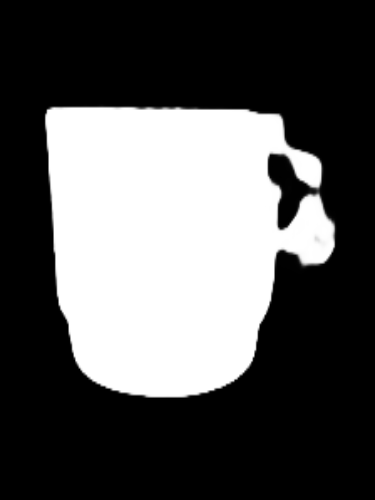}}
		&
		\makecell[c]{\includegraphics[width=0.06\linewidth,height=0.05\linewidth]{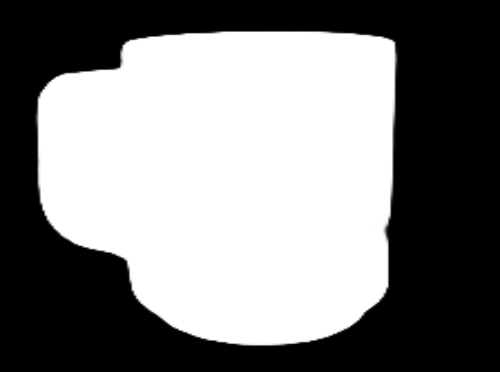}}
		&
		\makecell[c]{\includegraphics[width=0.06\linewidth,height=0.05\linewidth]{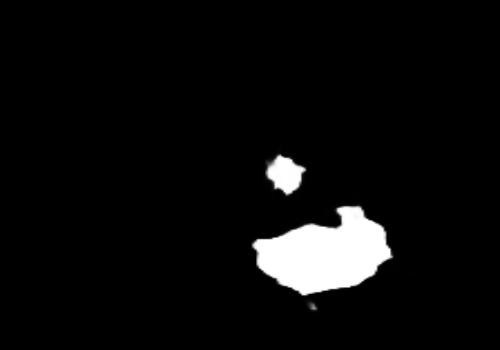}} 
		&
		\makecell[c]{\includegraphics[width=0.06\linewidth,height=0.05\linewidth]{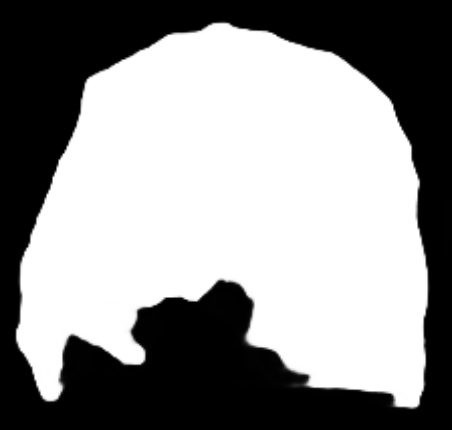}} 
		&
		\makecell[c]{\includegraphics[width=0.06\linewidth,height=0.05\linewidth]{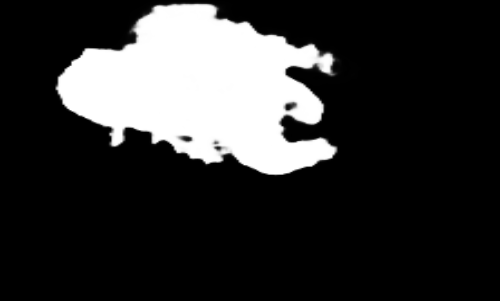}} 	
		&
		\makecell[c]{\includegraphics[width=0.06\linewidth,height=0.05\linewidth]{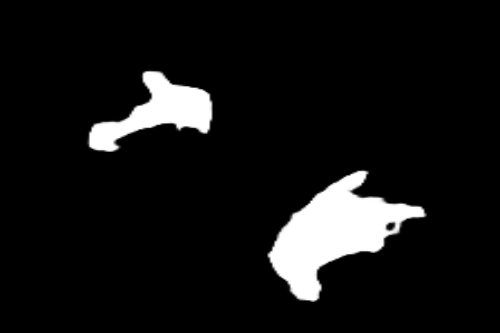}} 
            \\
		\end{tabular}
		\vspace{-4mm}
		\caption{
		\textbf{Qualitative comparisons of our model with other state-of-the-art methods.}
		}
		\label{SOTAFig}
		\vspace{-4mm}
\end{figure*}

\Paragraph{Effectiveness of TGFR.}
Finally, we add TGFR to leverage the learned tokens for refining the segmentation features. Table~\ref{ablationTab} shows that adopting TGFR
can bring more performance gains, thus demonstrating its effectiveness. We also visualize some feature maps and predictions of using and without using TGFR in Figure~\ref{fig:visual_TGFR}.
It can be seen that using TGFR obtains more discriminative features for distinguishing co-saliency objects from distractors, thus generating better segmentation results.

To dive deeper into the effectiveness of TGFR, we report more experimental results in Table~\ref{tab: detail_TGFR} for further analysis. First, we directly fuse the tokens and the segmentation features without performing the distillation process (``w/o Distillation"). we find this model brings limited improvements compared to the ``w/o TGFR" model. It is probably because the tokens and features might exist semantic gap, being detrimental for their fusion, hence verifying the necessity of our distillation mechanism. Next, we supplement the distillation process and explore four strategies for the refusion process, \ie individually refusing the distilled co-saliency (``w/ co'') or BG features (``w/ bg'') to the segmentation features, or refusing both with the order of co-saliency feature first (``w/ co\&bg'') or BG feature first (``w/ bg\&co''). We can find refusing both achieves better performance, thus verifying the necessity of leveraging both features for discrimination enhancement.
We also find first refusing the co-saliency feature and then integrating the BG feature obtains the best results. Thus, we adopt this strategy in our final TGFR design.

\Paragraph{Effectiveness of BG Exploration.} 
We remove all BG-related modules in our final model and only explore co-saliency regions, shown in the last row of Table~\ref{ablationTab}.
In this setting, CtP2T can not be used while only the co-saliency feature is used in TGFR. We find that the performance significantly drops compared to our final model, thus verifying the necessity of explicit BG modeling.

\Paragraph{Quantitative Analysis}. As shown in Figure~\ref{ablationFig}, we also provide some visual comparison samples for the four key components. We find that the baseline model is easily distracted by complex BG regions, while progressively introducing our four components can gradually exclude these distractors and achieve more and more accurate results.

\section{Comparison with State-of-the-Art Methods}
\vspace{-1mm}
We compare our model with other seven state-of-the-art methods, \ie
CSMG \cite{zhang2019co}, 
GICD \cite{zhang2020gicd}, ICNet \cite{jin2020icnet}, GCoNet \cite{fan2021GCoNet}, CADC \cite{zhang2021summarize}, UFO \cite{su2022unified}, and DCFM \cite{yu2022democracy}.
We report the quantitative comparison results in Table~\ref{SOTATab}. We can observe that our proposed DMT achieves the best performance on all three benchmark datasets.
Especially, on CoSal2015 and CoSOD3k, our DMT model surpasses the second-best model by a large margin, \eg 3.14\% $S_m$ and 4.07\% maxF on CoSal2015 and 3.23\% $S_m$ and 3.08\% maxF on CoSOD3k. We also show some visual comparison results in Figure~\ref{SOTAFig}. We can find that our method can precisely detect co-salient objects in complex scenarios, \eg the existence of extraneous salient objects with similar appearances to target objects, and target objects with small sizes. Nevertheless, other models are heavily distracted in these challenging scenes.

\vspace{-2mm}
\section{Conclusions}
\vspace{-1mm}
In this paper, we propose DMT, a transformer-based CoSOD model for explicitly mining both co-saliency and BG information and effectively modeling their discrimination. Specifically, we propose several economic multi-grained correlations, \ie R2R, CtP2T, and CoT2T to model inter-image and intra-image relations.
Besides, we propose a TGFR module to leverage the detection information for improving the discriminability of the segmentation features. 
It is an improvement to the MaskFormer that allows the mutual promotion of two sub-paths.  
Our model achieves a new state-of-the-art result.

\vspace{-4mm}
\paragraph{Acknowledgments:} This work was supported in part by Key-Area Research and Development Program of
Guangdong Province (No.2021B0101200001),  the National Key R\&D Program of China under Grant 2021B0101200001, and the National Science Foundation of China under Grant 62036011, U20B2065, 721A0001, 62136004.

{\small
\bibliographystyle{ieee_fullname}
\bibliography{egbib}
}

\end{document}